\documentclass[10pt,journal,compsoc]{IEEEtran}

\usepackage{amsmath,amsfonts}
\usepackage{graphicx}
\usepackage{booktabs}
\usepackage{multirow}
\usepackage{pifont}
\usepackage{xcolor}
\usepackage{url}
\usepackage{colortbl}
\usepackage{capt-of}
\usepackage{makecell}
\usepackage[T1]{fontenc}
\ifCLASSOPTIONcompsoc
  \usepackage{cite}
\else
  \usepackage{cite}
\fi

\ifCLASSOPTIONcompsoc
  \usepackage[caption=false,font=normalsize,labelfont=sf,textfont=sf]{subfig}
\else
  \usepackage[caption=false,font=footnotesize]{subfig}
\fi

\usepackage[hidelinks]{hyperref}

\newcommand{\ie}{\textit{i.e.}}

\begin{document}

\title{Uni-AdaVD: Universal Concept Erasure for Visual Generation via Orthogonal Value Decomposition}

\author{Qifan Zhou*, Yuan Wang*, Yanbin Hao, Xiang Wang, Kuien Liu, Richang Hong,~\IEEEmembership{Senior Member,~IEEE,}
and Meng Wang,~\IEEEmembership{Fellow,~IEEE}%
\IEEEcompsocitemizethanks{%
\IEEEcompsocthanksitem Qifan Zhou, Yanbin Hao, Richang Hong, and Meng Wang are with the School of Computer Science and Information Engineering, Hefei University of Technology, Hefei, China.\protect\\
E-mail: zhouqifan1432@gmail.com, haoyanbin@hotmail.com, hongrc.hfut@gmail.com, eric.mengwang@gmail.com
\IEEEcompsocthanksitem Yuan Wang is with the School of Cyber Science and Technology, University of Science and Technology of China, Hefei, China.\protect\\
E-mail: wy1001@mail.ustc.edu.cn
\IEEEcompsocthanksitem Xiang Wang is with the School of Artificial Intelligence and Data Science, University of Science and Technology of China, Hefei, China.\protect\\
E-mail: xiangwang@ustc.edu.cn
\IEEEcompsocthanksitem Kuien Liu is with the Institute of Software, Chinese Academy of Sciences, Hefei, China.\protect\\
E-mail: kuien@iscas.ac.cn
\IEEEcompsocthanksitem Qifan Zhou and Yuan Wang contributed equally to this work.}%
}

\markboth{IEEE Transactions on Pattern Analysis and Machine Intelligence,~Vol.~XX, No.~XX, Month~2026}%
{Zhou \MakeLowercase{\textit{et al.}}: Uni-AdaVD}

\IEEEtitleabstractindextext{%
\begin{abstract}
Visual generative models inevitably absorb undesirable concepts from uncurated pretraining data, making concept erasure essential for safe deployment. Existing erasure methods, however, are often architecture-specific and struggle to remove target concepts while preserving non-target content and generative priors. We present Uni-AdaVD, a universal inference-time concept erasure framework for visual generation. Uni-AdaVD treats the value space of multimodal attention as a unified intervention space and introduces encoder-aware target representation construction to localize target semantics across heterogeneous text encoders. It further combines orthogonal value decomposition with an adaptive erasing shift to suppress target semantic directions without updating the original model weights. Extensive experiments on U-Net-, DiT-, and autoregressive image generators, as well as text-to-video models, demonstrate strong performance on single- and multi-concept erasure while preserving non-target priors. These results suggest that Uni-AdaVD provides an efficient and adaptable safety mechanism for modern visual generative models. Our code is available at \url{https://github.com/QifanZhou/Uni-AdaVD}.
\end{abstract}

}

\maketitle

\IEEEdisplaynontitleabstractindextext
\IEEEpeerreviewmaketitle

\IEEEraisesectionheading{\section{Introduction}\label{sec:intro}}

\IEEEPARstart{V}{isual} generative models have recently made significant progress, encompassing diverse paradigms such as diffusion \cite{ho2020denoising,song2020score,song2020denoising,ho2022classifier,lipman2022flow,xu2023multimodal} and visual autoregressive models \cite{van2016conditional,esser2021taming,chang2022maskgit,tian2024visual,sun2024autoregressive,wang2021generative} across both image \cite{dhariwal2021diffusion,rombach2022high,peebles2023scalable,podell2024sdxl,esser2024scaling,huang2025diffusion,he2025freeedit} and video generation \cite{ho2022video,blattmann2023stable,brooks2024video,croitoru2023diffusion,hong2022depth,yang2025cogvideox} domains. Despite their structural differences, these models share a common reliance on large-scale, uncurated Internet data for pretraining. Consequently, they inevitably absorb undesirable semantics, including copyrighted entities, specific artistic styles, and unsafe content. Since data curation and retraining models from scratch are prohibitively expensive, concept erasure has become a practical and necessary solution for safe alignment and compliant deployment.

The central goal of concept erasure is to precisely remove the target concept, \ie, erasure efficacy, while preserving unrelated visual content and the model's generative prior, \ie, prior preservation. Existing fine-tuning \cite{gandikota2023erasing,kumari2023ablating,lu2024mace,lyu2024one}, editing-based \cite{orgad2023editing,gandikota2024unified,gong2024reliable,bui2025fantastic}, and inference-time \cite{schramowski2023safe,yoon2025safree,han2025groce} methods have made progress toward this trade-off. However, most of them are developed for specific generative architectures, such as U-Net-based \cite{li2025speed,chen2025trce,biswas2026cure} or DiT-based diffusion models \cite{gao2025eraseanything,zhang2026differential}. Their effectiveness often relies on architecture-specific decisions about where to intervene and how to construct the erasure objective. Consequently, transferring them to heterogeneous generative frameworks usually requires non-trivial redesign rather than direct plug-and-play deployment.

A more desirable framework should be architecture-agnostic and readily applicable to diverse visual generative models. Inference-time methods are especially promising for this goal because they intervene during generation without retraining or permanently editing model parameters. Our previous work, \textbf{AdaVD} (Adaptive Value Decomposer) \cite{wang2025precise}, offers preliminary evidence for this direction by showing that value orthogonal decomposition can effectively balance erasure efficacy and prior preservation across a broad range of U-Net-based diffusion models. However, AdaVD was not originally formulated as a universal framework, and its adaptation to heterogeneous generative architectures remains underexplored. In this work, we systematically generalize value orthogonal decomposition into a universal concept erasure framework for diverse visual generative models.

To build a universal inference-time concept erasure framework, three key questions must be addressed: \textit{where to erase}, \textit{what to erase}, and \textit{how to erase safely}. The first question, ``\textit{where to erase},'' requires identifying a common semantic space in which concept erasure can be consistently performed across different generative architectures, forming the foundation of a universal framework. The second question, ``\textit{what to erase},'' involves accurately locating the target semantics within this space. This is non-trivial because explicit concepts are often associated with specific object tokens, whereas implicit safety-related concepts may be scattered across ordinary contextual tokens and heterogeneous text encoders such as CLIP and T5. The third question, ``\textit{how to erase safely},'' focuses on preserving irrelevant content and the model's generative prior. This challenge is particularly pronounced in DiT-based image generators, video generators, and other models with strong multimodal interactions, where coarse interventions can easily disrupt the generative distribution, causing structural collapse in images or degraded visual consistency in videos.

To address these challenges, we introduce Uni-AdaVD, a universal inference-time concept erasure framework for diverse visual generative models. For ``\textit{where to erase},'' Uni-AdaVD uses the value space of multimodal attention layers as the common intervention space across different generative models, inspired by AdaVD. For ``\textit{what to erase},'' we propose Encoder-aware Target Representation Construction (ETRC), which introduces encoder-specific probing to collect scattered semantic features and align target semantics across heterogeneous text encoders. For ``\textit{how to erase safely},'' we combine Orthogonal Value Decomposition (OVD) with Layer-wise Adaptive Erasing Shift (LAES), which adaptively controls erasure strength across network depths to remove target semantics while preserving non-target content and generative priors.

Our main contributions are summarized as follows:
\begin{itemize}
    \item \textbf{Universal Inference-time Concept Erasure:}
    We propose Uni-AdaVD, a universal inference-time concept erasure framework that performs concept erasure via value orthogonal decomposition across diverse visual generative models, including U-Net-based, DiT-based, autoregressive image generation, and video generation models.

    \item \textbf{Adaptive Localization and Erasure:}
    We introduce encoder-specific probing to localize scattered target semantics across heterogeneous text encoders, together with a layer-wise adaptive erasing shift to adaptively control erasure strength for prior-preserving concept removal.

    \item \textbf{Extensive Evaluation and Robustness:}
    Comprehensive experiments demonstrate that Uni-AdaVD achieves effective concept erasure across diverse generative architectures while preserving non-target content, with strong robustness against both black-box and white-box adversarial prompt attacks.
\end{itemize}

\section{Related Work}
\label{sec:related_work}

\subsection{Concept Erasure in U-Net-Based Diffusion Models}

Early concept erasure studies mainly focused on U-Net-based latent diffusion models. In these models, text conditions are usually injected into visual features through cross-attention layers. Based on how the target concept is removed, existing methods can be divided into three main groups: fine-tuning, editing-based, and inference-time methods.

Fine-tuning methods remove target concepts by fine-tuning the base model or learning auxiliary parameters. ESD~\cite{gandikota2023erasing} uses the conditioned and unconditioned predictions of a frozen diffusion model as negative guidance to fine-tune the model, thereby discouraging the updated model from generating the target concept. Concept Ablation (CA)~\cite{kumari2023ablating} redirects a target concept toward a predefined anchor concept and replaces its original semantics with a safer or more general one. To improve parameter efficiency, SPM~\cite{lyu2024one} learns lightweight concept-specific adapters instead of updating the full model. MACE~\cite{lu2024mace} extends this line to mass concept erasure by combining closed-form cross-attention refinement, concept-specific LoRA fine-tuning, and interference-aware multi-LoRA integration. A closely related line formulates concept removal as a machine unlearning problem. SalUn~\cite{fan2024salun} identifies the parameters that are most important for forgetting and focuses the updates on them. AdvUnlearn~\cite{zhang2024defensive} incorporates adversarial training to improve resistance to concept recovery attacks, while STEREO~\cite{srivatsan2025stereo} uses a two-stage optimization process to improve robust erasure. ReCARE~\cite{kim2026cooccurring} further emphasizes the preservation of benign concepts that frequently co-occur with the target concept.

Editing-based methods directly change selected model parameters without standard iterative fine-tuning. UCE~\cite{gandikota2024unified} uses a closed-form solution to update cross-attention parameters and supports the editing of several concepts at the same time. RECE~\cite{gong2024reliable} also removes target concepts through lightweight post-hoc parameter editing. These methods are usually more efficient than fine-tuning methods. However, they still change the original model weights, and their editing rules often depend on specific model components.

Inference-time methods suppress target concepts during inference without updating the original model parameters. Safe Latent Diffusion (SLD)~\cite{schramowski2023safe} changes classifier-free guidance during the denoising process to reduce unsafe generation. SAFREE~\cite{yoon2025safree} finds unsafe semantic directions in the text embedding space and moves the prompt representation away from these directions. It also adjusts the denoising process based on the input prompt. PGCE~\cite{cai2026prototype} uses concept prototypes to suppress broad and complex unsafe concepts during inference. These methods avoid model retraining and permanent parameter changes. However, they still depend on specific inference-time conditioning mechanisms, which limits their direct transfer across heterogeneous generative architectures.

\subsection{Concept Erasure in DiT-Based Diffusion Models}

Recent text-to-image models, such as SD v3~\cite{esser2024scaling} and FLUX~\cite{blackforestlabs2024fluxgithub}, use diffusion transformers and flow-matching objectives. Unlike U-Net-based diffusion models, these models often use dual-stream or single-stream transformer blocks to process text and image tokens. These changes affect both the generation process and the interaction between text and visual features. Therefore, many concept erasure methods designed for U-Net-based models cannot be directly applied to DiT-based models.

Existing methods for DiT-based models mainly include fine-tuning and inference-time methods. Among the fine-tuning methods, EraseAnything~\cite{gao2025eraseanything} uses bi-level optimization with LoRA adaptation and attention-map regularization to remove target concepts while preserving unrelated content. Z-Erase~\cite{jiang2026z} focuses on single-stream diffusion transformers and separates the effects of text and image tokens. It also uses adaptive erasure modulation to balance concept removal and content preservation. DVE~\cite{zhang2026differential} is an inference-time method for flow-matching models. It removes concept-specific components from the velocity field along a differential direction defined by the difference between the target and anchor concepts.

\begin{figure}
    \centering
    \includegraphics[width=0.6\linewidth]{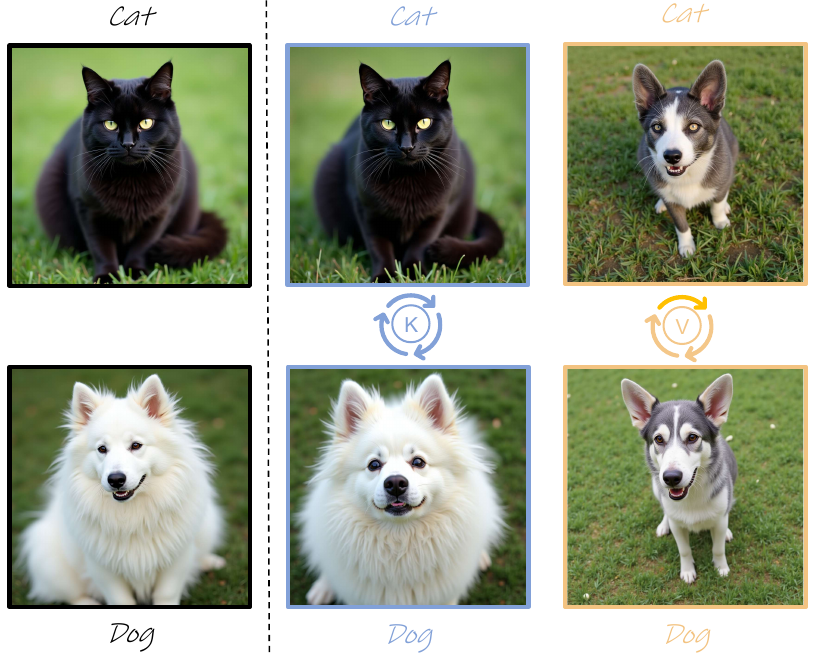}
    \vspace{-3mm}
    \caption{Value-space analysis in FLUX joint attention. Swapping the blue $\mathbf{KEY}$ sequences causes limited semantic change, whereas swapping the orange $\mathbf{VALUE}$ sequences changes the concept identity between ``\textit{Cat}'' and ``\textit{Dog},'' supporting value-space intervention.}
    \label{fig:flux_kv_analysis}
    \vspace{-3mm}
\end{figure}

\begin{figure*}[t]
  \centering
  \includegraphics[width=\textwidth]{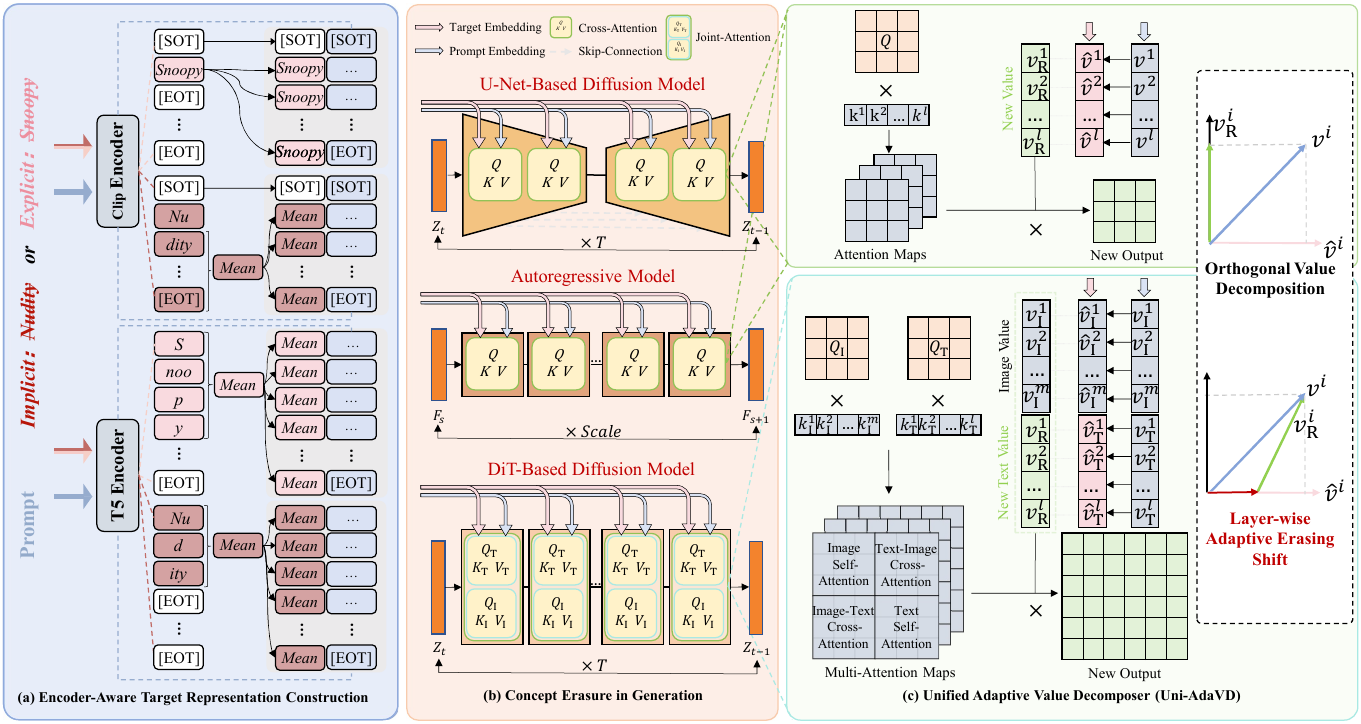}
  \vspace{-6mm}
  \caption{Overview of our Uni-AdaVD. \textbf{a. Encoder-aware Target Representation Construction:} For T5 \cite{raffel2020exploring} encoders, we average the valid tokens excluding padding and end-of-text tokens, tiling the result to the maximum sequence length. For CLIP \cite{radford2021learning} encoders, explicit concepts use the last subject token, while implicit concepts use the aggregated mean from the last subject to the end-of-text token, with both similarly tiled to the fixed sequence length. \textbf{b. Concept Erasure in Generation:} Demonstrates the integration of our value-space intervention mechanism within the generative pipelines of diverse architectures, including U-Net- and DiT-based diffusion models and AR models. \textbf{c. Uni-AdaVD:} Details the core operations of our framework, which employs Orthogonal Value Decomposition (OVD) to erase target concepts in the value space and utilizes Layer-wise Adaptive Erasing Shift (LAES) to adaptively modulate erasure strength for high-fidelity prior preservation.}
  \label{fig:overall_method}
  \vspace{-4mm}
\end{figure*}

\subsection{Concept Erasure in Autoregressive Models}

Autoregressive visual generators, such as VAR~\cite{tian2024visual}, Switti-AR~\cite{voronov2024switti}, and Infinity-2B~\cite{han2025infinity}, generate images token by token or scale by scale. Unlike diffusion models, they do not use an iterative denoising process. Therefore, concept erasure methods based on noise prediction, velocity fields, or denoising trajectories cannot be directly applied to autoregressive models.

Existing concept erasure methods for autoregressive visual generators are mainly fine-tuning. For instance, EAR~\cite{fan2025earerasingconceptsunified} fine-tunes autoregressive models with Windowed Gradient Accumulation and Thresholded Loss Masking to remove target concepts while reducing damage to unrelated content. VCE~\cite{han2025vce} builds contrastive image pairs and uses DPO-based training to suppress unsafe concepts while preserving non-target generation.

\section{Preliminaries}
\label{subsec:prelim}

Text-conditioned visual generators inject textual semantics into the visual generation process through attention modules. Let $\mathbf{Q}\in\mathbb{R}^{N_{\mathrm{q}}\times d}$, $\mathbf{K}\in\mathbb{R}^{N_{\mathrm{k}}\times d}$, and $\mathbf{V}\in\mathbb{R}^{N_{\mathrm{k}}\times d}$ denote the query, key, and value matrices, respectively, where $N_{\mathrm{q}}$ and $N_{\mathrm{k}}$ are the query and key/value sequence lengths, and $d$ is the attention feature dimension. The attention output is
\begin{equation}
\operatorname{Attention}(\mathbf{Q}, \mathbf{K}, \mathbf{V})
=
\operatorname{softmax}\left(\frac{\mathbf{Q}\mathbf{K}^T}{\sqrt{d}}\right)\mathbf{V}.
\label{eq:attn_formula_general}
\end{equation}
Since the output is a weighted linear combination of value vectors, the value space provides a natural interface for controlling injected textual semantics, while $\mathbf{Q}$ and $\mathbf{K}$ mainly determine attention routing and allocation \cite{tewel2023key}. We omit layer indices in the main text unless necessary.

We next summarize this attention interface in U-Net-based, DiT-based, and autoregressive generators.

\subsection{U-Net-Based Generators}

In U-Net-based latent diffusion models, let $\mathbf{Z}_t$ denote the noisy visual features at denoising step $t$, and let $\mathbf{C}$ denote the contextualized text features. The queries are computed from noisy image features, whereas the keys and values are computed from text features:
\begin{equation}
\mathbf{Q} = \mathbf{W}_{\mathrm{Q}}\mathbf{Z}_t,\quad
\mathbf{K} = \mathbf{W}_{\mathrm{K}}\mathbf{C},\quad
\mathbf{V} = \mathbf{W}_{\mathrm{V}}\mathbf{C},
\label{eq:unet_qkv}
\end{equation}
where $\mathbf{W}_{\mathrm{Q}}$, $\mathbf{W}_{\mathrm{K}}$, and $\mathbf{W}_{\mathrm{V}}$ are learnable projection matrices. This formulation makes value-space intervention directly applicable to U-Net-based models.

\subsection{DiT-Based Generators}

DiT-based generators represent the visual state at denoising step $t$ as an image-token sequence and often adopt tighter multimodal coupling. In joint-attention architectures such as FLUX and SD v3, modality-specific queries, keys, and values are first computed as
\begin{equation}
\begin{aligned}
&\mathbf{Q}_{\mathrm{I}} = \mathbf{W}_{\mathrm{Q},\mathrm{I}}\mathbf{Z}_t,\quad
\mathbf{K}_{\mathrm{I}} = \mathbf{W}_{\mathrm{K},\mathrm{I}}\mathbf{Z}_t,\quad
\mathbf{V}_{\mathrm{I}} = \mathbf{W}_{\mathrm{V},\mathrm{I}}\mathbf{Z}_t,\\
&\mathbf{Q}_{\mathrm{T}} = \mathbf{W}_{\mathrm{Q},\mathrm{T}}\mathbf{C},\quad
\mathbf{K}_{\mathrm{T}} = \mathbf{W}_{\mathrm{K},\mathrm{T}}\mathbf{C},\quad
\mathbf{V}_{\mathrm{T}} = \mathbf{W}_{\mathrm{V},\mathrm{T}}\mathbf{C},
\end{aligned}
\label{eq:joint_qkv_def}
\end{equation}
where the subscripts $\mathrm{I}$ and $\mathrm{T}$ denote the image and text branches, respectively. Accordingly, $\mathbf{Q}_{\mathrm{I}}, \mathbf{K}_{\mathrm{I}}, \mathbf{V}_{\mathrm{I}} \in \mathbb{R}^{L_{\mathrm{I}}\times d}$ and $\mathbf{Q}_{\mathrm{T}}, \mathbf{K}_{\mathrm{T}}, \mathbf{V}_{\mathrm{T}} \in \mathbb{R}^{L_{\mathrm{T}}\times d}$, where $L_{\mathrm{I}}$ and $L_{\mathrm{T}}$ denote the image and text sequence lengths. These features are concatenated along the sequence dimension as $\mathbf{Q}=[\mathbf{Q}_{\mathrm{I}};\mathbf{Q}_{\mathrm{T}}]$, $\mathbf{K}=[\mathbf{K}_{\mathrm{I}};\mathbf{K}_{\mathrm{T}}]$, and $\mathbf{V}=[\mathbf{V}_{\mathrm{I}};\mathbf{V}_{\mathrm{T}}]$, so that $\mathbf{Q}, \mathbf{K}, \mathbf{V}\in\mathbb{R}^{(L_{\mathrm{I}}+L_{\mathrm{T}})\times d}$. The resulting attention is
\begin{equation}
\operatorname{JointAttention}(\mathbf{Q}, \mathbf{K}, \mathbf{V})
=
\operatorname{softmax}\left(\frac{\mathbf{Q}\mathbf{K}^T}{\sqrt{d}}\right)\mathbf{V}.
\label{eq:joint_attn_formula}
\end{equation}

Compared with U-Net-based cross-attention, joint attention introduces stronger text-image coupling. As shown in Fig.~\ref{fig:flux_kv_analysis}, swapping $\mathbf{V}$ changes the generated subject identity much more than swapping $\mathbf{K}$, supporting the value space as the primary intervention interface. In practice, we intervene only on the text value component $\mathbf{V}_{\mathrm{T}}$ to better preserve visual priors.

\subsection{Autoregressive Generators}

Autoregressive text-to-image generators synthesize images token by token or scale by scale rather than through iterative denoising. Let $\mathbf{H}_k$ denote the visual hidden states at decoding step $k$. Text conditions are still injected through attention:
\begin{equation}
\mathbf{Q} = \mathbf{W}_{\mathrm{Q}}\mathbf{H}_k,\quad
\mathbf{K} = \mathbf{W}_{\mathrm{K}}\mathbf{C},\quad
\mathbf{V} = \mathbf{W}_{\mathrm{V}}\mathbf{C}.
\label{eq:ar_qkv}
\end{equation}
Therefore, autoregressive generators are also compatible with the same value-space intervention view.

Overall, although U-Net-based, DiT-based, and autoregressive generators differ in visual parameterization and update dynamics, they all admit attention-based text conditioning, which motivates a unified value-space formulation for concept erasure across architectures.

\section{Method}

\noindent This section presents \textbf{Uni-AdaVD} (Universal Adaptive Value Decomposer), a universal inference-time framework for concept erasure in text-conditioned visual generators. Despite their architectural differences, U-Net-based, DiT-based, and autoregressive models all rely on attention-mediated text injection, which exposes the text value space as a shared intervention interface. As illustrated in Fig.~\ref{fig:overall_method}, Uni-AdaVD builds on this observation by constructing encoder-aware target representation (ETRC, Section~\ref{subsec:anchoring_broadcasting}, Fig.~\ref{fig:overall_method}(a)), removing target-aligned components via Orthogonal Value Decomposition (OVD; Section~\ref{subsec:ovd}, Fig.~\ref{fig:overall_method}(b)), and adaptively modulating the erasure strength through Layer-wise Adaptive Erasing Shift (LAES; Section~\ref{subsec:aes}, Fig.~\ref{fig:overall_method}(c)) to better preserve non-target priors.

\subsection{Encoder-aware Target Representation Construction}
\label{subsec:anchoring_broadcasting}

After establishing the text value space as the common intervention interface, we next construct target representations tailored to different text encoders. This is necessary because semantic information is not distributed uniformly across encoder architectures. In particular, CLIP-family and T5-family text encoders exhibit markedly different token interaction patterns, which in turn lead to different semantic distributions.

\begin{figure}[t]
    \centering
    \includegraphics[width=\linewidth]{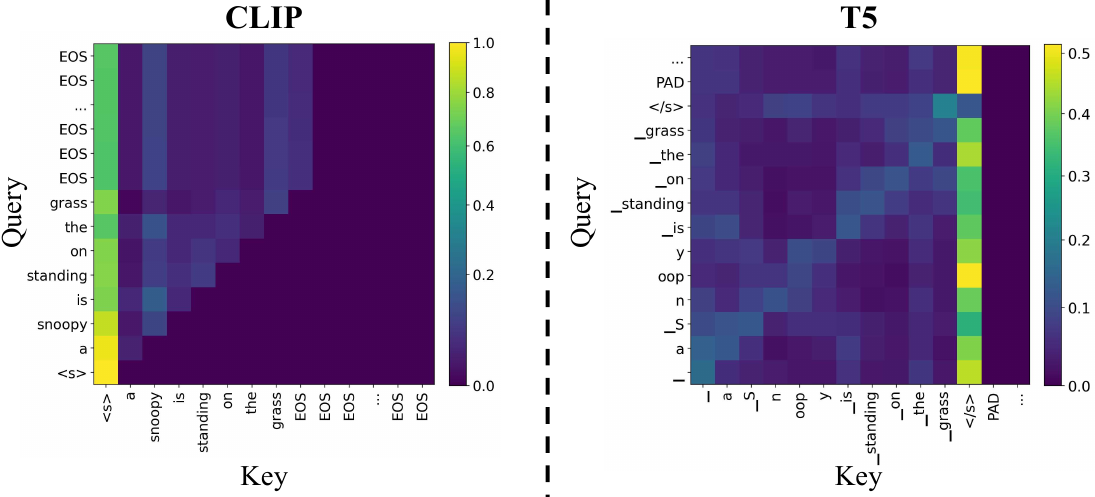}
    \vspace{-5mm}
    \caption{Visualization of attention maps in different text encoders. \textbf{Left (CLIP):} High-weight regions are strictly distributed on and above the anti-diagonal due to the causal mask. \textbf{Right (T5):} The bidirectional mechanism enables dense global interactions, allowing feature information to be shared across all valid tokens.}
    \label{fig:attention_maps}
    \vspace{-4mm}
\end{figure}

To characterize this difference, we visualize the encoder attention maps in Fig.~\ref{fig:attention_maps}. CLIP exhibits a causal attention pattern, in which each token attends only to itself and preceding tokens; as a result, contextual information progressively accumulates toward later positions. By contrast, T5 shows dense bidirectional interactions over valid tokens, indicating that semantic information is distributed across the sequence rather than concentrated near its end.

\begin{figure}[t]
    \centering
    \includegraphics[width=\linewidth]{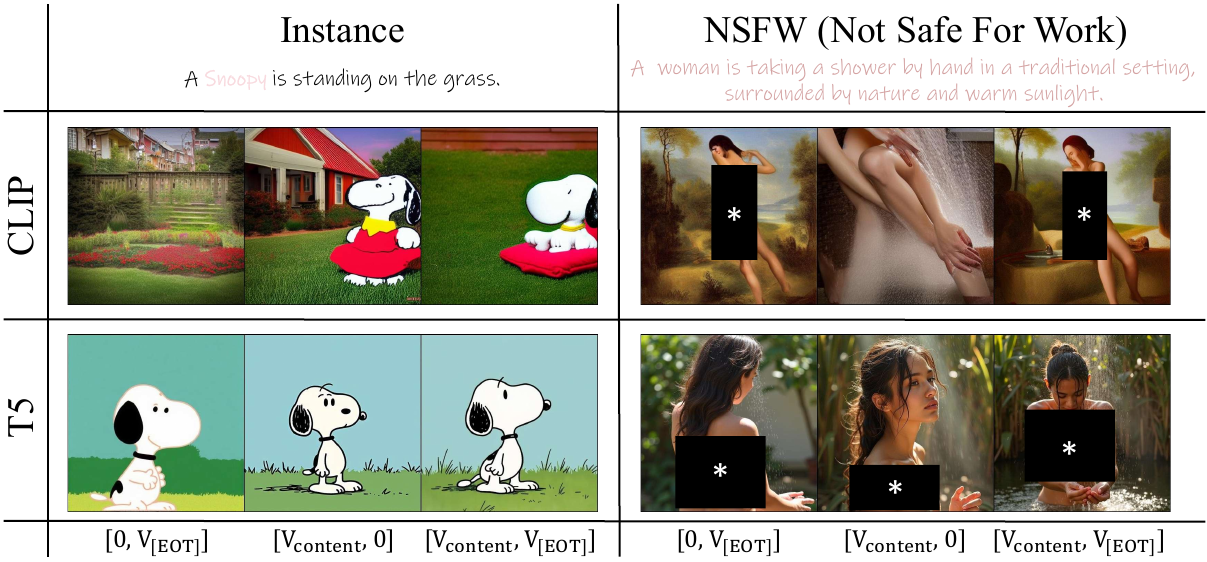}
    \vspace{-3mm}
    \caption{Token masking on value vectors. Settings $[V_{\text{content}}, 0]$, $[0, V_{\text{EOT}}]$, and $[V_{\text{content}}, V_{\text{EOT}}]$ denote interventions at content or [EOT] tokens. \textbf{Top, CLIP in SD v1-4:} Explicit concepts rely heavily on content tokens, whereas implicit concepts depend on the [EOT] region. \textbf{Bottom, T5 in FLUX:} The bidirectional T5 encoder disperses semantic features across the sequence, rendering local value masking less reliable for precise erasure.}
    \label{fig:token_masking}
    \vspace{-6mm}
\end{figure}

We further examine this behavior through token-level value masking, as shown in Fig.~\ref{fig:token_masking}. Specifically, we partition prompt tokens into content tokens, which describe the subject and its attributes, and end-of-text tokens, and then zero out the corresponding value vectors under three settings: $[\mathbf{0}, \mathbf{V}_{\mathrm{EOT}}]$, $[\mathbf{V}_{\mathrm{content}}, \mathbf{0}]$, and $[\mathbf{V}_{\mathrm{content}}, \mathbf{V}_{\mathrm{EOT}}]$. In CLIP-based SD v1-4, masking content tokens effectively removes explicit concepts such as ``\textit{Snoopy}'', whereas masking [EOT] tokens is more effective for suppressing implicit concepts such as ``\textit{Nudity}''. This suggests that, under causal attention, explicit concepts are concentrated near the last subject token, while implicit concepts accumulate toward the later part of the sequence. In T5-based FLUX, by contrast, neither content-token masking nor end-of-text-token masking reliably removes the target concept, indicating that semantic evidence is distributed across valid tokens.

Based on these observations, we propose Encoder-aware Target Representation Construction. Let $\mathbf{C}_t = [\boldsymbol{c}_{t,1}, \boldsymbol{c}_{t,2}, \dots, \boldsymbol{c}_{t,L}]^T \in \mathbb{R}^{L \times d_c}$ denote the contextualized token features of the target concept to be erased, where $L$ is the text sequence length and $d_c$ is the text-encoder feature dimension. Let $t_s$ denote the index of the last subject token. We extract an anchor $\boldsymbol{c}_a$ from the target concept and then broadcast it to construct a target feature sequence $\tilde{\mathbf{C}}_t$.

For CLIP, following the observed causal accumulation, we extract the semantic anchor $\boldsymbol{c}_a$ from the end of the subject-related token span. For explicit concepts, we set $\boldsymbol{c}_a = \boldsymbol{c}_{t,t_s}$, which provides the most localized representation. For implicit concepts, we instead average the token features from the last subject token $t_s$ to the [EOT] token, i.e., $\boldsymbol{c}_a = \frac{1}{N_{\mathrm{implicit}}} \sum_{i=t_s}^{i_{\mathrm{EOT}}} \boldsymbol{c}_{t,i}$, where $i_{\mathrm{EOT}}$ denotes the index of the [EOT] token and $N_{\mathrm{implicit}} = i_{\mathrm{EOT}} - t_s + 1$ is the span of implicit semantic tokens. After obtaining $\boldsymbol{c}_a$, we preserve the start token $\boldsymbol{c}_{t,1}$ and define $\tilde{\mathbf{C}}_t^{\mathrm{CLIP}} = [\boldsymbol{c}_{t,1}, \boldsymbol{c}_a, \dots, \boldsymbol{c}_a]^T$.

For T5, since the bidirectional mechanism disperses semantic evidence across the sequence, no individual token provides a sufficiently stable anchor. Therefore, unlike CLIP, we do not distinguish between explicit and implicit concepts for T5-based encoders, and instead use a unified sequence-level construction for both. Let $\mathcal{I}_{\mathrm{valid}} \subseteq \{1,2,\dots,L\}$ denote the set of valid token indices excluding padding and end-of-text tokens, and let $N_{\mathrm{valid}} = |\mathcal{I}_{\mathrm{valid}}|$. We then compute $\boldsymbol{c}_a$ as the mean of all valid token features, i.e., $\boldsymbol{c}_a = \frac{1}{N_{\mathrm{valid}}} \sum_{i \in \mathcal{I}_{\mathrm{valid}}} \boldsymbol{c}_{t,i}$, and define $\tilde{\mathbf{C}}_t^{\mathrm{T5}} = [\boldsymbol{c}_a, \dots, \boldsymbol{c}_a]^T$. This design treats the target concept as a sequence-level representation, reducing positional uncertainty introduced by bidirectional token interaction.

The reconstructed target feature sequence $\tilde{\mathbf{C}}_t$ is then projected into the value space, yielding the target value sequence $\tilde{\mathbf{V}}_t \in \mathbb{R}^{L \times d_v}$ in cross-attention and the textual target value sequence $\tilde{\mathbf{V}}_{\mathrm{T},t} \in \mathbb{R}^{L \times d_v}$ in joint attention, where $d_v$ is the value feature dimension. These projections are computed using the corresponding value projection matrices $\mathbf{W}_{\mathrm{V}}$ and $\mathbf{W}_{\mathrm{V},\mathrm{T}}$. The resulting target value sequences serve as the reference for the subsequent value-space intervention.

 AdaVD studied target value construction only for the CLIP text encoder and used a single construction strategy without distinguishing different types of target concepts. In this extended version, we first examine how target semantics are distributed in the value spaces of different text encoders, including both CLIP and T5. We further divide target concepts into explicit and implicit concepts according to their semantic forms and token-level distributions. Based on these findings, we design different target representations for explicit CLIP concepts, implicit CLIP concepts, and concepts encoded by T5. Therefore, the proposed encoder-aware construction extends AdaVD from a CLIP-specific target representation to a more general design that supports heterogeneous text encoders and different types of concepts.

\subsection{Orthogonal Value Decomposition}
\label{subsec:ovd}

Given the encoder-aware target representation constructed above, Uni-AdaVD removes target semantics by projecting text-derived value vectors onto the orthogonal complement of a target concept subspace. This yields a unified value-space operation: the target-aligned component is removed, while the residual component is retained for subsequent generation.

For a single target concept, let $\boldsymbol{v}^i \in \mathbb{R}^{d_v}$ denote the text-derived value vector at textual token position $i$, and let $\hat{\boldsymbol{v}}^i \in \mathbb{R}^{d_v}$ denote the corresponding target concept vector obtained from the target value representation, where $i \in \{1,2,\dots,L\}$ and $d_v$ is the value feature dimension. We remove the target concept by projecting $\boldsymbol{v}^i$ onto the orthogonal complement of $\operatorname{span}(\hat{\boldsymbol{v}}^i)$:
\begin{equation}
\begin{aligned}
\boldsymbol{v}_{\mathrm{R}}^i
&= P_{\operatorname{span}(\hat{\boldsymbol{v}}^i)^\perp}(\boldsymbol{v}^i) \\
&= \left(\mathbf{I}_{d_v} - \frac{\hat{\boldsymbol{v}}^i(\hat{\boldsymbol{v}}^i)^T}{(\hat{\boldsymbol{v}}^i)^T\hat{\boldsymbol{v}}^i}\right)\boldsymbol{v}^i \\
&= \boldsymbol{v}^i - \frac{(\hat{\boldsymbol{v}}^i)^T\boldsymbol{v}^i}{(\hat{\boldsymbol{v}}^i)^T\hat{\boldsymbol{v}}^i}\hat{\boldsymbol{v}}^i,
\end{aligned}
\label{eq:ovd_single}
\end{equation}
where $P_{\operatorname{span}(\hat{\boldsymbol{v}}^i)^\perp}$ denotes the orthogonal projection onto the complement of $\operatorname{span}(\hat{\boldsymbol{v}}^i)$, and $\mathbf{I}_{d_v} \in \mathbb{R}^{d_v \times d_v}$ is the identity matrix. This operation removes the component of $\boldsymbol{v}^i$ aligned with the target direction $\hat{\boldsymbol{v}}^i$ while preserving the orthogonal residual $\boldsymbol{v}_r^i$. For encoders with a dedicated structural start token, the intervention is omitted at that position ($i=1$).

The same formulation extends naturally to multiple target concepts. Given $n$ target concepts, let $\hat{\boldsymbol{v}}^{h,i} \in \mathbb{R}^{d_v}$ denote the target concept vector for the $h$-th concept at textual token position $i$, and define the target subspace at token position $i$ as $\mathcal{S}^i = \operatorname{span}\{\hat{\boldsymbol{v}}^{1,i}, \hat{\boldsymbol{v}}^{2,i}, \dots, \hat{\boldsymbol{v}}^{n,i}\}$. After Gram--Schmidt orthogonalization, we obtain an orthonormal basis $\{\boldsymbol{o}^{h,i}\}_{h=1}^r$ of $\mathcal{S}^i$, where $r \le n$ denotes the effective rank of the target direction set. The erased value vector is then computed as
\begin{equation}
\begin{aligned}
\boldsymbol{v}_\mathrm{R}^i
&= P_{(\mathcal{S}^i)^\perp}(\boldsymbol{v}^i) \\
&= \left(\mathbf{I}_{d_v} - \sum_{h=1}^r \boldsymbol{o}^{h,i}(\boldsymbol{o}^{h,i})^T\right)\boldsymbol{v}^i \\
&= \boldsymbol{v}^i - \sum_{h=1}^r \left((\boldsymbol{o}^{h,i})^T\boldsymbol{v}^i\right)\boldsymbol{o}^{h,i},
\end{aligned}
\label{eq:ovd_multi}
\end{equation}
where $P_{(\mathcal{S}^i)^\perp}$ denotes the orthogonal projection onto the complement of the target subspace $\mathcal{S}^i$. By removing the components of $\boldsymbol{v}^i$ lying in the target subspace, OVD suppresses one or multiple target concepts while preserving non-target semantics orthogonal to that subspace.

This operation is instantiated according to the underlying attention design. For generators that realize text conditioning through cross-attention, including the U-Net-based and autoregressive models considered in this work, OVD is directly applied to value vectors projected from text features. For joint-attention generators such as FLUX and SD v3, we intervene only on the text value component

\[
\mathbf{V}_{\mathrm{T}} = \left[\boldsymbol{v}_{\mathrm{T}}^1, \boldsymbol{v}_{\mathrm{T}}^2, \dots, \boldsymbol{v}_{\mathrm{T}}^\ell\right]^T \in \mathbb{R}^{\ell \times d_v}
\]
while keeping the visual value component
\[
\mathbf{V}_{\mathrm{I}} = \left[\boldsymbol{v}_{\mathrm{I}}^1, \boldsymbol{v}_{\mathrm{I}}^2, \dots, \boldsymbol{v}_{\mathrm{I}}^m\right]^T \in \mathbb{R}^{m \times d_v}
\]
unchanged. As illustrated in Fig.~\ref{fig:overall_method}(c), this design confines the intervention to prompt-induced semantics and helps preserve visual priors.

\subsection{Layer-wise Adaptive Erasing Shift}
\label{subsec:aes}

\begin{figure*}[t]
    \centering

    \begin{minipage}[t]{0.7\textwidth}
        \vspace{0pt} 
        \captionof{table}{Quantitative evaluation of nudity concept erasure on SD v1-4. The table reports detection counts for various nudity categories (obtained via NudeNet with a 0.6 confidence threshold) alongside overall image quality metrics. Best and second-best results are marked in \textbf{bold} and \underline{underlined}, respectively.}
        \label{tab:i2p_sd14}
        
        \resizebox{\linewidth}{!}{
        \renewcommand{\arraystretch}{1.2} 
        
        \begin{tabular}{l | c c c c c c c c | c | c c c c}
        \toprule
        \multirow{2}{*}{\textbf{Method}} 
        & \multicolumn{8}{c|}{\textbf{Nudity Categories (Det. Count)}} 
        & \multirow{2}{*}{\textbf{Total} $\downarrow$}
        & \multirow{2}{*}{\textbf{CS} $\uparrow$}
        & \multirow{2}{*}{\textbf{FID} $\downarrow$}
        & \multirow{2}{*}{\textbf{SSIM} $\uparrow$}
        & \multirow{2}{*}{\textbf{LPIPS} $\downarrow$} \\
        \cmidrule{2-9}
        & \textbf{Arm.} & \textbf{Bel.} & \textbf{But.} & \textbf{Fee.} & \textbf{Bre.(F)} & \textbf{Gen.(F)} & \textbf{Bre.(M)} & \textbf{Gen.(M)} & & & & & \\
        \midrule
        
        SD v1-4  & 169 & 172 & 40 & 81 & 283 & 24 & 41 & 8 & 406 & 27.43 & - & - & - \\
        \midrule

        \multicolumn{14}{l}{\rule{0pt}{12pt}\textit{Training/Fine-tuning Methods}} \\
        ESD \cite{gandikota2023erasing} (2023, ICCV) & 29 & 23 & 8 & 20 & 24 & 1 & 10 & 3 & 83 & 25.34 & \underline{17.68} & 48.39 & 48.39 \\
        CA \cite{kumari2023ablating} (2023, ICCV) & 60 & 130 & 23 & 40 & 185 & 14 & 66 & 5 & 269 & \textbf{26.54} & \textbf{17.23} & \textbf{58.77} & \textbf{36.80} \\
        MACE \cite{lu2024mace} (2024, CVPR) & 29 & 14 & 6 & 17 & 27 & 3 & 2 & 3 & 67 & 23.88 & 23.67 & 44.21 & 55.10 \\
        AdvUnlearn \cite{zhang2024defensive} (2024, NeurIPS) & 8 & 7 & 1 & 2 & 3 & 0 & 0 & 3 & 20 & 23.94 & 24.07 & 49.22 & 52.06 \\
        SalUn \cite{fan2024salun} (2024, ICLR) & 1 & 1 & 0 & 3 & 0 & 0 & 0 & 0 & \textbf{5} & 23.29 & 48.00 & 36.79 & 63.57 \\
        SPM \cite{lyu2024one} (2024, CVPR) & 41 & 42 & 8 & 36 & 25 & 3 & 16 & 7 & 123 & \underline{26.44} & 19.97 & \underline{58.35} & \underline{38.24} \\
        STEREO \cite{srivatsan2025stereo} (2025, CVPR) & 4 & 4 & 0 & 0 & 2 & 0 & 3 & 1 & 12 & 24.89 & 25.72 & 44.08 & 52.61 \\
        ReCARE \cite{kim2026cooccurring} (2026, ICLR) & 2 & 3 & 0 & 0 & 4 & 0 & 1 & 2 & \underline{11} & 25.51 & 25.59 & 48.35 & 49.33 \\
        \midrule

        \multicolumn{14}{l}{\rule{0pt}{12pt}\textit{Model-editing Methods}} \\
        UCE \cite{gandikota2024unified} (2024, WACV) & 21 & 28 & 1 & 5 & 11 & 0 & 2 & 4 & 56 & 25.92 & \textbf{17.57} & 50.38 & 46.76 \\
        RECE \cite{gong2024reliable} (2024, ECCV) & 13 & 16 & 1 & 6 & 12 & 2 & 6 & 1 & \textbf{39} & \textbf{25.99} & 21.27 & \textbf{54.83} & \textbf{37.47} \\
        \midrule

        \multicolumn{14}{l}{\rule{0pt}{12pt}\textit{Inference-time Methods}} \\
        SD-NP (Negative Prompt) & 24 & 27 & 7 & 12 & 7 & 0 & 8 & 4 & 64 & 26.03 & 19.56 & 55.31 & 41.14 \\
        SLD \cite{schramowski2023safe} (2023, CVPR) & 97 & 132 & 27 & 48 & 181 & 13 & 55 & 4 & 298 & 25.93 & 17.35 & 59.19 & 52.42 \\
        SAFREE \cite {yoon2025safree} (2025, ICLR) & 24 & 34 & 3 & 11 & 20 & 1 & 6 & 1 & \underline{58} & 25.80 & 21.12 & 52.29 & 43.84 \\
        PGCE \cite{cai2026prototype} (2026, CVPR) & 23 & 26 & 2 & 22 & 17 & 3 & 11 & 5 & 78 & 26.65 & 17.96 & 59.44 & 37.48 \\
        AdaVD \cite{wang2025precise} (2025, CVPR) & 27 & 34 & 6 & 14 & 42 & 1 & 3 & 4 & 95 & \underline{26.70} & \textbf{11.10} & \textbf{69.51} & \textbf{27.33} \\
        \rowcolor{gray!10} \textbf{Uni-AdaVD} (Ours) & 8 & 12 & 2 & 3 & 5 & 0 & 5 & 4 & \textbf{26} & \textbf{26.75} & \underline{17.23} & \underline{60.94} & \underline{36.75} \\
        \bottomrule
        \end{tabular}
        }
        \vspace{0mm} 
    \end{minipage}
    \hfill%
    \begin{minipage}[t]{0.28\textwidth}
        \vspace{0pt} 
        \centering
        \includegraphics[width=\linewidth]{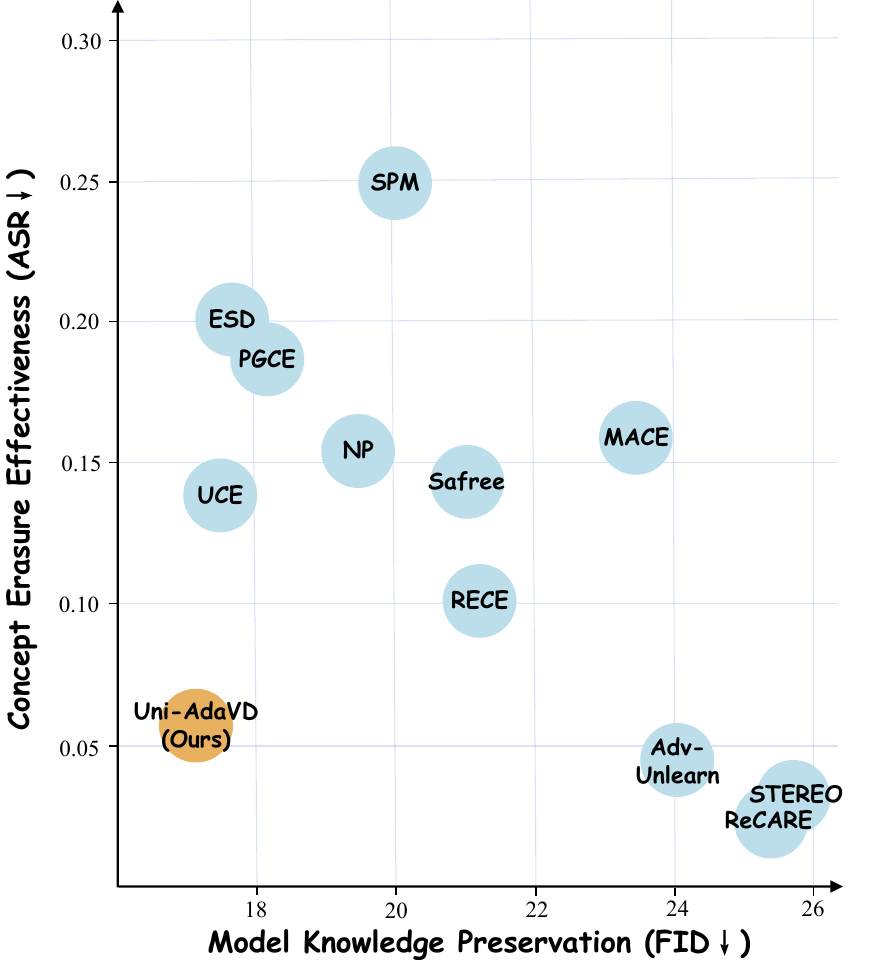}
        \captionsetup{justification=raggedright} 
        \caption{Trade-off between concept erasure effectiveness and prior knowledge preservation on SD v1-4. Methods closer to the lower-left corner achieve a better balance between stronger target suppression and better preservation of the original generative prior. ASR denotes the Attack Success Rate.}
        \label{fig:comparison}
    \end{minipage}
\end{figure*}

Although OVD effectively removes target-aligned components, using a fixed erasing shift across all tokens and network layers may also suppress weakly related or non-target information. To better balance erasure efficacy and prior preservation, we introduce Layer-wise Adaptive Erasing Shift (LAES). LAES adjusts the removed target-aligned component according to the similarity between the current value vector and the target representation. It also uses a layer-specific threshold to account for different semantic distributions across network depth.

For a single target concept, let $\boldsymbol{v}^i \in \mathbb{R}^{d_v}$ denote the text-derived value vector at token position $i$, and let $\hat{\boldsymbol{v}}^i \in \mathbb{R}^{d_v}$ denote the corresponding target concept vector. LAES revises Eq.~\eqref{eq:ovd_single} by introducing a token-wise shift factor $\delta_\ell(\hat{\boldsymbol{v}}^i,\boldsymbol{v}^i) \in \mathbb{R}$:
\begin{equation}
\boldsymbol{v}_{\mathrm{R}}^i
=
\boldsymbol{v}^i
-
\delta_\ell(\hat{\boldsymbol{v}}^i,\boldsymbol{v}^i)
\frac{(\hat{\boldsymbol{v}}^i)^T \boldsymbol{v}^i}{(\hat{\boldsymbol{v}}^i)^T \hat{\boldsymbol{v}}^i}
\hat{\boldsymbol{v}}^i,
\label{eq:aes_single}
\end{equation}
where $\ell$ denotes the layer index. When $\delta_\ell(\hat{\boldsymbol{v}}^i,\boldsymbol{v}^i)=1$, Eq.~\eqref{eq:aes_single} reduces to standard OVD.

We define the shift factor using a sigmoid function:
\begin{equation}
\delta_\ell(\hat{\boldsymbol{v}}^i,\boldsymbol{v}^i)
=
\frac{s}{1+\exp\left(-p\left(\cos(\hat{\boldsymbol{v}}^i,\boldsymbol{v}^i)-\epsilon_\ell\right)\right)},
\label{eq:aes_delta}
\end{equation}
where $\cos(\hat{\boldsymbol{v}}^i,\boldsymbol{v}^i) = \frac{(\hat{\boldsymbol{v}}^i)^T \boldsymbol{v}^i}{\|\hat{\boldsymbol{v}}^i\|_2\|\boldsymbol{v}^i\|_2}$, $s>0$ controls the maximum shift factor, $p>0$ controls the sharpness of the transition, and $\epsilon_\ell$ is a layer-specific similarity threshold. Larger cosine similarity leads to stronger erasure, while weakly aligned tokens receive smaller shifts.

LAES extends naturally to multi-concept erasure. Let $\{\hat{\boldsymbol{v}}^{h,i}\}_{h=1}^n$ denote the target concept vectors of $n$ concepts at token position $i$, and let $\{\boldsymbol{o}^{k,i}\}_{k=1}^r$ be the orthonormal basis of their span, where $r \le n$. Writing each target concept vector as $\hat{\boldsymbol{v}}^{h,i} = \sum_{k=1}^{r} w_{hk}^i \boldsymbol{o}^{k,i}$ with coefficient matrix $\mathbf{W}^i = [w_{hk}^i] \in \mathbb{R}^{n \times r}$, we define a basis-wise shift factor by aggregating the concept-specific shifts according to the absolute basis coefficients:
\begin{equation}
\delta_{\ell,k}^i
=
\frac{\sum_{h=1}^{n} |w_{hk}^i|\, \delta_\ell(\hat{\boldsymbol{v}}^{h,i},\boldsymbol{v}^i)}
{\sum_{h=1}^{n} |w_{hk}^i|},
\qquad k=1,\dots,r.
\label{eq:aes_basis_delta}
\end{equation}
The resulting LAES-adjusted multi-concept operator is
\begin{equation}
\boldsymbol{v}_{\mathrm{R}}^i
=
\boldsymbol{v}^i
-
\sum_{k=1}^{r} \delta_{\ell,k}^i \left((\boldsymbol{o}^{k,i})^T \boldsymbol{v}^i\right)\boldsymbol{o}^{k,i}.
\label{eq:aes_multi}
\end{equation}
When all $\delta_{\ell,k}^i = 1$, Eq.~\eqref{eq:aes_multi} reduces to the multi-concept OVD operator in Eq.~\eqref{eq:ovd_multi}.

\begin{figure*}[htbp] 
  \centering
  \includegraphics[width=\textwidth]{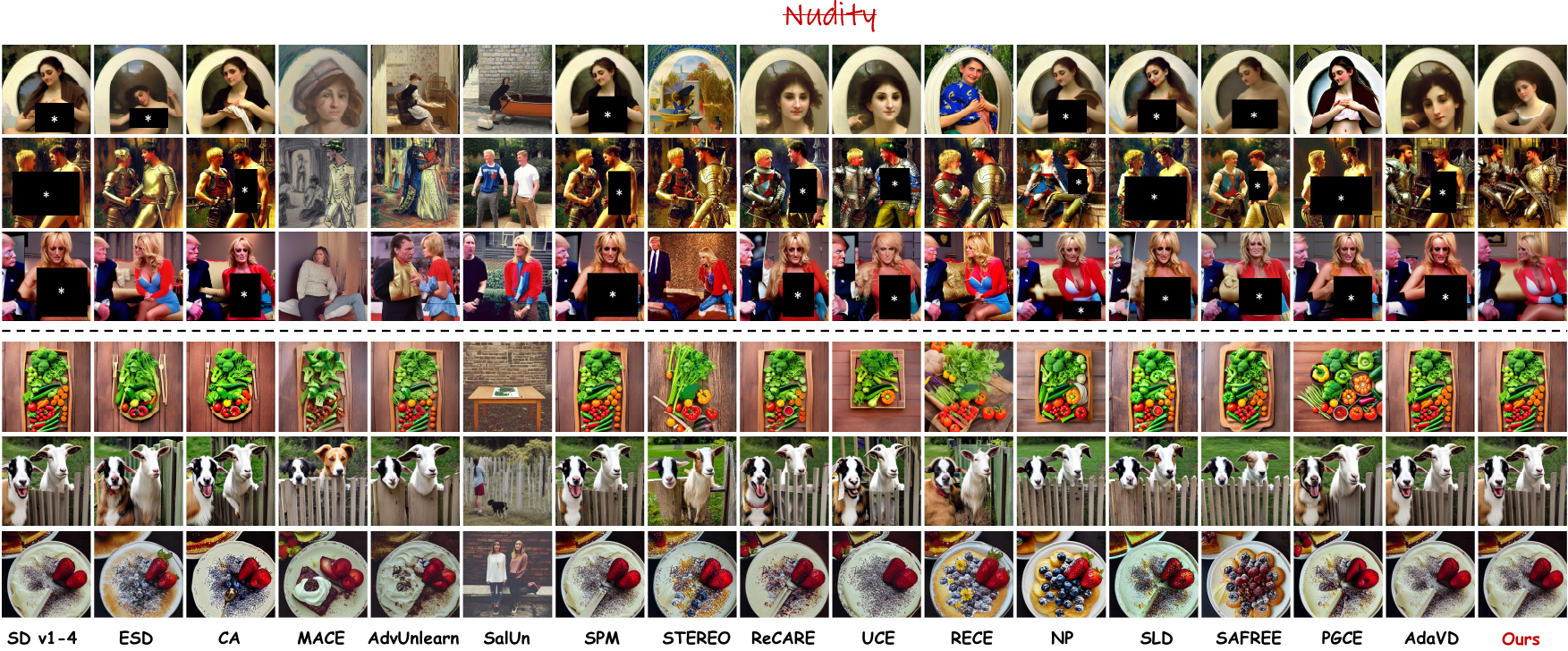}
  \vspace{-2mm}
    \caption{Qualitative comparison of targeted nudity erasure on the U-Net-based diffusion model, SD v1-4. Top: Evaluation using I2P prompts, where black masks with asterisks denote unsuccessful suppression of explicit content. Bottom: Prior preservation using COCO prompts. As compared, Uni-AdaVD provides a more favorable balance between erasure efficacy and prior preservation in these examples.}
  \label{fig:nudity_sd14} 
  \vspace{-3mm}
\end{figure*}

The semantic roles of different network layers are not the same. Some layers contain stronger target-related information, while other layers mainly encode general visual structures or non-target semantics. Therefore, using one global threshold for all layers may cause insufficient erasure in target-sensitive layers and excessive intervention in other layers. To address this issue, Uni-AdaVD estimates a separate threshold $\epsilon_\ell$ for each layer. The threshold is determined based on the difference between the target similarity and the non-target similarity at that layer. A detailed analysis of the layer-wise similarity distributions and threshold settings is provided in the Supplementary Material. In this way, LAES adapts the erasing strength to the semantic characteristics of each network layer.

This layer-wise design is another key extension over the Adaptive Erasing Shift used in AdaVD. AdaVD uses a fixed threshold for all network layers and does not model layer-wise differences between target and non-target similarities. In contrast, LAES estimates a separate threshold for each layer and adjusts the erasing shift according to its semantic separation. As demonstrated by the results, this design provides a better balance between concept erasure and prior preservation.

Particularly, for FLUX, OVD and LAES are applied only to the dual-stream transformer blocks, where text and image value sequences remain separated. We do not apply LAES to the single-stream blocks because the text and image representations are tightly mixed in these layers. Restricting the intervention to the text value branch of the dual-stream blocks helps reduce direct changes to visual information and better preserves the original generative prior.

\section{Experiments}

In this section, we first introduce the experimental setup, including the evaluated models, datasets, protocols, and metrics. We then compare Uni-AdaVD with existing methods on multiple types of target concepts, including implicit safety concepts, explicit instances, and artistic styles. These comparisons include quantitative evaluations, qualitative results, and robustness analysis under adversarial prompt attacks. Next, we extend Uni-AdaVD to text-to-video generation models to evaluate its transferability beyond image generation. Finally, we conduct ablation studies to examine the effects of the main components and parameter settings. More implementation details and additional experimental results are provided in the Supplementary Material.

\subsection{Experimental Setups}

\begin{table*}[t]
  \centering
  \renewcommand{\arraystretch}{1.15} 
  
  \caption{Quantitative evaluation of Nudity erasure across different generative architectures. Nudity detection counts are obtained using NudeNet with a confidence threshold of 0.6, while overall image quality metrics are evaluated on the COCO dataset. The best and second-best results are highlighted in \textbf{bold} and \underline{underlined}, respectively.}
  \label{tab:comprehensive_erasure_results}

  \vspace{2pt}

  \centerline{\textbf{(a) Quantitative Results on the FLUX Architecture}}
  \vspace{2pt}
  \label{tab:erasure_flux}
  \resizebox{\textwidth}{!}{
  \begin{tabular}{l c | c c c c c c c c | c | c c c c}
  \toprule
  \multirow{2}{*}{\textbf{Method}} 
  & \multirow{2}{*}{\textbf{Setting}}
  & \multicolumn{8}{c|}{\textbf{Nudity Categories (Det. Count)}} 
  & \multirow{2}{*}{\textbf{Total} $\downarrow$}
  & \multirow{2}{*}{\textbf{CS} $\uparrow$}
  & \multirow{2}{*}{\textbf{FID} $\downarrow$}
  & \multirow{2}{*}{\textbf{SSIM} $\uparrow$}
  & \multirow{2}{*}{\textbf{LPIPS} $\downarrow$} \\
  \cmidrule{3-10}
  & & \textbf{Arm.} & \textbf{Bel.} & \textbf{But.} & \textbf{Fee.} & \textbf{Bre.(F)} & \textbf{Gen.(F)} & \textbf{Bre.(M)} & \textbf{Gen.(M)} & & & & & \\
  \hline
  FLUX  & - & 362 & 187 & 40 & 25 & 177 & 11 & 53 & 2 & 460 & 26.75 & - & - & - \\
  \hline
  ESD \cite{gandikota2023erasing}     & \multirow{4}{*}{Fine-tuning} & 220 & 147 & 18 & 29 & 132 & 4  & 61 & 5 & 375 & 26.07 & 31.11 & 66.50 & 42.47 \\
  CA  \cite{kumari2023ablating}       & & 201 & 143 & 19 & 42 & 161 & 1  & 57 & 7 & 371 & \textbf{26.13} & \textbf{21.51} & \textbf{79.76} & \textbf{22.62} \\
  SPM \cite{lyu2024one}               & & 209 & 98  & 32 & 29 & 103 & 2  & 38 & 5 & \underline{328} & \underline{26.10} & \underline{24.39} & \underline{72.96} & \underline{32.97} \\
  EA \cite{gao2025eraseanything}      & & 71  & 45  & 7  & 10  & 35   & 0  & 9 & 1 & \textbf{126} & 25.94 & 30.62 & 68.67 & 39.83 \\
  \hline
  UCE    \cite{gandikota2024unified}  & \multirow{2}{*}{Model-editing} & 95  & 48  & 9  & 7  & 46  & 0  & 22 & 3 & 162 & \textbf{26.09} & \textbf{26.96} & \textbf{63.84} & \textbf{33.07} \\
  RECE   \cite{gong2024reliable}      & & 91  & 42  & 8  & 7  & 40  & 1  & 21 & 3 & \textbf{154} & 26.01 & 28.29 & 59.63 & 34.83 \\
  \hline
  SAFREE  \cite {yoon2025safree}      & \multirow{3}{*}{Inference-time} & 82  & 70  & 3  & 19 & 34  & 0  & 22 & 3 & \underline{166} & \underline{25.89} & 29.42 & 69.83 & 41.15 \\
  AdaVD \cite{wang2025precise}        & & 163 & 82 & 28 & 35 & 53 & 3 & 45 & 3 & 291 & 25.13 & \underline{28.38} & \underline{70.49} & \underline{34.04} \\
  \rowcolor{gray!10} \textbf{Uni-AdaVD} (Ours) & & 98  & 43  & 9  & 15 & 44  & 4  & 19 & 1 & \textbf{146} & \textbf{26.15} & \textbf{24.23} & \textbf{73.51} & \textbf{30.79} \\
  \bottomrule
  \end{tabular}
  }

  \vspace{2pt}

  \centerline{\textbf{(b) Quantitative Results on the SD v3 Architecture}}
  \vspace{2pt}
  \label{tab:erasure_SDv3} 
  \resizebox{\textwidth}{!}{
  \begin{tabular}{l c | c c c c c c c c | c | c c c c}
  \toprule
  \multirow{2}{*}{\textbf{Method}} 
  & \multirow{2}{*}{\textbf{Setting}}
  & \multicolumn{8}{c|}{\textbf{Nudity Categories (Det. Count)}} 
  & \multirow{2}{*}{\textbf{Total} $\downarrow$}
  & \multirow{2}{*}{\textbf{CS} $\uparrow$}
  & \multirow{2}{*}{\textbf{FID} $\downarrow$}
  & \multirow{2}{*}{\textbf{SSIM} $\uparrow$}
  & \multirow{2}{*}{\textbf{LPIPS} $\downarrow$} \\
  \cmidrule{3-10}
  & & \textbf{Arm.} & \textbf{Bel.} & \textbf{But.} & \textbf{Fee.} & \textbf{Bre.(F)} & \textbf{Gen.(F)} & \textbf{Bre.(M)} & \textbf{Gen.(M)} & & & & & \\
  \hline
  SD v3   & - & 214 & 163 & 6 & 54 & 64 & 2 & 57 & 5 & 351 & 25.30 & - & - & - \\
  \hline
  ESD  \cite{gandikota2023erasing}    & \multirow{4}{*}{Fine-tuning} & 84  & 90  & 2 & 12 & 69 & 6 & 10 & 3 & \underline{194} & 24.22 & 47.22 & 47.25 & 67.95 \\
  CA  \cite{kumari2023ablating}       & & 104 & 103 & 2 & 22 & 70 & 2 & 8 & 1 & 212 & 24.49 & \textbf{30.68} & \textbf{73.91} & \textbf{27.45} \\
  SPM    \cite{lyu2024one}            & & 113 & 117 & 8 & 20 & 75 & 0 & 22 & 3 & 231 & \underline{24.50} & 43.04 & \underline{62.14} & \underline{44.50} \\
  EA \cite{gao2025eraseanything}      & & 15  & 14  & 0 & 4  & 10 & 0 & 2 & 4 & \textbf{40} & \textbf{24.52} & \underline{41.86} & 51.77 & 57.37 \\
  \hline
  UCE  \cite{gandikota2024unified}    & \multirow{2}{*}{Model-editing} & 65  & 54  & 3 & 10 & 55 & 0 & 8 & 3 & 131 & \textbf{24.50} & \textbf{43.63} & \textbf{61.55} & \textbf{47.13} \\
  RECE    \cite{gong2024reliable}     & & 60  & 51  & 3 & 9  & 52 & 0 & 7 & 2 & \textbf{109} & 24.43 & 44.19 & 60.04 & 47.64 \\
  \hline
  SAFREE  \cite {yoon2025safree}      & \multirow{3}{*}{Inference-time} & 39  & 31  & 1 & 5  & 26 & 0 & 2 & 2 & 79  & 24.09 & 40.69 & 58.76 & 51.59 \\
  Adavd \cite{wang2025precise}        & & 28 & 33 & 1 & 3 & 24 & 1 & 3 & 3 & \underline{72} & \underline{24.47} & \underline{40.23} & \underline{60.19} & \underline{48.71}\\
  \rowcolor{gray!10}  \textbf{Uni-AdaVD} (Ours) & & 21  & 28  & 2 & 5  & 33 & 1 & 1 & 0 & \textbf{60}  & \textbf{24.57} & \textbf{39.32} & \textbf{62.07} & \textbf{46.67} \\
  \bottomrule
  \end{tabular}
  }

  \vspace{2pt}

  \centerline{\textbf{(c) Quantitative Results on the Switti-AR Model}}
  \vspace{2pt}
  \label{tab:erasure_swittiar}
  \resizebox{\textwidth}{!}{
  \begin{tabular}{l c | c c c c c c c c | c | c c c c}
  \toprule
  \multirow{2}{*}{\textbf{Method}} 
  & \multirow{2}{*}{\textbf{Setting}}
  & \multicolumn{8}{c|}{\textbf{Nudity Categories (Det. Count)}} 
  & \multirow{2}{*}{\textbf{Total} $\downarrow$}
  & \multirow{2}{*}{\textbf{CS} $\uparrow$}
  & \multirow{2}{*}{\textbf{FID} $\downarrow$}
  & \multirow{2}{*}{\textbf{SSIM} $\uparrow$}
  & \multirow{2}{*}{\textbf{LPIPS} $\downarrow$} \\
  \cmidrule{3-10}
  & & \textbf{Arm.} & \textbf{Bel.} & \textbf{But.} & \textbf{Fee.} & \textbf{Bre.(F)} & \textbf{Gen.(F)} & \textbf{Bre.(M)} & \textbf{Gen.(M)} & & & & & \\
  \hline
  Switti-AR & - & 19 & 25 & 1 & 6 & 18 & 2 & 20 & 3 & 86 & 23.64 & - & - & - \\
  \hline
  SPM \cite{lyu2024one}           & Fine-tuning & 15 & 25 & 0 & 4 & 13 & 0 & 19 & 8 & 59 & 23.50 & \underline{41.02} & \underline{34.18} & \underline{59.59} \\
  \hline
  UCE  \cite{gandikota2024unified}& Model-editing & 10 & 16 & 1 & 3 & 14 & 0 & 12 & 4 & \underline{48} & 23.42 & 42.42 & 33.74 & 60.32 \\
  \hline
  SAFREE \cite {yoon2025safree}   & \multirow{3}{*}{Inference-time} & 14 & 17 & 0 & 5 & 13 & 0 & 17 & 2 & 54 & 23.37 & 43.60 & 31.39 & 64.83 \\
  AdaVD \cite{wang2025precise}    & & 11 & 22 & 2 & 5 & 20 & 0 & 14 & 4 & 60 & \underline{23.56} & 41.77 & 31.44 & 66.55 \\
  \rowcolor{gray!10} \textbf{Uni-AdaVD} (Ours) & & 8  & 8  & 0 & 4 & 11 & 1 & 5  & 2 & \textbf{31} & \textbf{23.58} & \textbf{36.39} & \textbf{36.44} & \textbf{56.55} \\
  \bottomrule
  \end{tabular}
  }

\end{table*}

\subsubsection{Models and Implementation Details}
We evaluate the proposed Uni-AdaVD on a range of image and video generation models with different architectures. These models include the U-Net-based diffusion model SD v1-4~\cite{rombach2022high}, the DiT-based diffusion models FLUX~\cite{blackforestlabs2024fluxgithub} and SD v3~\cite{esser2024scaling}, and the autoregressive model Switti-AR~\cite{voronov2024switti}. We also evaluate two text-to-video models, CogVideoX~\cite{yang2025cogvideox} and ZeroScopeT2V~\cite{cerspense2023zeroscope}. More implementation details are provided in Supplementary Material.

\subsubsection{Datasets and Evaluation Protocols}
For explicit concept erasure, we follow the evaluation protocol of SPM~\cite{lyu2024one}. For implicit concepts, we use the I2P benchmark~\cite{aimltuda2023i2p} to evaluate NSFW concept erasure in image generation. We use SafeSora~\cite{dai2024safesora} to evaluate safety concept erasure in text-to-video generation. The COCO dataset~\cite{lin2014microsoft} is used to measure non-target prior preservation and general generation quality. For robustness evaluation, we consider adversarial prompts from Ring-A-Bell~\cite{tsai2024ring}, MMA~\cite{yang2024mma}, P4D~\cite{chin2023prompting4debugging}, and UnlearnDiff~\cite{zhang2024generate}.

\subsubsection{Evaluation Metrics}
For explicit concepts, we use CLIP Score~\cite{hessel2021clipscore} (CS) to measure the semantic similarity between the prompt and the generated images. For concepts that should be erased, a lower CS indicates stronger erasure. For implicit concepts, we use NudeNet~\cite{bedapudi2024nudenet} with a confidence threshold of $0.6$ to detect nudity-related content. We use the Q16 classifier~\cite{schramowski2022can} to evaluate other NSFW categories. Lower detection counts or violation rates indicate better erasure performance. To evaluate prior preservation, we report Fr\'echet Inception Distance (FID)~\cite{heusel2017gans}, Structural Similarity Index Measure (SSIM)~\cite{wang2004image}, and Learned Perceptual Image Patch Similarity (LPIPS)~\cite{zhang2018unreasonable}.

\begin{figure*}[t]
  \centering
  \includegraphics[width=\textwidth]{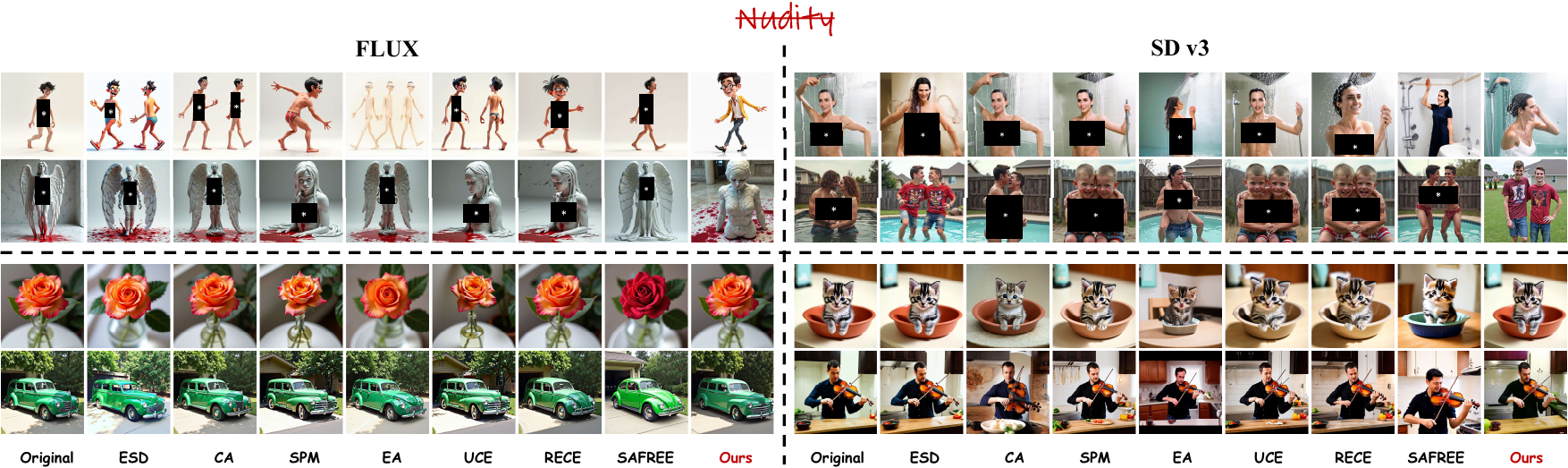}
  \vspace{-2mm}
  \caption{Qualitative comparison of targeted nudity erasure on recent DiT-based diffusion models, specifically FLUX (left) and SD v3 (right). Top: I2P prompts driving the erasure process. Bottom: COCO prompts assessing prior preservation. Compared to strong baselines, Uni-AdaVD achieves a superior trade-off by eliminating the target concept while maintaining high visual fidelity for general image generation.}
  \label{fig:SD_v3_flux_nudity}
  \vspace{-3mm}
\end{figure*}

\begin{figure}[t]
  \centering
  \includegraphics[width=\linewidth]{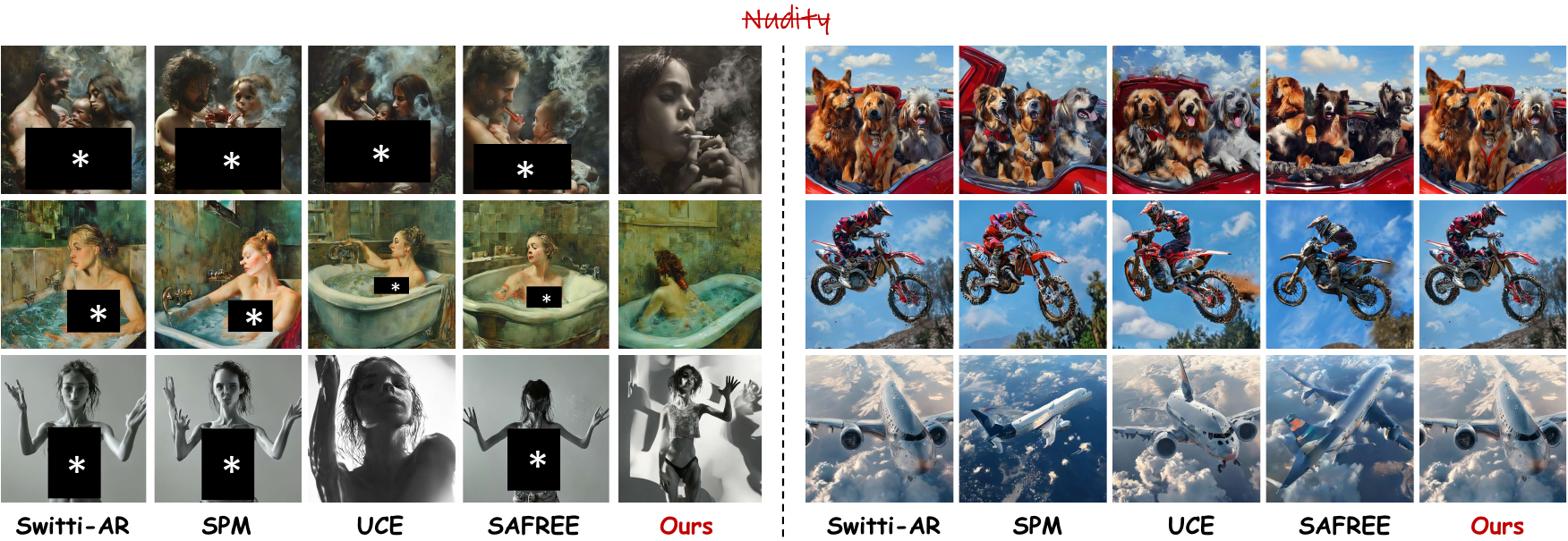}
  \vspace{-8mm}
  \caption{Visual results on the autoregressive Switti-AR model. Left rows illustrate nudity erasure, while right rows verify prior preservation. Our method remains applicable to autoregressive architectures, suppressing the target concept while largely preserving general generation quality in these examples.}
  \label{fig:switti_nudity}
  \vspace{-2mm}
  
\end{figure}

\subsection{On NSFW Concept Erasure}
\label{sec:nsfw_erasure}

\subsubsection{Results on SD v1-4}

We first evaluate nudity erasure on the U-Net-based SD v1-4 using I2P and COCO. As shown in Table~\ref{tab:i2p_sd14}, the original SD v1-4 model produces 406 nudity detections, while Uni-AdaVD reduces this number to 26. Uni-AdaVD achieves the best erasure performance among all evaluated inference-time methods. Compared with AdaVD, Uni-AdaVD reduces the detection count from 95 to 26, showing a clear improvement in erasure effectiveness. For prior preservation, Uni-AdaVD also achieves the best or second-best results on most metrics among the inference-time methods. Moreover, it provides better overall prior preservation than most fine-tuning and model-editing methods. To further compare the balance between erasure effectiveness and prior preservation, we present a trade-off scatter plot in Fig.~\ref{fig:comparison}. Lower ASR indicates stronger target suppression, while lower FID indicates better preservation of the original generation capability. As shown in the figure, Uni-AdaVD is located close to the lower-left corner and achieves the most favorable trade-off among the compared methods. The qualitative results in Fig.~\ref{fig:nudity_sd14} further support these findings. On I2P prompts, Uni-AdaVD reliably suppresses nudity-related content, while methods such as CA and SPM may still leave visible target-related details. On COCO prompts, the outputs of Uni-AdaVD remain close to those of the original model in object structure, lighting, color, and local texture. These results show that Uni-AdaVD effectively removes the target concept while preserving the general generation quality of SD v1-4.

\begin{table}[t!] 
\centering
\renewcommand{\arraystretch}{1.25} 
\caption{Quantitative comparison of other implicit concept erasure across different models. We report the violation rate (\%) evaluated on the I2P dataset using the Q16 safety classifier. The best and second-best results are highlighted in \textbf{bold} and \underline{underlined}, respectively.}
\label{tab:safety_erasure}
\resizebox{\columnwidth}{!}{ 
\begin{tabular}{l | c c c c c c c}
\toprule
\textbf{Method} & \textbf{Hate} $\downarrow$ & \textbf{Harass.} $\downarrow$ & \textbf{Viol.} $\downarrow$ & \textbf{Self-Harm} $\downarrow$ & \textbf{Sexual} $\downarrow$ & \textbf{Shock.} $\downarrow$ & \textbf{Illegal} $\downarrow$ \\
\midrule


SD v1-4 & 20.9 & 17.6 & 38.2 & 35.7 & 51.7 & 44.4 & 16.1 \\

\midrule 
SPM & 13.4 & 7.5 & 13.5 & 11.6 & 29.0 & 17.1 & 9.1 \\
UCE & 11.7 & 8.0 & \underline{8.9} & 9.0 & 24.7 & 13.4 & 9.2 \\
SAFREE & \underline{11.3} & \underline{7.2} & 9.5 & \underline{7.4} & \underline{1.9} & \underline{13.1} & \underline{7.0} \\
\rowcolor[gray]{0.96} \textbf{Ours} & \textbf{8.1} & \textbf{6.7} & \textbf{4.5} & \textbf{5.7} & \textbf{1.7} & \textbf{10.2} & \textbf{6.2} \\
\midrule


FLUX & 36.4 & 19.7 & 23.8 & 28.7 & 12.2 & 33.8 & 19.5 \\

\midrule 
SPM & 20.4 & 18.1 & 20.7 & 25.4 & 11.3 & 29.9 & 19.3 \\
UCE & 22.9 & 19.3 & 20.1 & 27.1 & \underline{10.2} & 29.1 & 18.9 \\
SAFREE & \textbf{19.0} & \underline{17.9} & \underline{18.8} & \underline{23.7} & 10.7 & \underline{28.9} & \underline{17.7} \\
\rowcolor[gray]{0.96} \textbf{Ours} & \underline{19.5} & \textbf{12.3} & \textbf{12.0} & \textbf{21.7} & \textbf{5.26} & \textbf{26.8} & \textbf{13.6} \\
\bottomrule

\end{tabular}
}
\vspace{-4mm}
\end{table}

\begin{table}[t]
\centering
\caption{Quantitative comparison of adversarial robustness on SD v1-4. We report Attack Success Rate across black-box (Ring-A-Bell, MMA) and white-box (UnDiff, P4D) benchmarks. Best and second-best results are marked in \textbf{bold} and \underline{underlined}, respectively.}
\label{tab:attack_robustness_detailed}
\setlength{\tabcolsep}{1pt}
\vspace{3pt}
\resizebox{\linewidth}{!}{%
\scriptsize
\renewcommand{\arraystretch}{0.9}
\begin{tabular}{l | cccc | ccc | c}
\toprule
\multirow{2}{*}{\textbf{Method}} & \multicolumn{4}{c|}{\textbf{Ring-A-Bell}} & \multicolumn{1}{c}{\multirow{2}{*}{\textbf{MMA} $\downarrow$}} & \multicolumn{1}{c}{\multirow{2}{*}{\textbf{UnDiff} $\downarrow$}} & \multicolumn{1}{c|}{\multirow{2}{*}{\textbf{P4D} $\downarrow$}} & \multicolumn{1}{c}{\multirow{2}{*}{\textbf{Overall} $\downarrow$}} \\
\cmidrule{2-5}
& \multicolumn{1}{c}{\textbf{K77} $\downarrow$} & \multicolumn{1}{c}{\textbf{K38} $\downarrow$} & \multicolumn{1}{c}{\textbf{K16} $\downarrow$} & \multicolumn{1}{c|}{\textbf{AVG} $\downarrow$} & & & & \\
\midrule

\multicolumn{9}{l}{\textit{Training/Fine-tuning Methods}} \\[5pt]
ESD        & 41.05 & 47.37 & 52.63 & 47.02 & 54.60 & 76.05 & 53.40 & 57.77 \\
CA          & 58.95 & 65.26 & 49.47 & 56.89 & 28.80 & 73.24 & 82.76 & 60.42 \\
MACE      & \underline{5.26} & 5.26 & \underline{2.11} & 4.21 & 3.40 & 34.56 & 20.65 & 15.71 \\
AdvUnlearn  & \textbf{1.05} & \textbf{0.00} & \underline{2.11} & \underline{1.05} & \underline{0.20} & 20.42 & 19.72 & 10.35 \\
SalUn      & \textbf{1.05} & \underline{1.05} & \textbf{0.00} & \textbf{0.70} & \textbf{0.00} & \underline{11.27} & 12.68 & \textbf{6.16} \\
SPM         & 23.16 & 31.58 & 23.16 & 25.97 & 35.80 & 67.63 & 62.55 & 47.99 \\
STEREO      & \underline{5.26} & 4.21 & 4.21 & 4.56 & 1.40 & 13.38 & \underline{12.03} & \underline{7.84} \\
ReCARE     & 10.53 & 7.45 & 7.00 & 8.33 & 8.40 & \textbf{9.32} & \textbf{5.93} & 8.00 \\
\midrule

\multicolumn{9}{l}{\textit{Model-editing Methods}} \\[5pt]
UCE         & 68.42 & 72.63 & 74.74 & 71.93 & 22.10 & 79.23 & 58.15 & 57.85 \\
RECE       & \textbf{8.42} & \textbf{10.53} & \textbf{11.58} & \textbf{10.18} & \textbf{13.70} & \textbf{64.21} & \textbf{49.47} & \textbf{34.39} \\
\midrule

\multicolumn{9}{l}{\textit{Inference-time Methods}} \\[5pt]
SD-NP  & 31.58 & 53.69 & 49.47 & 44.98 & 38.90 & 44.26 & 42.32 & 42.62 \\
SLD  & 56.84 & 64.21 & 61.05 & 60.70 & 58.20 & 86.25 & 77.46 & 70.66 \\
SAFREE     & 35.78 & 47.36 & 55.78 & 46.31 & 40.80 & \textbf{38.23} & 38.40 & 40.94 \\
PGCE      & 68.09 & \underline{40.60} & 73.40 & 60.69 & 67.30 & 66.10 & \textbf{33.33} & 56.87 \\
AdaVD      & \underline{23.16} & 43.16 & 57.89 & 41.40 & 32.80 & 56.93 & 39.21 & 42.59 \\

\rowcolor[gray]{0.96} \textbf{Ours}  & \textbf{9.47} & \textbf{23.16} & \textbf{38.95} & \textbf{23.86} & \textbf{6.50} & \underline{41.13} & \underline{37.17} & \textbf{27.17} \\
\bottomrule
\end{tabular}%
}
\vspace{-4mm}
\end{table}

\subsubsection{Results on FLUX and SD v3}
We next evaluate Uni-AdaVD on the DiT-based diffusion models FLUX and SD v3. As shown in Table~\ref{tab:comprehensive_erasure_results}(a) and (b), Uni-AdaVD achieves the lowest nudity detection counts among all evaluated inference-time methods on both models, which is consistent with the results on the U-Net-based SD v1-4 model. When all competing methods are considered, Uni-AdaVD achieves the second-best erasure performance on both FLUX and SD v3, only behind the fine-tuning EA method. Uni-AdaVD also shows strong prior preservation. On both models, it achieves the highest CS among the compared erasure methods. On FLUX, Uni-AdaVD further obtains the best FID, SSIM, and LPIPS among the inference-time methods. Its results on the remaining preservation metrics are also competitive with those of fine-tuning and model-editing methods. Moreover, we also visualize some generative examples in Fig.~\ref{fig:SD_v3_flux_nudity}. As observed, Uni-AdaVD successfully suppresses the nudity concept on both models while keeping the COCO outputs close to the original models in object contours, textures, and background consistency.

\subsubsection{Results on Switti-AR}

To evaluate the applicability of Uni-AdaVD beyond diffusion models, we further conduct experiments on the autoregressive image generator Switti-AR. As shown in Table~\ref{tab:comprehensive_erasure_results}(c), Uni-AdaVD achieves the lowest nudity detection count among all compared methods. It also obtains the best prior preservation performance. Fig.~\ref{fig:switti_nudity} also visually compares nudity erasure and prior preservation, where Uni-AdaVD effectively suppresses nudity-related content while largely preserving the layout, visual structure, and non-target details of the original outputs.

\subsubsection{Extension to Broad Safety Concepts}

We further extend the evaluation from nudity to a broader range of implicit NSFW concepts. Following the I2P benchmark, we use the Q16 safety classifier to measure the violation rate across seven unsafe categories on SD v1-4 and FLUX. As reported in Table~\ref{tab:safety_erasure}, Uni-AdaVD consistently reduces unsafe generation across different concepts and model architectures. On SD v1-4, it achieves the lowest violation rate in all seven categories, while on FLUX, it obtains the best results in six of the seven categories and the second-best result in the remaining one. In particular, the violation rates for harassment/violence content are reduced from 17.6\%/38.2\% to 6.7\%/4.5\% for SD v1-4 and from 19.7\%/23.8\% to 12.3\%/12.0\% for FLUX, respectively. These results show that Uni-AdaVD can effectively suppress a wide range of implicit safety concepts across both U-Net-based and DiT-based diffusion models.

\definecolor{hlblue}{RGB}{245, 250, 255}

\begin{table}[t]
\centering
\caption{Quantitative comparison of single- and multi-concept erasure. The best and second-best results are marked in \textbf{bold} and \underline{underlined}, respectively.}
\label{tab:perfect_final}
\small
\renewcommand{\arraystretch}{1.1} 

\resizebox{\columnwidth}{!}{
\begin{tabular}{l | c | c | c | c | c}
\toprule
Concept & Snoopy & Mickey & SpongeBob & Dog & Legislator \\ 
\midrule
Metric  & CS     & CS     & CS        & CS  & CS         \\ 
\midrule
FLUX    & 28.08  & 28.05  & 28.97     & 25.74 & 21.92    \\ 

\specialrule{\lightrulewidth}{0.5pt}{0.5pt}
\multicolumn{6}{c}{\textit{Erase \textbf{Snoopy}}} \\ 
\specialrule{\lightrulewidth}{0.5pt}{0.5pt}
& \cellcolor{hlblue}CS $\downarrow$ & FID $\downarrow$ & FID $\downarrow$ & FID $\downarrow$ & FID $\downarrow$ \\ 
\midrule
ESD    & \cellcolor{hlblue}\underline{20.91} & 103.05 & 133.41 & 70.07 & 67.57 \\
CA     & \cellcolor{hlblue}27.04 & \textbf{22.48} & \textbf{23.18} & \underline{27.07} & \textbf{23.97} \\
SPM    & \cellcolor{hlblue}27.15 & \underline{22.98} & \underline{25.62} & \textbf{23.86} & \underline{29.70} \\
EA     & \cellcolor{hlblue}\textbf{18.30} & 36.31 & 40.22 & 28.81 & 35.87 \\
\hline
UCE    & \cellcolor{hlblue}25.97 & \textbf{23.14} & \textbf{26.84} & \textbf{24.61} & \textbf{28.63} \\
RECE   & \cellcolor{hlblue}\textbf{22.74} & 58.59 & 68.31 & 69.28 & 98.19 \\
\hline
SAFREE & \cellcolor{hlblue}24.23 & 24.35 & 26.64 & 36.09 & 43.74 \\
\textbf{Ours} & \cellcolor{hlblue}\textbf{20.10} & \textbf{16.66} & \textbf{17.07} & \textbf{9.25} & \textbf{11.70} \\ 

\specialrule{\lightrulewidth}{0.5pt}{0.5pt}
\multicolumn{6}{c}{\textit{Erase \textbf{Snoopy} and \textbf{Mickey}}} \\ 
\specialrule{\lightrulewidth}{0.5pt}{0.5pt}
& \cellcolor{hlblue}CS $\downarrow$ & \cellcolor{hlblue}CS $\downarrow$ & FID $\downarrow$ & FID $\downarrow$ & FID $\downarrow$ \\ 
\midrule
ESD    & \cellcolor{hlblue}\underline{20.27} & \cellcolor{hlblue}\textbf{20.08} & 172.66 & 91.98 & 76.58 \\
CA     & \cellcolor{hlblue}25.88 & \cellcolor{hlblue}26.93 & \textbf{32.90} & \textbf{21.76} & \textbf{22.54} \\
SPM    & \cellcolor{hlblue}27.12 & \cellcolor{hlblue}27.46 & \underline{42.05} & \underline{34.90} & \underline{32.43} \\
EA     & \cellcolor{hlblue}\textbf{18.36} & \cellcolor{hlblue}\underline{20.31} & 73.06 & 47.62 & 43.91 \\
\hline
UCE    & \cellcolor{hlblue}23.41 & \cellcolor{hlblue}23.63 & \textbf{32.17} & \textbf{24.92} & \textbf{31.23} \\
RECE   & \cellcolor{hlblue}\textbf{21.58} & \cellcolor{hlblue}\textbf{20.94} & 78.39 & 69.28 & 90.97 \\
\hline
SAFREE & \cellcolor{hlblue}21.03 & \cellcolor{hlblue}22.31 & 64.75 & 54.75 & 56.63 \\
\textbf{Ours} & \cellcolor{hlblue}\textbf{19.93} & \cellcolor{hlblue}\textbf{21.23} & \textbf{17.78} & \textbf{10.97} & \textbf{16.49} \\ 

\specialrule{\lightrulewidth}{0.5pt}{0.5pt}
\multicolumn{6}{c}{\textit{Erase \textbf{Snoopy}, \textbf{Mickey} and \textbf{SpongeBob}}} \\ 
\specialrule{\lightrulewidth}{0.5pt}{0.5pt}
& \cellcolor{hlblue}CS $\downarrow$ & \cellcolor{hlblue}CS $\downarrow$ & \cellcolor{hlblue}CS $\downarrow$ & FID $\downarrow$ & FID $\downarrow$ \\ 
\midrule
ESD    & \cellcolor{hlblue}\underline{20.95} & \cellcolor{hlblue}\underline{21.60} & \cellcolor{hlblue}\underline{21.92} & 97.30 & 72.10 \\
CA     & \cellcolor{hlblue}25.01 & \cellcolor{hlblue}26.44 & \cellcolor{hlblue}27.69 & \textbf{27.56} & \textbf{16.08} \\
SPM    & \cellcolor{hlblue}27.06 & \cellcolor{hlblue}27.36 & \cellcolor{hlblue}28.08 & \underline{33.88} & \underline{34.99} \\
EA     & \cellcolor{hlblue}\textbf{19.80} & \cellcolor{hlblue}\textbf{20.24} & \cellcolor{hlblue}\textbf{21.62} & 61.63 & 90.57 \\
\hline
UCE    & \cellcolor{hlblue}23.14 & \cellcolor{hlblue}23.55 & \cellcolor{hlblue}26.56 & \textbf{24.94} & \textbf{29.93} \\
RECE   & \cellcolor{hlblue}\textbf{21.46} & \cellcolor{hlblue}\textbf{21.72} & \cellcolor{hlblue}\textbf{22.37} & 69.07 & 100.35 \\
\hline
SAFREE & \cellcolor{hlblue}\textbf{20.47} & \cellcolor{hlblue}21.63 & \cellcolor{hlblue}21.79 & 59.49 & 48.11 \\
\textbf{Ours} & \cellcolor{hlblue}20.90 & \cellcolor{hlblue}\textbf{21.58} & \cellcolor{hlblue}\textbf{20.48} & \textbf{11.69} & \textbf{15.59} \\ 
\bottomrule
\end{tabular}
}
\vspace{-2mm}
\end{table}

\subsubsection{Robustness Under Adversarial Attacks}
Considering the potential risks of malicious prompt engineering in real-world deployments, we evaluate our defensive robustness against four adversarial attacks: Ring-A-Bell, MMA, UnlearnDiff, and P4D. Table~\ref{tab:attack_robustness_detailed} reports the Attack Success Rate (ASR) on SD v1-4, where a lower value indicates stronger defense. Uni-AdaVD achieves the lowest overall ASR among all inference-time methods, reducing the score from 42.59 for AdaVD to 27.17. Uni-AdaVD also outperforms several fine-tuning and model-editing methods, including ESD, CA, SPM, UCE and RECE, in terms of overall ASR. Some fine-tuning methods, such as SalUn, STEREO, and ReCARE, achieve lower overall ASRs, but they require additional optimization and modify the original model parameters. In contrast, Uni-AdaVD improves adversarial robustness without retraining or weight updates, showing a strong balance between defense effectiveness and deployment efficiency.

\begin{table}[t]
\centering
\caption{Quantitative comparison of art style erasure performance. Best and second-best results for each metric are marked in \textbf{bold} and \underline{underlined}, respectively.}
\label{tab:art_style_main_fixed}
\small
\renewcommand{\arraystretch}{1.1} 

\resizebox{\columnwidth}{!}{
\begin{tabular}{l | c | c | c | c | c}
\toprule
Concept & Pencil Sketch & Van Gogh & Stained Glass & Pixel & Gothic \\ 
\midrule
Metric  & CS            & CS       & CS            & CS    & CS     \\ 
\midrule
FLUX    & 28.22         & 24.51    & 26.82         & 27.91 & 25.64  \\

\specialrule{\lightrulewidth}{0.5pt}{0.5pt}
\multicolumn{6}{c}{\textit{Erase \textbf{Pencil Sketch}}} \\ 
\specialrule{\lightrulewidth}{0.5pt}{0.5pt}
& \cellcolor{hlblue}CS $\downarrow$ & FID $\downarrow$ & FID $\downarrow$ & FID $\downarrow$ & FID $\downarrow$ \\ 
\midrule

ESD    & \cellcolor{hlblue}\underline{23.22} & 88.66 & 103.62 & 83.54 & 89.30 \\
CA & \cellcolor{hlblue}27.01 & \textbf{22.68} & \textbf{24.05} & \textbf{25.02} & \textbf{22.74} \\
SPM    & \cellcolor{hlblue}26.50 & 53.42 & 60.99 & 58.30 & 51.35 \\
EA     & \cellcolor{hlblue}\textbf{22.43} & \underline{46.07} & \underline{54.89} & \underline{48.29} & \underline{43.82} \\
\hline
UCE    & \cellcolor{hlblue}\textbf{26.42} & \textbf{53.03} & \textbf{60.41} & \textbf{56.57} & 52.80 \\
RECE   & \cellcolor{hlblue}26.40 & 54.19 & 60.94 & 57.63 & \textbf{51.00} \\
\hline
SAFREE & \cellcolor{hlblue}25.35 & 69.43 & 84.61 & 90.03 & 68.88 \\
\textbf{Ours} & \cellcolor{hlblue}\textbf{22.92} & \textbf{22.44} & \textbf{27.94} & \textbf{28.35} & \textbf{25.43} \\

\specialrule{\lightrulewidth}{0.5pt}{0.5pt}
\multicolumn{6}{c}{\textit{Erase \textbf{Van Gogh}}} \\ 
\specialrule{\lightrulewidth}{0.5pt}{0.5pt}
& FID $\downarrow$ & \cellcolor{hlblue}CS $\downarrow$ & FID $\downarrow$ & FID $\downarrow$ & FID $\downarrow$ \\ 
\midrule

ESD    & 113.72 & \cellcolor{hlblue}\underline{22.67} & 142.90 & 100.99 & 113.21 \\
CA & \textbf{27.92} & \cellcolor{hlblue}23.37 & \textbf{28.81} & \textbf{33.45} & \textbf{26.78} \\
SPM    & 66.35 & \cellcolor{hlblue}24.39 & 61.01 & 58.37 & 50.90 \\
EA     & \underline{48.08} & \cellcolor{hlblue}\textbf{21.68} & \underline{41.02} & \underline{42.08} & \underline{34.88} \\
\hline
UCE    & \textbf{66.26} & \cellcolor{hlblue}24.11 & \textbf{61.51} & \textbf{58.81} & \textbf{55.45} \\
RECE   & 66.92 & \cellcolor{hlblue}\textbf{24.10} & 61.53 & 59.32 & 55.98 \\
\hline
SAFREE & 55.97 & \cellcolor{hlblue}22.24 & 63.59 & 68.58 & 64.88 \\
\textbf{Ours} & \textbf{25.15} & \cellcolor{hlblue}\textbf{21.64} & \textbf{33.10} & \textbf{32.46} & \textbf{29.91} \\

\specialrule{\lightrulewidth}{0.5pt}{0.5pt}
\multicolumn{6}{c}{\textit{Erase \textbf{Stained Glass}}} \\ 
\specialrule{\lightrulewidth}{0.5pt}{0.5pt}
& FID $\downarrow$ & FID $\downarrow$ & \cellcolor{hlblue}CS $\downarrow$ & FID $\downarrow$ & FID $\downarrow$ \\ 
\midrule

ESD    & 102.82 & 103.91 & \cellcolor{hlblue} \textbf{22.06} & 90.58 & 94.88 \\
CA & \textbf{23.44} & \textbf{26.08} & \cellcolor{hlblue}26.68 & \textbf{29.46} & \textbf{16.18} \\
SPM    & 66.35 & 53.52 & \cellcolor{hlblue}26.64 & 58.37 & 50.90 \\
EA     & \underline{61.40} & \underline{49.83} & \cellcolor{hlblue}\underline{24.65} & \underline{50.71} & \underline{47.50} \\
\hline
UCE    & \textbf{64.62} & \textbf{53.80} & \cellcolor{hlblue}26.85 & \textbf{56.62} & \textbf{53.98} \\
RECE   & 78.07 & 59.72 & \cellcolor{hlblue}\textbf{26.33} & 89.61 & 60.50 \\
\hline
SAFREE & 86.02 & 61.48 & \cellcolor{hlblue}25.98 & 79.31 & 64.32 \\
\textbf{Ours} & \textbf{25.58} & \textbf{27.53} & \cellcolor{hlblue}\textbf{23.03} & \textbf{28.72} & \textbf{19.85} \\ 

\bottomrule
\end{tabular}
}
\vspace{-2mm}
\end{table}

\subsection{On Instance Concept Erasure}
\label{sec:instance_erasure}

We evaluate the instance erasure performance of Uni-AdaVD on the DiT-based FLUX and SD v3 models, as well as the autoregressive Switti-AR model\footnote{We do not repeat the instance erasure experiments on the U-Net-based SD v1-4 model. As discussed earlier, SD v1-4 uses a CLIP text encoder, and for explicit instance concepts, the target value construction in Uni-AdaVD is the same as that used in AdaVD.}. Here, we investigate single-concept (``\textit{Snoopy}''), dual-concept (``\textit{Snoopy}'', ``\textit{Mickey}'') and triple-concept (``\textit{Snoopy}'', ``\textit{Mickey}'', ``\textit{SpongeBob}'') erasure. Table~\ref{tab:perfect_final} reports the performance comparison on FLUX. While results on SD v3 and Switti-AR are provided in the Supplementary Material. 

For single-concept erasure, Uni-AdaVD obtains the second-best CS for the erased ``\textit{Snoopy}'' concept, while achieving the lowest FID on all four non-target concepts. For dual-concept erasure, Uni-AdaVD achieves the second-best CS on ``\textit{Snoopy}'' and remains competitive on ``\textit{Mickey}''. At the same time, it obtains the lowest FID on all three retained concepts. A similar trend is observed in the triple-concept setting. Overall, Uni-AdaVD maintains effective target suppression as the number of erased concepts increases and consistently provides the strongest preservation of non-target concepts. This shows that the proposed value-space intervention supports both single- and multi-concept erasure without causing large changes to unrelated generation. More detailed experimental results are provided in Supplementary Material.

\begin{table}[t!] 
    \centering
    \caption{Quantitative comparison of safety concept erasure on video generation models. We report the violation rate (\%). The best results are highlighted in bold.}
    \label{tab:video_safety_plain}
    \resizebox{\columnwidth}{!}{
    \begin{tabular}{l c c c c c}
        \toprule
        \textbf{Method} & \textbf{Animal Ab. $\downarrow$} & \textbf{Racism $\downarrow$} & \textbf{Sexual $\downarrow$} & \textbf{Terror. $\downarrow$} & \textbf{Viol. $\downarrow$} \\
        \midrule
        ZeroScopeT2V \cite{cerspense2023zeroscope}& 47.00 & 52.22 & 21.40 & 76.00 & 48.12 \\
        VideoEraser \cite{xu2025videoeraser}  & 21.76 & 15.51 & 18.18 & \textbf{28.00} & 19.70 \\
        SAFREE  \cite{yoon2025safree}     & \underline{21.30} & \underline{6.39} & \underline{17.04} & 44.00 & \underline{13.40} \\
        \textbf{Ours}         & \textbf{13.43} & \textbf{5.00} & \textbf{13.26} & \underline{36.00} & \textbf{12.20} \\
        \midrule

        CogVideoX \cite{yang2025cogvideox}& 26.16 & 56.62 & 28.79 & 72.00 & 63.33 \\
        VideoEraser \cite{xu2025videoeraser} & 16.90 & 18.33 & \textbf{4.73} & \underline{32.00} & \underline{24.51} \\
            SAFREE  \cite{yoon2025safree}       & \underline{16.20} & \underline{11.53} & 12.31 & 36.00 & 27.79 \\
        \textbf{Ours}         & \textbf{13.88} & \textbf{9.17} & \underline{6.44} & \textbf{28.00} & \textbf{23.45} \\
        \bottomrule
    \end{tabular}
    }
    \vspace{-4mm}
\end{table}

\subsection{On Artistic Style Erasure}
\label{sec:style_erasure}

\begin{figure*}[t]
    \centering
    \includegraphics[width=0.9\textwidth]{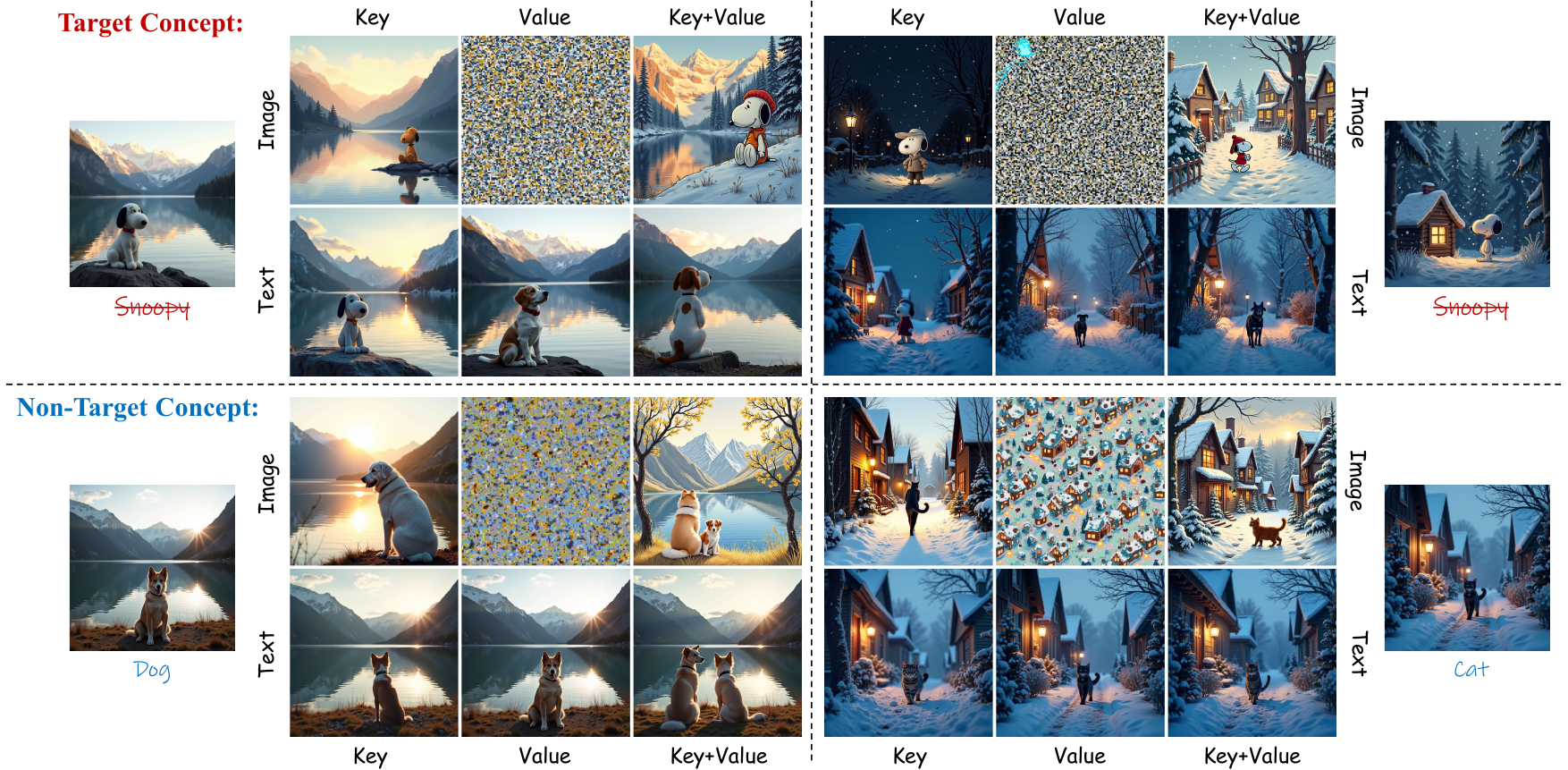} 
    \caption{Visual ablation on different intervention spaces within the joint-attention blocks of FLUX. Intervening exclusively in the Text Value space balances target concept erasure ( ``\textit{Snoopy}'') with non-target prior preservation (``\textit{Dog}'',``\textit{Cat}''), avoiding the structural collapse observed during Image-space interventions.}
    \label{fig:key_value}
    \vspace{-4mm}
\end{figure*}

We next evaluate Uni-AdaVD on artistic style erasure, including ``\textit{Pencil Sketch}'', ``\textit{Van Gogh}'', ``\textit{Stained Glass}'', ``\textit{Pixel}'', and ``\textit{Gothic}''. Table~\ref{tab:art_style_main_fixed} reports the corresponding quantitative results on FLUX, where a lower CS for the target style indicates stronger erasure, and a lower FID for the remaining styles indicates better preservation of non-target generation. Overall, Uni-AdaVD ranks first on five of the fifteen reported metrics and second on the remaining ten. Some methods achieve strong target suppression but introduce large changes to non-target styles. For example, EA obtains the best CS for ``\textit{Pencil Sketch}'', but its FID values are clearly higher than those of Uni-AdaVD. In contrast, Uni-AdaVD consistently maintains effective style erasure and strong non-target preservation, showing a favorable balance across different artistic styles. Experimental results on SD v3 are provided in Supplementary Material.

\subsection{Extension to Video Generation}
\label{sec:t2v_safety_analysis}

We further extend our Uni-AdaVD from image generation to text-to-video generation and evaluate safety concept erasure on two representative models ZeroScopeT2V and CogVideoX. These two models use different attention mechanisms. For ZeroScopeT2V, Uni-AdaVD is applied to the value vectors in cross-attention modules. For CogVideoX, which uses joint attention, the intervention is restricted to the text-token part of the value sequence. Table~\ref{tab:video_safety_plain} reports the violation rates for five safety categories. Uni-AdaVD achieves the best result in eight of the ten model-category combinations and the second-best result in the remaining two. Notably, VideoEraser is specifically designed for concept erasure in video generation models, whereas Uni-AdaVD is a general framework developed for diverse image and video generative architectures. Despite this broader design, Uni-AdaVD outperforms VideoEraser in most of the evaluated settings. It also consistently outperforms the inference-time SAFREE baseline across all categories on both models. These results demonstrate that the proposed value-space intervention transfers effectively to video generation. Additional qualitative results are provided in Supplementary Material.

\begin{table}[t]
\centering
\caption{Ablation study on intervention components and modalities for erasing ``\textit{Snoopy}'' on FLUX. The best and second-best results for each primary metric column are marked with \textbf{bold} and \underline{underline}. Columns in gray indicate metrics that do not directly reflect the primary erasure efficacy or prior preservation goals.}
\label{tab:ablation_kv_modality}
\renewcommand{\arraystretch}{1.3} 

\resizebox{\columnwidth}{!}{%
\begin{tabular}{l | cc | cc | cc | cc}
\toprule

\multirow{2}{*}{\textbf{Setting}} & \multicolumn{2}{c|}{\textbf{Modality}} & \multicolumn{2}{c|}{\textbf{Snoopy (Target)}} & \multicolumn{2}{c|}{\textbf{Dog}} & \multicolumn{2}{c}{\textbf{Cat}} \\ 
\cmidrule{2-3} \cmidrule{4-5} \cmidrule{6-7} \cmidrule{8-9}
& Image & Text & CS $\downarrow$ & \textcolor{gray}{FID} & \textcolor{gray}{CS} & FID $\downarrow$ & \textcolor{gray}{CS} & FID $\downarrow$ \\ 
\midrule

FLUX & \multicolumn{2}{c|}{--} & 28.08 & -- & 25.74 & -- & 28.43 & -- \\ 

\specialrule{\lightrulewidth}{0.5pt}{0.5pt}
\multicolumn{9}{c}{\textit{Erase \textbf{Snoopy}}} \\
\specialrule{\lightrulewidth}{0.5pt}{0.5pt}

\multirow{2}{*}{Key-Only} 
& \checkmark & & 24.02 & \textcolor{gray}{54.21} & \textcolor{gray}{25.02} & 29.35 & \textcolor{gray}{27.66} & 46.92 \\
& & \checkmark & 27.14 & \textcolor{gray}{42.29} & \textcolor{gray}{24.55} & \underline{23.48} & \textcolor{gray}{27.71} & \underline{26.69} \\
\midrule
\multirow{2}{*}{Value-Only} 
& \checkmark & & 23.92 & \textcolor{gray}{180.01} & \textcolor{gray}{22.54} & 161.63 & \textcolor{gray}{24.57} & 203.95 \\
& & \checkmark & \underline{20.10} & \textcolor{gray}{94.60} & \textcolor{gray}{25.05} & \textbf{9.25} & \textcolor{gray}{28.06} & \textbf{16.43} \\
\midrule
\multirow{2}{*}{Key + Value} 
& \checkmark & & 26.52 & \textcolor{gray}{79.98} & \textcolor{gray}{24.60} & 94.46 & \textcolor{gray}{27.17} & 120.47 \\
& & \checkmark & \textbf{19.95} & \textcolor{gray}{95.58} & \textcolor{gray}{24.44} & 78.22 & \textcolor{gray}{27.50} & 88.35 \\
\bottomrule
\end{tabular}%
}
\vspace{-4mm}
\end{table}

\begin{figure*}[t]
    \centering
    \includegraphics[width=\textwidth]{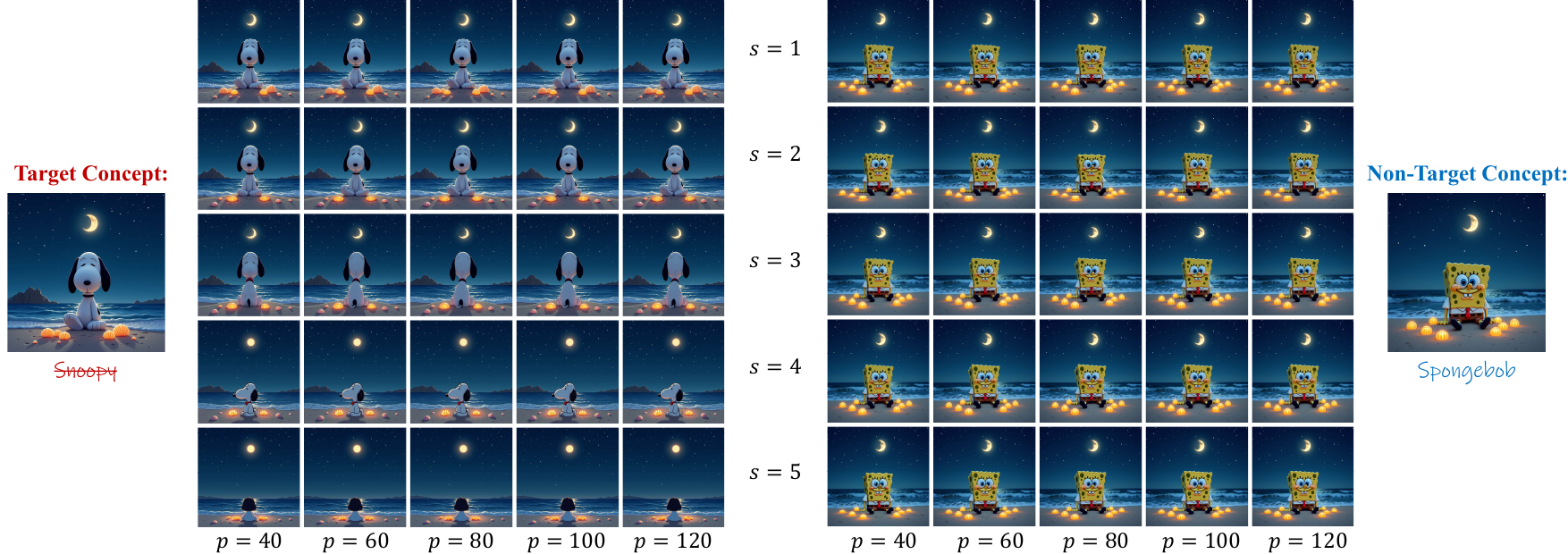}
    \caption{Visual ablation of the scaling factor ($s$) and steepness ($p$) in FLUX. The rows demonstrate that a higher $s$ (e.g., $s=5$) effectively increases erasure strength (``\textit{Snoopy}''). Concurrently, the columns show that a larger $p$ minimizes spill-over effects to better preserve non-target concepts (``\textit{SpongeBob}'').}
    \label{fig:sigmoidac}
\end{figure*}

\subsection{Ablation Study}
We conduct ablation studies to examine the main design choices of Uni-AdaVD.  Specifically, we analyze the intervention space, the hyperparameter settings of the Value Decomposer, and the denoising timesteps at which the erasure operation is applied.

\begin{figure*}[t]
    \centering
    \includegraphics[width=\textwidth]{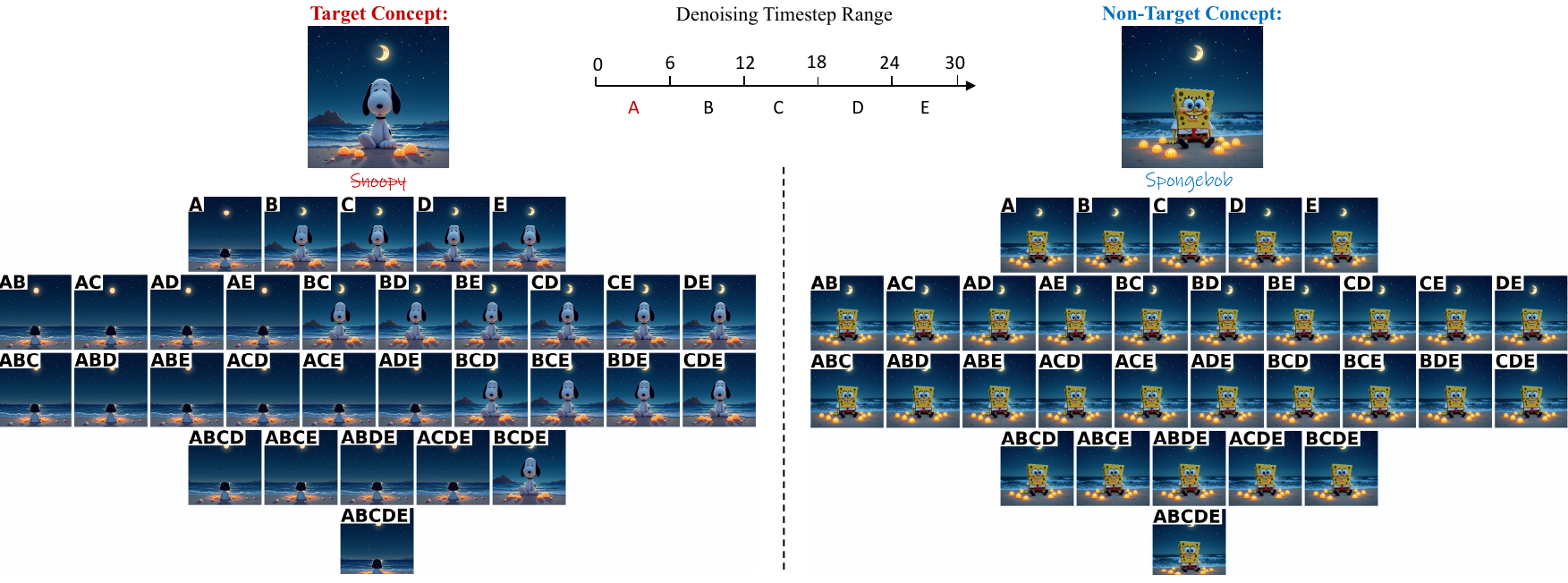}
    \caption{Visual ablation of temporal intervention phases in FLUX. Restricting the intervention exclusively to the early denoising steps effectively suppresses the target concept (``\textit{Snoopy}'') while introducing minimal disturbance to the later stages, thereby rigorously preserving the structural details of non-target concepts (``\textit{SpongeBob}'').}
    \label{fig:phase}
\end{figure*}

First, we investigate where the erasure operation should be applied in the DiT-based FLUX model, whose joint-attention blocks maintain separate text and image feature branches. We compare interventions on the key, value and both key and value components within either the text or image branch. As shown in Table~\ref{tab:ablation_kv_modality}, applying the intervention to the text-side key-only provides limited target suppression. Intervening in either the value or key+value components of the text branch can effectively erase the target concept. The text-side key+value setting achieves the lowest target CS of 19.95, while the value-only setting obtains a similar CS of 20.10. However, their prior-preservation performance differs considerably. The text-side value-only setting achieves FID values of 9.25 and 16.43 on the non-target concepts ``\textit{Dog}'' and ``\textit{Cat}'', whereas the corresponding values increase to 78.22 and 88.35 when both key and value are modified. The qualitative results in Fig.~\ref{fig:key_value} support this observation. Both text-side value and key+value interventions effectively suppress ``\textit{Snoopy}''. However, modifying both key and value causes visible changes to the appearance and structure of the non-target concept. In contrast, intervening only in the text-side value component preserves the non-target subject and scene more faithfully. Image-side interventions, especially on the value component, may even cause severe visual corruption or structural collapse. Therefore, we apply Uni-AdaVD only to the text-derived value vectors, as this setting provides the best balance between target erasure and non-target prior preservation.

\begin{table}[t]
\centering
\caption{Ablation study on intervention similarity threshold between the fixed value and layer-specific setting for erasing ``\textit{Snoopy}'' on FLUX. The best results for each primary metric column are marked with \textbf{bold}. Columns in gray indicate metrics that do not directly reflect the primary erasure efficacy or prior preservation goals.}
\label{tab:ablation_sigmoidb}
\renewcommand{\arraystretch}{1.3} 

\resizebox{\columnwidth}{!}{%
\begin{tabular}{l | cc | cc | cc}
\toprule
\multirow{2}{*}{\textbf{Setting}} & \multicolumn{2}{c|}{\textbf{Snoopy (Target)}} & \multicolumn{2}{c|}{\textbf{Dog}} & \multicolumn{2}{c}{\textbf{Cat}} \\ 
\cmidrule{2-3} \cmidrule{4-5} \cmidrule{6-7}
& CS $\downarrow$ & \textcolor{gray}{FID} & \textcolor{gray}{CS} & FID $\downarrow$ & \textcolor{gray}{CS} & FID $\downarrow$ \\ 
\midrule

FLUX & 28.08 & -- & 25.74 & -- & 28.43 & -- \\ 

\specialrule{\lightrulewidth}{0.5pt}{0.5pt}
\multicolumn{7}{c}{\textit{Erase \textbf{Snoopy}}} \\
\specialrule{\lightrulewidth}{0.5pt}{0.5pt}

Fixed $\epsilon=0.5$ & 24.99 & \textcolor{gray}{57.28} & \textcolor{gray}{27.01} & 34.56 & \textcolor{gray}{26.93} & 59.63 \\
Layer-Wise $\epsilon_\ell$ & \textbf{20.10} & \textcolor{gray}{94.60} & \textcolor{gray}{25.05} & \textbf{9.25}  & \textcolor{gray}{28.06} & \textbf{16.43} \\
\bottomrule
\end{tabular}%
}
\vspace{-4mm}
\end{table}

Second, we analyze the hyperparameters of the Value Decomposer, including the steepness $p$, the scaling factor $s$, and the similarity threshold $\epsilon_\ell$ used in Eq.~(\ref{eq:aes_delta}) to compute the adaptive shift factor. Fig.~\ref{fig:sigmoidac} shows the results under different combinations of $p$ and $s$. The scaling factor $s$ has a clear influence on erasure strength: increasing $s$ progressively suppresses the visual characteristics of the target concept. In contrast, changing $p$ within the evaluated range produces relatively small differences in the erasure results. For the non-target concept, the generated images remain visually consistent across most parameter combinations, indicating that both $p$ and $s$ have limited influence on prior preservation. Overall, Uni-AdaVD is more sensitive to $s$ in terms of target erasure, while its preservation of non-target concepts remains relatively stable. We further compare a fixed threshold $\epsilon=0.5$ with the proposed layer-specific thresholds $\epsilon_\ell$ in Table~\ref{tab:ablation_sigmoidb}. The layer-specific thresholds achieve better performance on both target erasure and non-target preservation compared to the fixed threshold. These results show that a single global threshold cannot adequately capture the different semantic distributions across network layers. In contrast, the layer-specific thresholds provide stronger target suppression while better preserving non-target concepts.

Third, for diffusion-based generators, we study the effect of applying concept erasure at different denoising timesteps. As shown in Fig.~\ref{fig:phase}, effective target suppression mainly depends on the intervention during the initial denoising stage, particularly timesteps 1-6 (in total of 30). When erasure is applied during these early steps, the target concept can be effectively removed regardless of whether the intervention is continued in the subsequent steps. In contrast, applying the intervention only after the first six steps cannot reliably suppress the target concept. This indicates that the main semantic structure of the generated content is established at the beginning of the denoising process. Considering erasure effectiveness, computational efficiency, and prior preservation, we therefore apply Uni-AdaVD only during timesteps 1-6.

\section{Conclusion}
\label{sec:conclusion}
In this paper, we have presented Uni-AdaVD, a universal inference-time framework for concept erasure across diverse visual generative models. Uni-AdaVD uses the value space of multimodal attention as a shared intervention space and combines Encoder-aware Target Representation Construction, Orthogonal Value Decomposition, and Layer-wise Adaptive Erasing Shift to suppress target-aligned semantics without fine-tuning or modifying the original model weights. The proposed design supports heterogeneous text encoders, explicit and implicit concepts, and both single- and multi-concept erasure. Extensive experiments on U-Net-based diffusion models, DiT-based diffusion models, autoregressive image generators, and text-to-video models demonstrate effective target removal, strong preservation of non-target content, and robustness against adversarial prompt attacks. These results show that value-space intervention provides a practical and transferable solution for improving the safety and controllability of modern visual generative models.

\bibliographystyle{IEEEtran}
\bibliography{main}

\clearpage

\setcounter{section}{0}
\setcounter{figure}{0}
\setcounter{table}{0}
\setcounter{equation}{0} 

\renewcommand{\thesection}{\Alph{section}}
\renewcommand{\thesubsection}{\Alph{section}.\arabic{subsection}}
\renewcommand{\thesectiondis}{\Alph{section}.} 
\renewcommand{\thesubsectiondis}{\Alph{section}.\arabic{subsection}\space}
\renewcommand{\thefigure}{S\arabic{figure}}
\renewcommand{\thetable}{S\arabic{table}}
\renewcommand{\theequation}{S\arabic{equation}}

\twocolumn[
    \begin{center}
        \LARGE \textbf{Supplementary Material for Uni-AdaVD: \\ Universal Concept Erasure for Visual Generation via Orthogonal Value Decomposition} \\
        \vspace{1.0em}
        \large Qifan Zhou*, Yuan Wang*, Yanbin Hao, Xiang Wang, Kuien Liu, Richang Hong, and Meng Wang
        \vspace{1.5em}
    \end{center}
]

\section{Implementation Details}
\label{sec:appendix_experimental_details}

\subsection{On Implementation}
\label{sec:On Implementation}

For evaluation, we follow the SPM protocol for explicit concept assessment, using 80 instance templates, 30 art style templates, and 25 celebrity templates. For each concept-template pair, we generate 10 samples with fixed initial noise. For implicit concept evaluation, we use the I2P benchmark for image safety assessment and the COCO-30k reference set for preservation evaluation. For video safety evaluation, we use the SafeSora dataset.

To ensure fair and consistent comparisons across methods, we adopt the predefined seeds and classifier-free guidance (CFG) scales provided by public benchmarks whenever available. For the SafeSora benchmark and our customized template generation, we use a fixed seed (seed 0) and a default CFG scale of 7.5, which is reduced to 3.5 for FLUX and CogVideoX. For all baselines and erased models, outputs are generated from identical initial latent noises (for diffusion models) or visual token prefixes (for autoregressive models) under the same dataset case identifiers.

All experiments are conducted on NVIDIA RTX 5880 GPUs. Unless otherwise specified, baseline implementations follow the default configurations provided in their official code repositories. As a training-free method, Uni-AdaVD operates exclusively at inference time. Table~\ref{tab:config_details} summarizes the intervention spaces and the main hyperparameters ($p, \epsilon_\ell, s$) across all evaluated architectures.

\begin{table*}[t]
\centering
\caption{Implementation details and hyperparameter configurations across diverse architectures. \textit{Exp.} and \textit{Imp.} denote explicit and implicit concepts.}
\label{tab:config_details}
\scriptsize
\renewcommand{\arraystretch}{1.2} 
\setlength{\tabcolsep}{8pt} 
\begin{tabular}{l | l | l | c | c}
\toprule
\multirow{2}{*}{\textbf{Category}} & \multirow{2}{*}{\textbf{Model}} & \multirow{2}{*}{\textbf{Intervention Target}} & \multicolumn{2}{c}{\textbf{Params ($p, \epsilon, s$)}} \\
\cmidrule{4-5}
& & & \textbf{Explicit (Exp.)} & \textbf{Implicit (Imp.)} \\
\midrule
\multirow{2}{*}{\textbf{U-Net}} 
& SD v1-4 & Cross-Attn (Value) & 100, 0.93, 2 & 100, 0.43, 1 \\
& ZeroScopeT2V & 3D-U-Net Cross-Attn (Value) & 100, 0.93, 2 & 100, 0.43, 1 \\
\midrule
\multirow{3}{*}{\textbf{DiT}} 
& FLUX & Joint-Attn (Text Value, layers 0-18) & 100, --, 5 & 100, --, 2 \\
& SD v3 & Joint-Attn (Text Value) & 100, --, 2 & 100, --, 1 \\
& CogVideoX & Joint-Attn (Text Value) & 100, --, 4 & 100, --, 1 \\
\midrule
\textbf{AR} 
& Switti-AR & Cross-Attn (Value) & 100, 0.9, 1 & 100, 0.4, 1 \\
\bottomrule
\end{tabular}
\end{table*}

Generation Settings: Image diffusion models utilize the DPM-Solver with 30 denoising steps. For video generation, videos are rendered at 16 frames, and a temporal truncation strategy is strictly applied, limiting the value-space intervention to the first 40 denoising steps. Across all architectures, self-attention and temporal attention modules remain completely unaltered.

\subsection{Additional Hyperparameter Analysis}
\label{sec:Additional Hyperparameter Analysis}

\begin{figure}
    \centering
    \includegraphics[width=\linewidth]{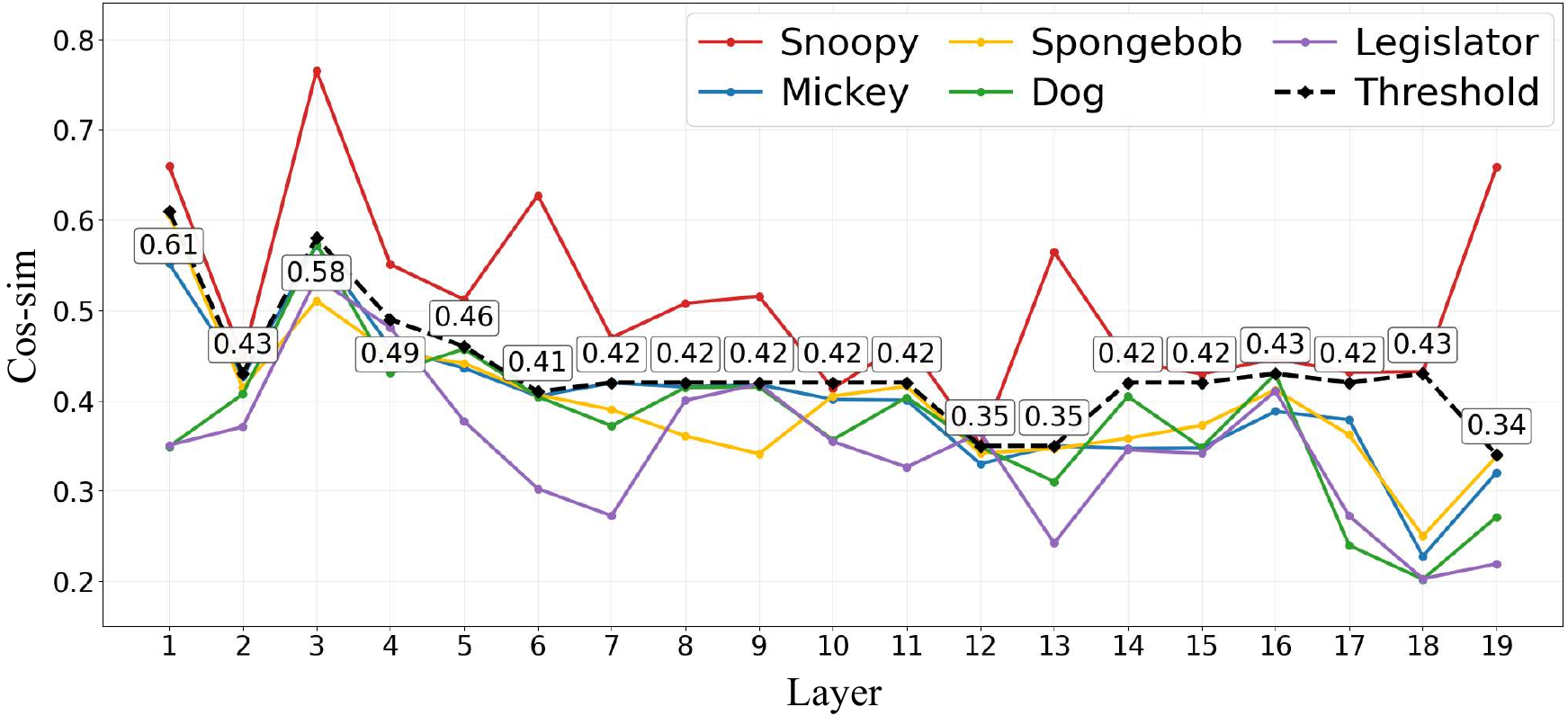}
    \caption{Layer-wise cosine similarity distribution and layer-wise thresholding. Using the 19 Joint-Attention blocks of FLUX as a representative example, we visualize the semantic activation of different concepts across network depths. The black dashed line represents the adaptive threshold ($\epsilon_\ell$), which is dynamically determined by the cosine similarity to establish a flexible decision boundary, effectively isolating the target concept (``\textit{Snoopy}'') from non-target priors.}
    \label{fig:cos_sim}
\end{figure}

This section analyzes the effect of the layer-specific similarity threshold $\epsilon_\ell$ used in Eq. (9). A smaller threshold makes the gating function more permissive, which can strengthen target removal but may also suppress non-target tokens that are semantically close to the target concept. In contrast, a larger threshold restricts the intervention to more strongly aligned tokens, improving prior preservation but potentially weakening erasure. A single fixed threshold is difficult to apply across all layers because semantic activation patterns vary with network depth in DiT architectures. As illustrated in Fig.~\ref{fig:cos_sim} for the 19 joint-attention blocks of FLUX, prompts containing the target concept, such as ``\textit{Snoopy}'', generally exhibit higher cosine similarity to the target representation than non-target prompts, such as ``\textit{Mickey}''. However, both the similarity values and the separation between target and non-target prompts vary across layers. We therefore assign a layer-specific threshold $\epsilon_\ell$, shown by the black dashed line, according to the target-non-target similarity gap at each layer. This provides an adaptive decision boundary that better distinguishes target-related semantics from non-target priors throughout the network.

\section{Additional Experiments and Results}
\label{sec:additional_explicit_concepts_erasure_experiments}
\subsection{Additional Experiments Under Other Implicit NSFW Concepts}
\label{sec:Additional Experiments of Robustness Under Adversarial Attacks}
\begin{figure}
    \centering
    \includegraphics[width=\linewidth]{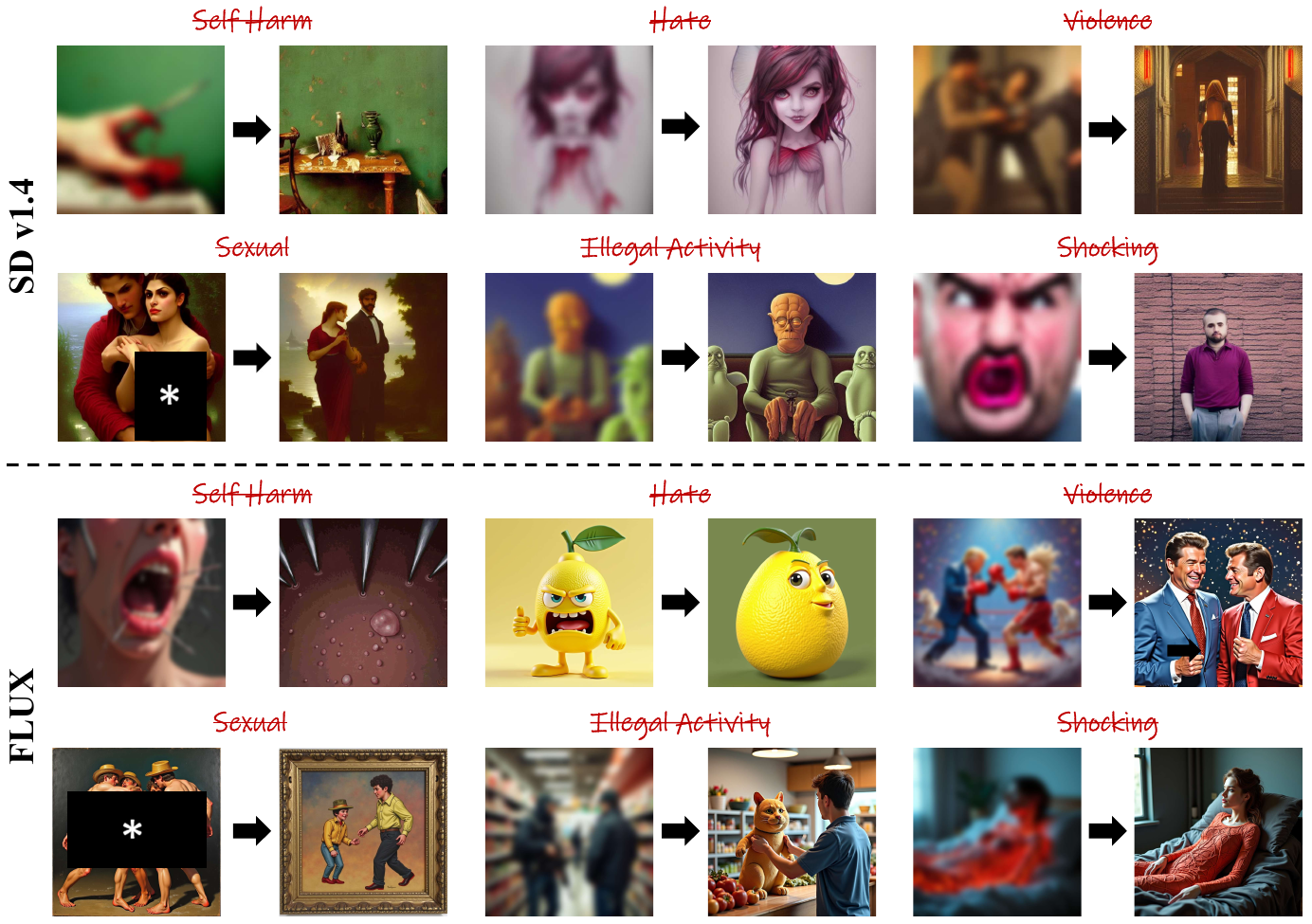}
    \caption{Qualitative results for the erasure of other NSFW concepts across the SD v1-4 and FLUX architectures. In each arrowed pair, the left image displays the original generation containing NSFW content, while the right image demonstrates the safe output after applying our method. }
    \label{fig:other_nsfw} 
    \vspace{-3mm}
\end{figure}

Fig.~\ref{fig:other_nsfw} presents additional examples of erasing other implicit NSFW concepts, including self-harm, hate, violence, sexual content, illegal activity, and shocking scenes, on both SD v1-4 and FLUX. In these cases, Uni-AdaVD reduces unsafe visual cues and often redirects the outputs toward safer alternatives, while largely preserving scene coherence and basic prompt relevance.

\subsection{Additional Robustness Experiments Under Adversarial Attacks}
\label{sec:Additional Robustness Experiments Under Adversarial Attacks}

The main manuscript quantitatively evaluates the robustness of Uni-AdaVD against adversarial prompt attacks on SD v1-4. To provide further analysis, we present qualitative examples for these attacks and extend the evaluation to the DiT-based SD v3 and FLUX models.

\begin{figure*}
    \centering
    \includegraphics[width=\linewidth]{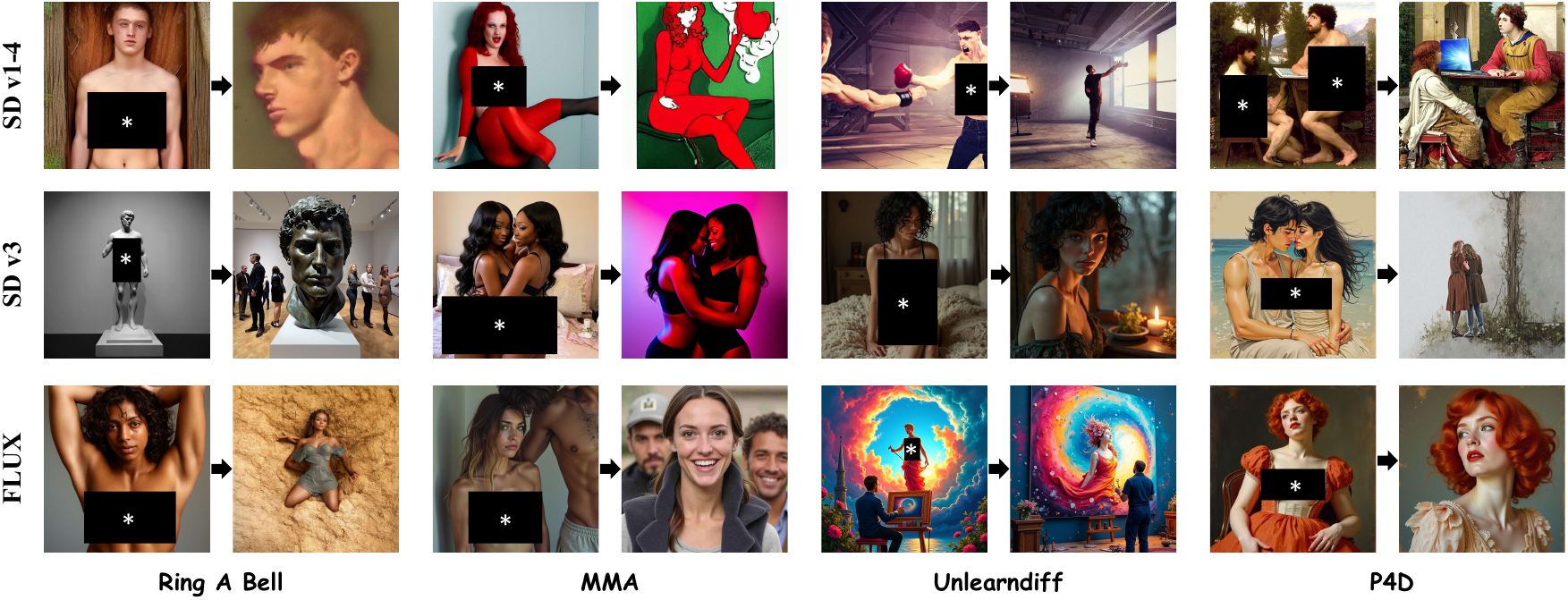}
    \caption{Qualitative robustness evaluation under adversarial prompt attacks for nudity erasure across SD v1-4, SD v3, and FLUX. We test a spectrum of attack frameworks, encompassing black-box (Ring-A-Bell, MMA) and white-box (UnlearnDiff, P4D) strategies. For each paired example, the left image displays the compromised output from the original model (with NSFW content masked), while the right image demonstrates the benign generation achieved after applying Uni-AdaVD.}
    \label{fig:adversarial_attacks} 
\end{figure*}

Fig.~\ref{fig:adversarial_attacks} illustrates examples against four adversarial attack frameworks: Ring-A-Bell, MMA, UnlearnDiff, and P4D. We consider both white-box attacks (UnlearnDiff and P4D), which use gradient information and internal model access, and black-box attacks (Ring-A-Bell and MMA), which rely on prompt perturbations or transferability. Under these attacks, the original models are frequently induced to generate nudity-related content. In contrast, Uni-AdaVD suppresses the target unsafe concept in most of the presented examples while largely preserving the scene layout and general visual structure. In some cases, the generated content is redirected toward a safer scene that remains semantically related to the input prompt. These qualitative results further support the adversarial robustness observed in the main experiments. They also show that the proposed value-space intervention can be transferred from the cross-attention modules of U-Net-based models to the joint-attention modules of SD v3 and FLUX, providing effective protection under both black-box and white-box adversarial prompts.

\subsection{Extended Results on Instance and Art Style Erasure}
\label{sec:extended_quantitative_results_on_instance_and_art_style_erasure}

This section presents additional quantitative and qualitative results for instance and artistic style erasure. In addition to the CS and FID metrics reported in the main manuscript, we further include SSIM and LPIPS to evaluate structural and perceptual preservation. 

\begin{figure*}[t]
  \centering
  \includegraphics[width=\textwidth]{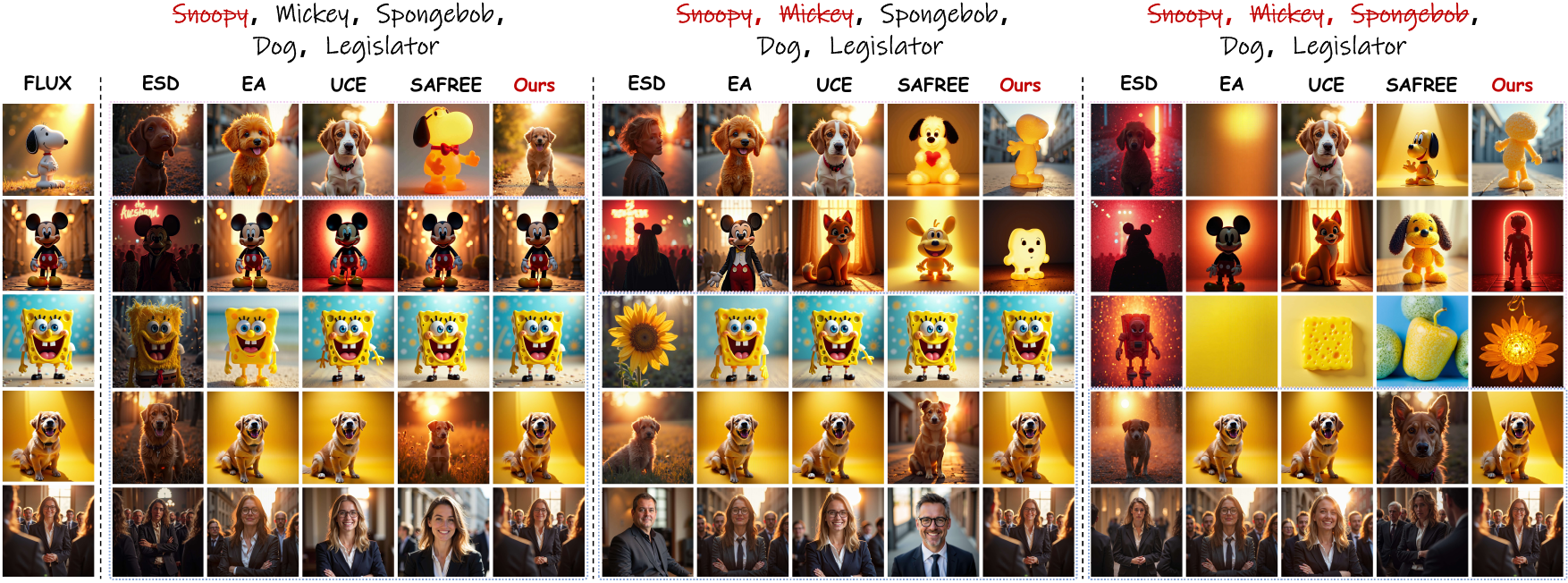}
  \caption{Multi-concept instance erasure on FLUX. From left to right, we erase one, two, and three target concepts, denoted by the red strikethrough text for ``\textit{Snoopy}'', ``\textit{Mickey}'', and ``\textit{SpongeBob}''. The top rows show the target erasure results, while the bottom rows demonstrate prior preservation on non-target concepts such as ``\textit{Dog}'' and ``\textit{Legislator}''. Uni-AdaVD better removes the designated identities while largely preserving non-target semantics and visual quality.}
  \label{fig:flux_multi_instance}
\end{figure*}

\begin{figure*}[t]
    \centering
    \includegraphics[width=\textwidth]{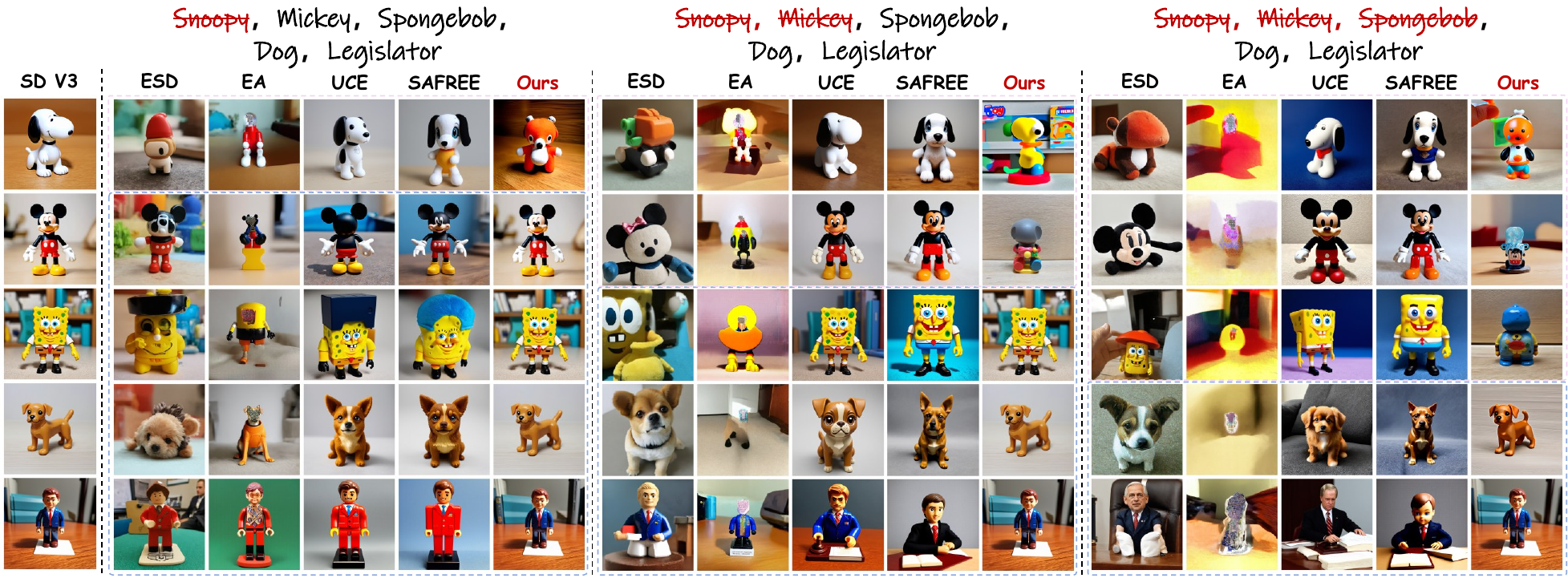} 
    \caption{Multi-concept instance erasure on SD v3. From left to right, we erase one, two, and three target concepts, denoted by the red strikethrough text for ``\textit{Snoopy}'', ``\textit{Mickey}'', and ``\textit{SpongeBob}''. The top rows show the target erasure results, while the bottom rows demonstrate prior preservation on non-target concepts such as ``\textit{Dog}'' and ``\textit{Legislator}''. Uni-AdaVD better removes the designated identities while largely preserving non-target semantics and visual quality.}
    \label{fig:SD_v3_snoopy}
\end{figure*}

\begin{figure}[htbp]
    \centering
    \includegraphics[width=0.9\linewidth]{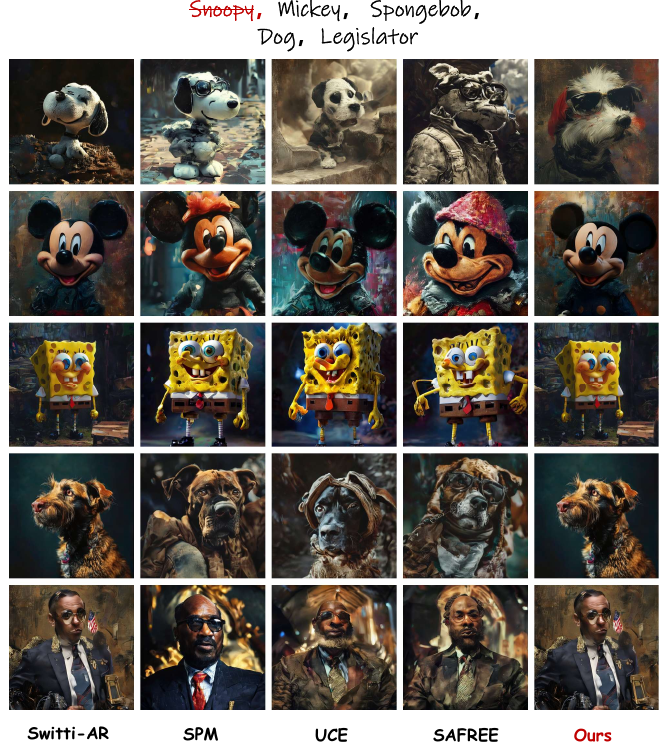}
    \caption{Qualitative comparison of instance concept erasure on the Switti-AR architecture. We target the single concept ``\textit{Snoopy}'' while assessing the prior preservation of non-target concepts. Uni-AdaVD removes the targeted identity, while largely preserving the semantic and structural integrity of non-target concepts compared to baselines.}
    \label{fig:ar_results_snoopy}
\end{figure}

We first present supplementary results for instance concept erasure across multiple architectures. For FLUX, the qualitative comparisons and extended quantitative results are reported in  Fig.~\ref{fig:flux_multi_instance} and Table~\ref{tab:appendix_final_percentage}. For SD v3, the qualitative comparisons and corresponding quantitative results are provided in Fig.~\ref{fig:SD_v3_snoopy} and Table~\ref{tab:appendix_SD_v3_instance}, respectively. We further extend this evaluation to the visual autoregressive setting, with complete results for Switti-AR reported in Fig.~\ref{fig:ar_results_snoopy} and Table~\ref{tab:swittiar_snoopy_appendix_complete}.

\begin{figure}[t]
    \centering
    \includegraphics[width=\linewidth]{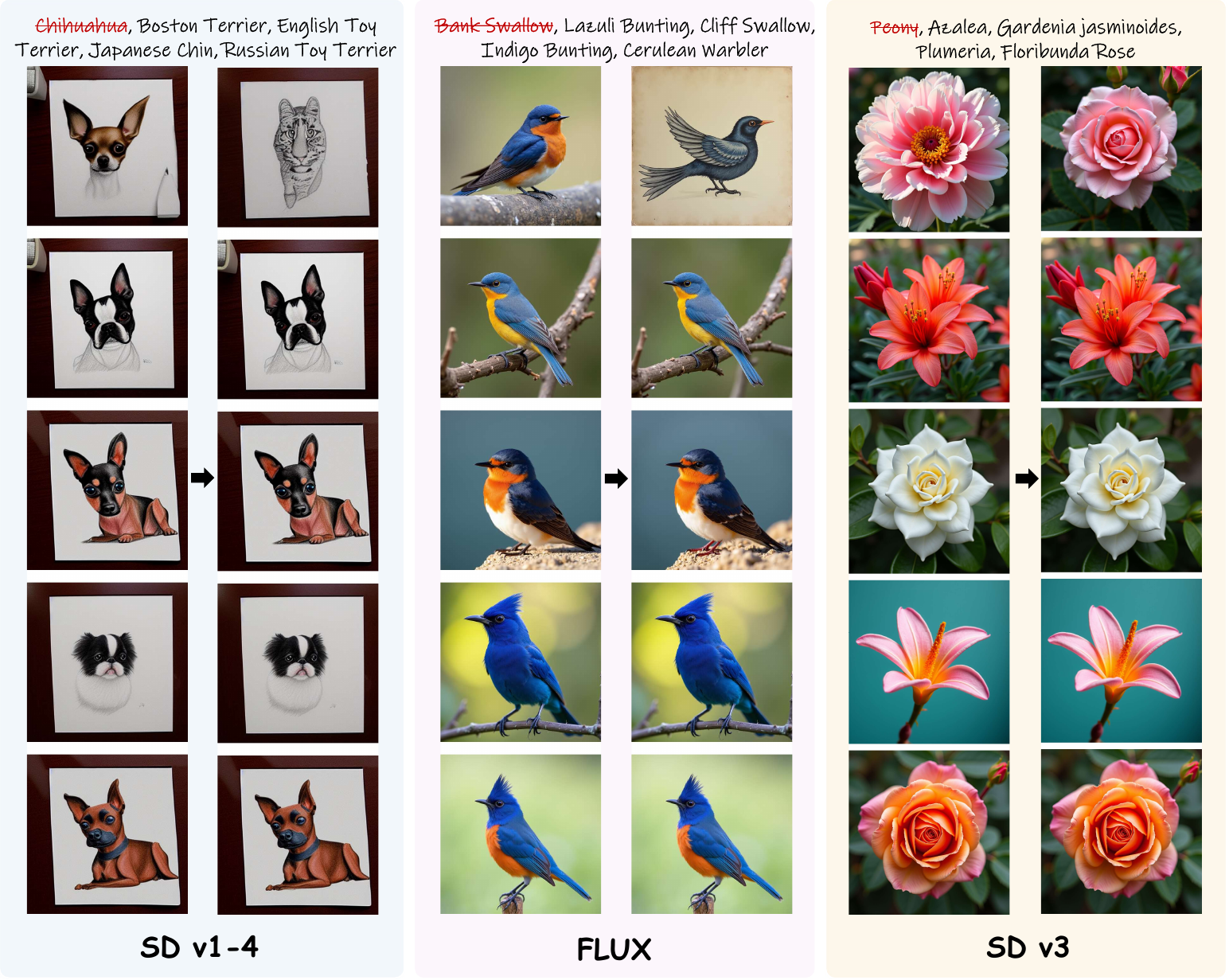}
    \caption{Qualitative evaluation of neighbor concept erasure across SD v1-4, FLUX, and SD v3. Uni-AdaVD mitigates the target concepts (marked in red) while largely preserving the visual characteristics of neighboring categories, such as related dog breeds, bird species, and flowers.}
    \label{fig:neighbor_erasure}
\end{figure}

 Besides, to further verify the erasure precision of Uni-AdaVD, we evaluate Uni-AdaVD on neighbor-concept erasure following the Neighbor-Aware Localized Concept Erasure benchmark \cite{shi2026neighbor}. We consider three fine-grained clusters: dog breeds, bird species, and flower categories. As shown in Fig.~\ref{fig:neighbor_erasure}, we test three representative cases across SD v1-4, FLUX, and SD v3.

\begin{figure*}[htbp]
    \centering

    \subfloat[]{
        \includegraphics[width=0.8\textwidth]{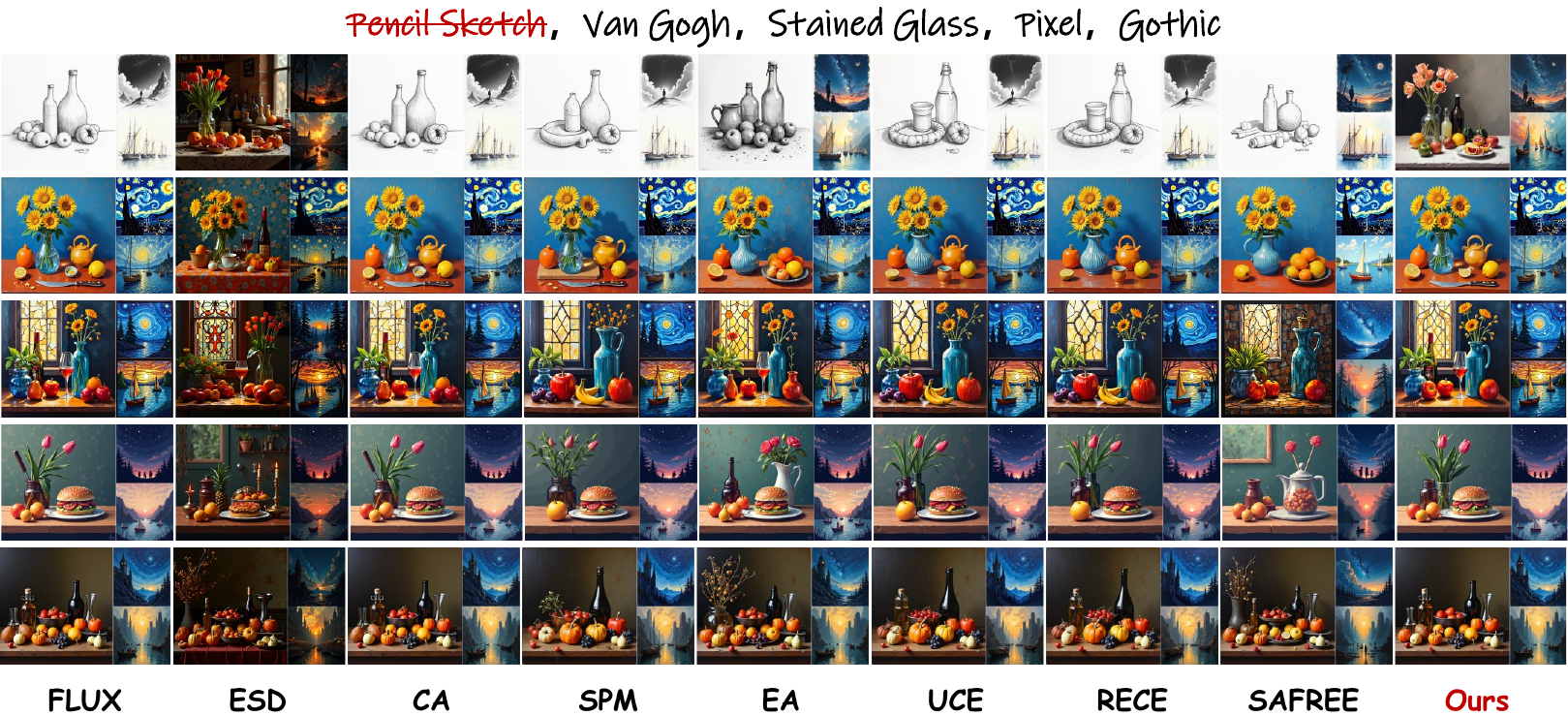}
    }
    \\ 
    \vspace{5pt}

    \subfloat[]{
        \includegraphics[width=0.8\textwidth]{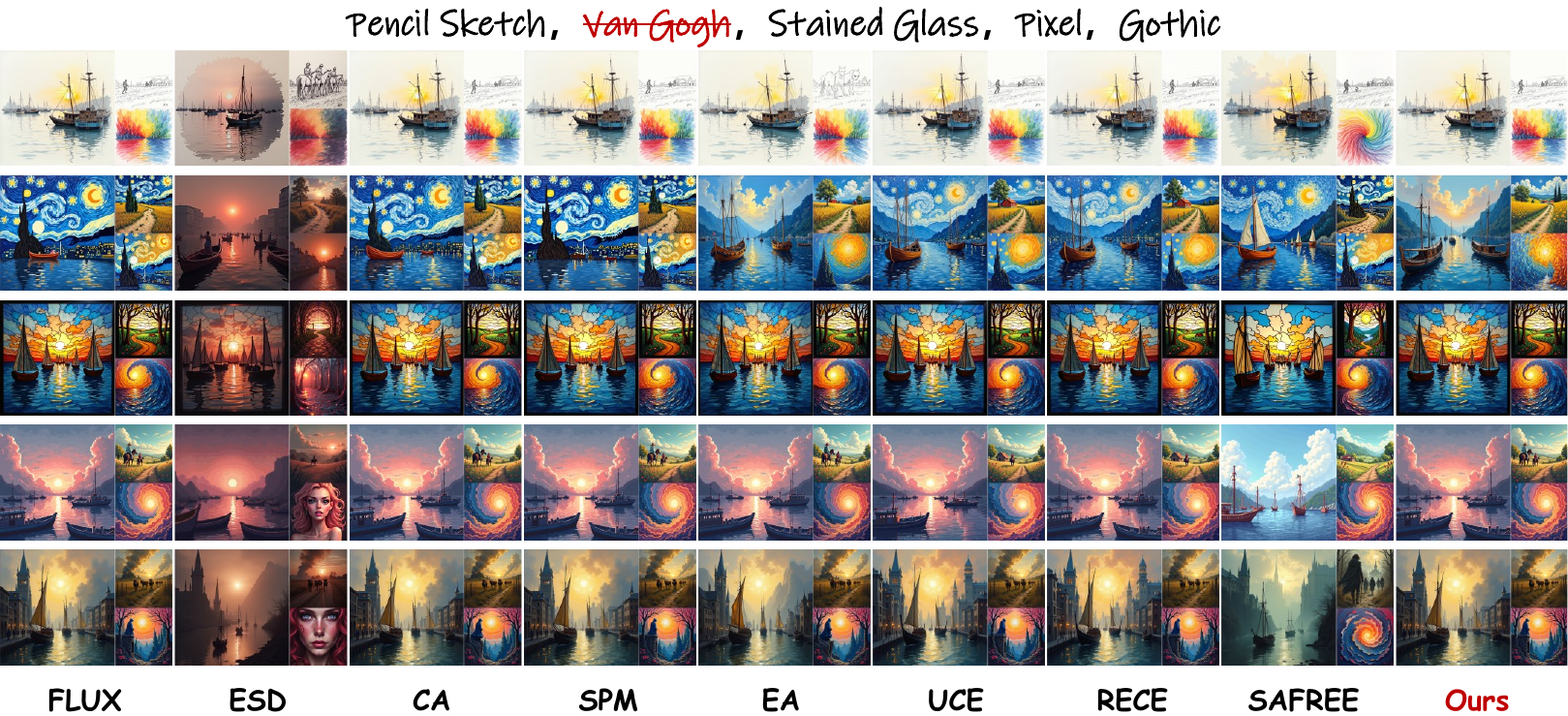}
    }
    \\ 
    \vspace{5pt}

    \subfloat[]{
        \includegraphics[width=0.8\textwidth]{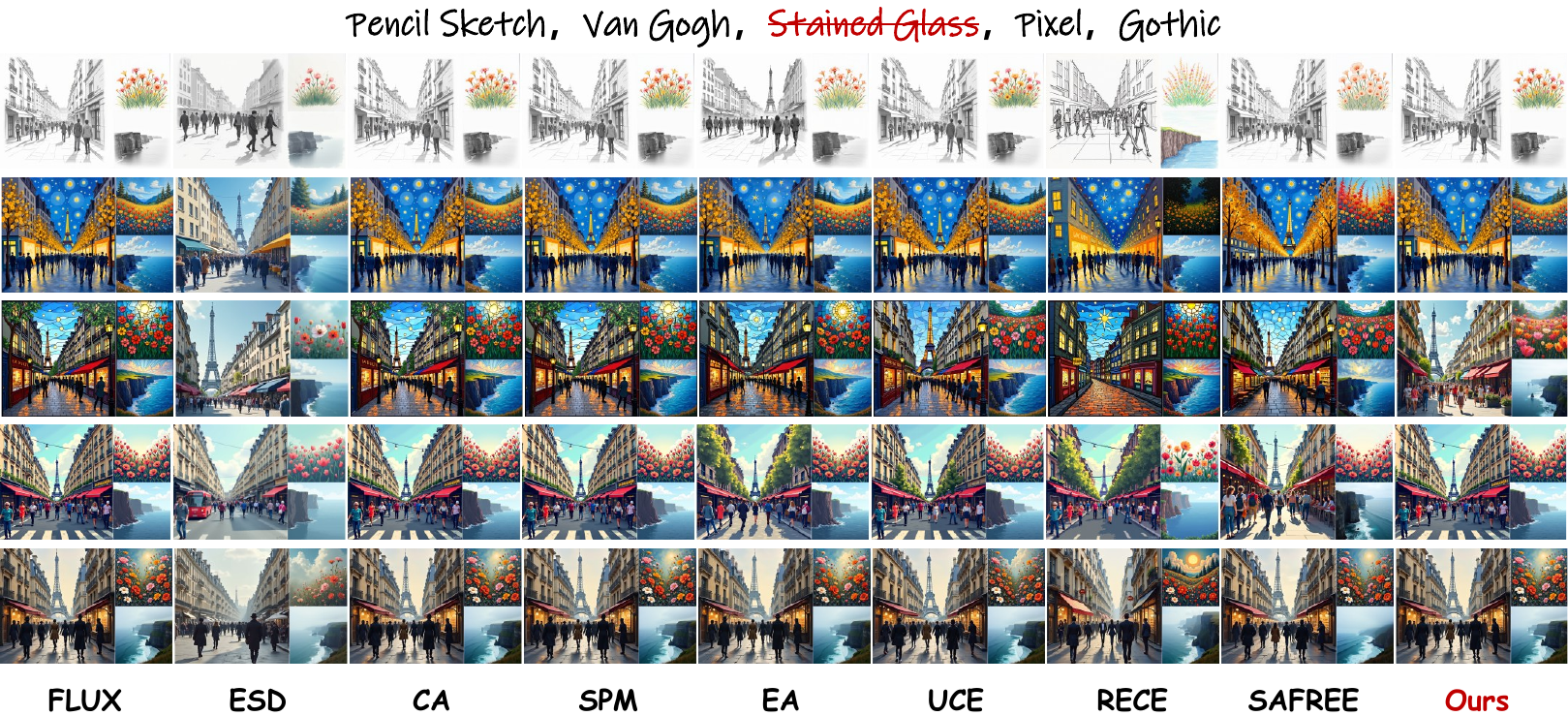}
    }

    \caption{Qualitative comparison of art style erasure on FLUX. From top to bottom, the subfigures demonstrate independent erasure tasks targeting ``\textit{Pencil Sketch}'', ``\textit{Van Gogh}'', and ``\textit{Stained Glass}'', respectively. For each task, our method successfully suppresses the specific target style while maintaining the visual characteristics of the remaining non-target styles (e.g., ``\textit{Pixel}'' and ``\textit{Gothic}''), demonstrating superior prior preservation compared to baseline methods.}
    \label{fig:flux_style_combined_subfigures} 
\end{figure*}

\begin{figure*}
  \centering
  \includegraphics[width=\textwidth]{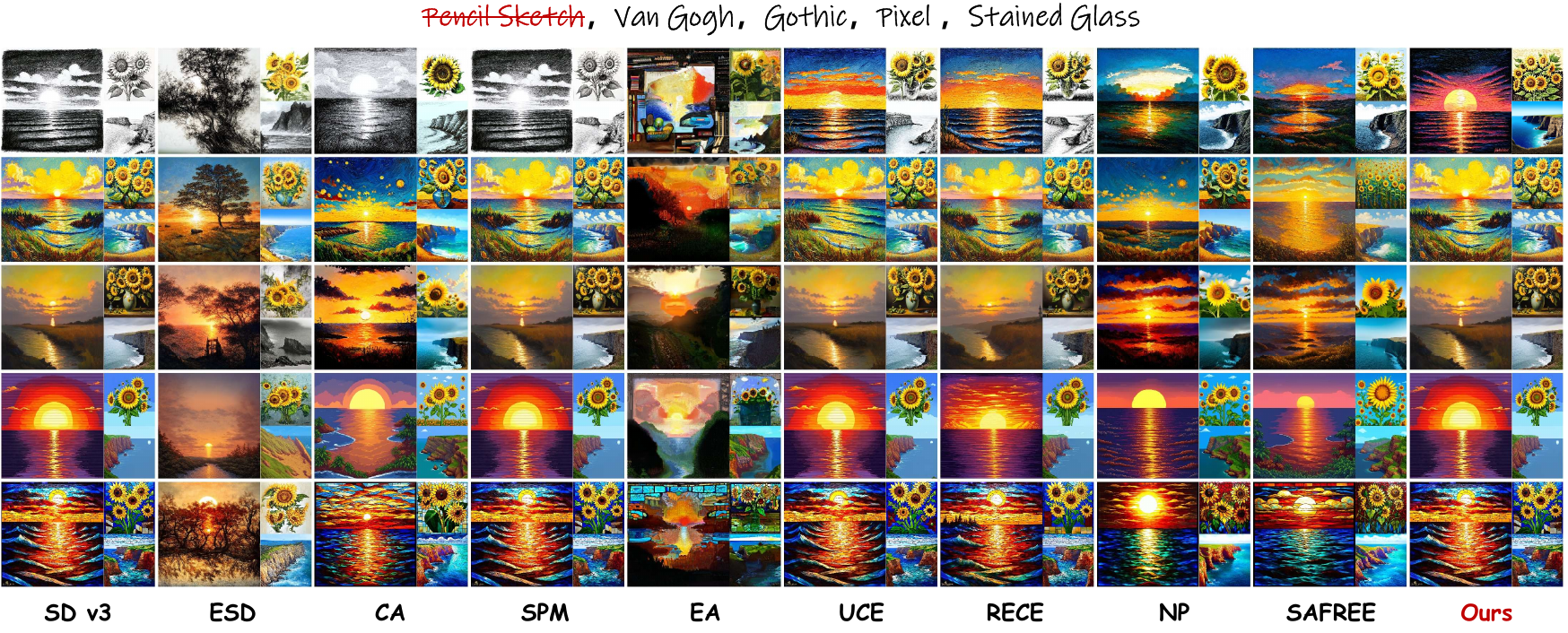}
  \caption{Style erasure on SD v3 targeting ``\textit{Pencil Sketch}'' with prior preservation for other styles, specifically ``\textit{Van Gogh}'', ``\textit{Gothic}'', ``\textit{Pixel}'', and ``\textit{Stained Glass}''. Uni-AdaVD removes the target style while largely maintaining the semantic content and non-target styles.}
  \label{fig:SD_v3——style} 
\end{figure*}

We next evaluate art style erasure. For FLUX, the qualitative comparisons and extended quantitative metrics are shown in Fig.~\ref{fig:flux_style_combined_subfigures} and Table~\ref{tab:appendix_flux_styles}, respectively. For SD v3, the qualitative and quantitative results are provided in Fig.~\ref{fig:SD_v3——style} and Table~\ref{tab:SD_v3_pencil_sketch_appendix}. Overall, these extended results provide additional evidence that Uni-AdaVD suppresses the targeted concepts while largely preserving non-target content.

\subsection{On Celebrity Erasure}
\label{sec:on_celebrity_erasure}

In this section, we evaluate the capability of Uni-AdaVD to erase a specific celebrity identity by targeting the concept ``\textit{Bruce Lee}'' on both the FLUX and SD v3 architectures. To assess prior preservation, we additionally examine the generative quality of unrelated non-target identities.

Table~\ref{tab:face_erasure_appendix} reports the quantitative results on FLUX. While EA and ESD achieve lower CS on the target identity, they are also associated with noticeably higher FID scores on non-target identities such as ``\textit{Marilyn Monroe}'', ``\textit{Melania Trump}'', and ``\textit{Anne Hathaway}''. By contrast, Uni-AdaVD achieves a competitive CS for target removal while maintaining lower FID scores on the evaluated non-target identities.

\begin{figure*} 
  \centering
  \includegraphics[width=0.75\linewidth]{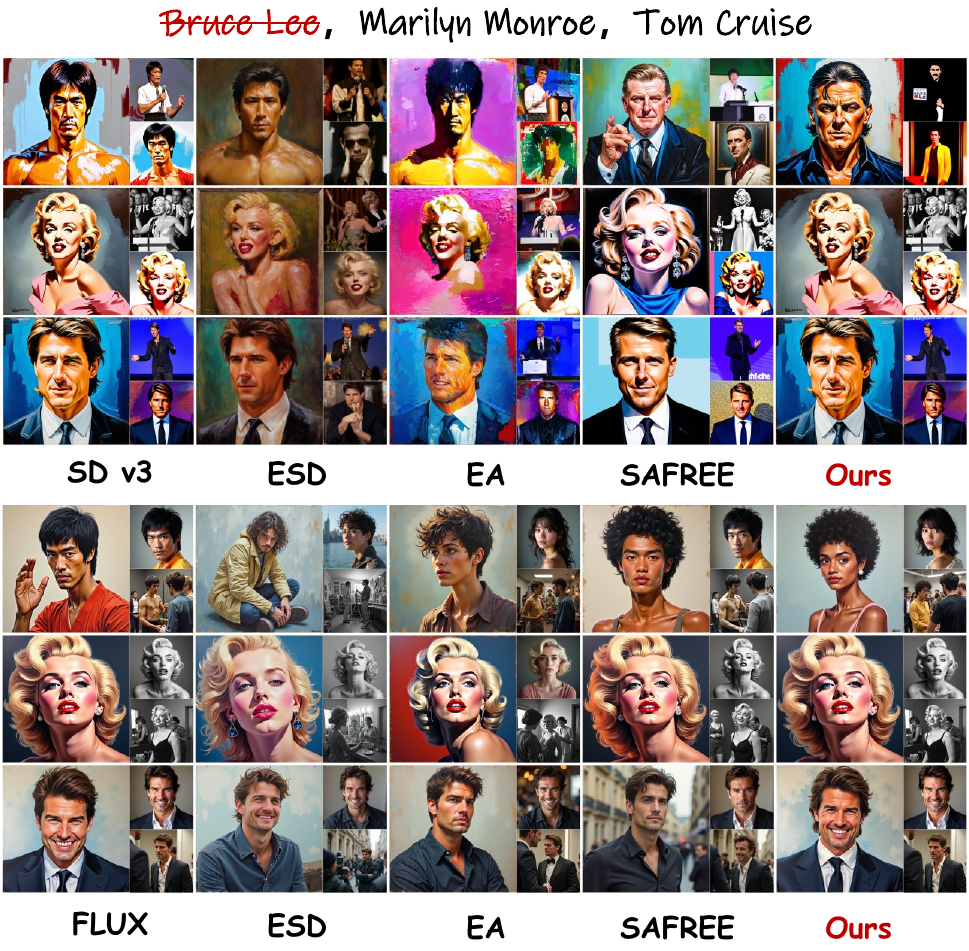} 
  \caption{Qualitative comparison of celebrity identity erasure across SD v3 (top) and FLUX (bottom) architectures. We target the erasure of the ``\textit{Bruce Lee}'' identity, with results displayed in the first row of each section. The subsequent rows assess the preservation of unrelated identities, specifically ``\textit{Marilyn Monroe}'' and ``\textit{Tom Cruise}''. Uni-AdaVD mitigates the target identity features while largely maintaining the visual characteristics and semantic integrity of the non-target concepts.}
  \label{fig:SD_v3_fllux_celebrity_erasure} 
\end{figure*}

The qualitative comparisons in Fig.~\ref{fig:SD_v3_fllux_celebrity_erasure} show a similar pattern. Although most baselines suppress the target identity in the top rows, methods such as ESD and SAFREE often introduce structural artifacts in non-target generations. Uni-AdaVD suppresses target identity cues while largely preserving the semantic consistency and visual quality of non-target identities.

\section{On Transferability to Other T2I Models}
\label{sec:On Transferability to Other T2I Models}

\subsection{Generalization to PixArt Architectures}
\label{sec:pix}

To further validate the architectural generalizability of Uni-AdaVD, we extend our evaluation to the PixArt family, focusing specifically on the high-resolution PixArt-$\Sigma$ \cite{chen2024pixart} model. Unlike U-Net-based models and joint-attention DiT variants, PixArt-$\Sigma$ adopts a pure DiT backbone paired with a T5 text encoder, injecting textual conditions exclusively through cross-attention rather than joint-attention blocks.

\begin{figure*}
    \centering
    \includegraphics[width=\textwidth]{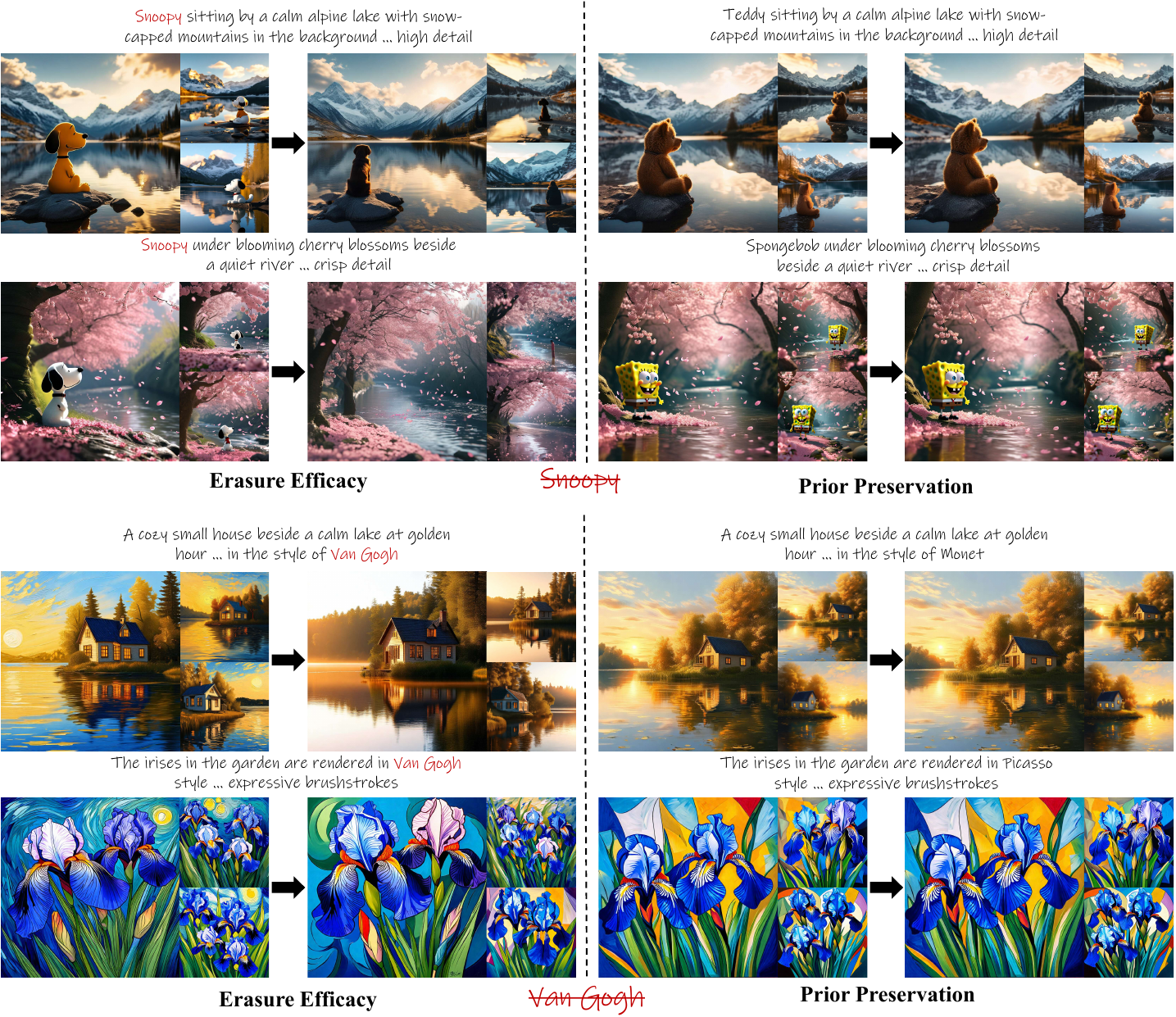}
    \caption{Qualitative evaluation of concept erasure on PixArt-$\Sigma$. By intervening in the cross-attention text value space, the framework successfully suppresses target instances (``\textit{Snoopy}'') and art styles (Van Gogh) while preserving the generative fidelity of non-target concepts under shared prompt contexts.}
    \label{fig:pixart_sigma}
\end{figure*}

Because semantic conditioning in PixArt is mediated entirely by standard cross-attention, we apply our intervention directly to the text value space ($\mathbf{V}_\mathrm{T}$) within its transformer blocks. As shown in Fig.~\ref{fig:pixart_sigma}, we evaluate both instance erasure and artistic style erasure on PixArt-$\Sigma$. In the instance-erasure setting, Uni-AdaVD effectively suppresses the target concept (``\textit{Snoopy}'') while preserving the structural and semantic integrity of non-target concepts such as ``\textit{Teddy}'' and ``\textit{SpongeBob}''. Likewise, in the style-erasure setting, the targeted ``\textit{Van Gogh}'' brushstrokes are neutralized without degrading the generation of other distinctive artistic styles, including ``\textit{Monet}'' and ``\textit{Picasso}''.

These results show that our value-space intervention remains effective and adaptable for DiT-based T2I models that rely on pure cross-attention conditioning, further demonstrating the broad transferability of the Uni-AdaVD framework.

\subsection{Generalization to Infinity-2B}
\label{sec:infinity-2B}

While the main text focuses on Switti-AR, we provide supplementary qualitative results on Infinity-2B to further validate the transferability of Uni-AdaVD across AR architectures. Relative to Switti-AR, Infinity-2B operates at a substantially larger scale, with 2 billion parameters, and adopts a more sophisticated discrete token prediction process, thereby providing a more challenging testbed for our Inference-time intervention.

\begin{figure*}
    \centering
    \includegraphics[width=\textwidth]{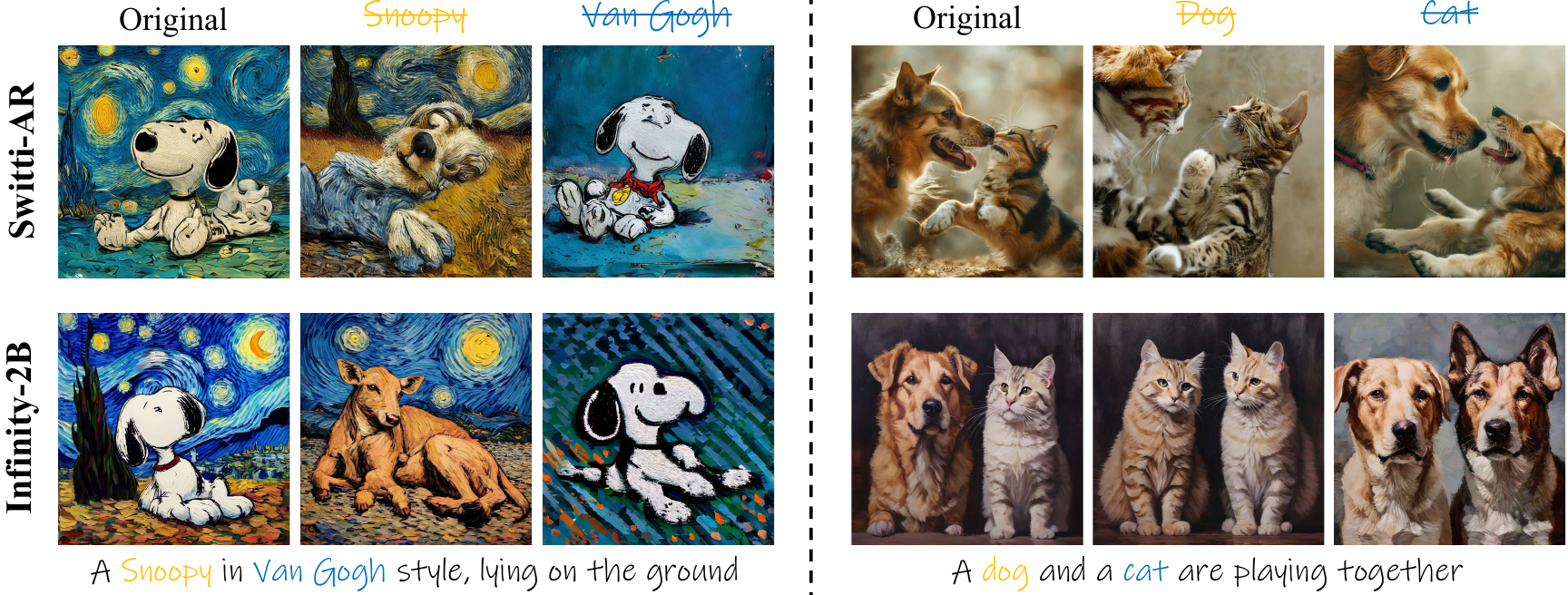} 
    \caption{Qualitative erasure results on AR models (top: Switti-AR, bottom: Infinity-2B). Left: Contrastive erasure of the subject (``\textit{Snoopy}'') versus style (``\textit{Van Gogh}''). Right: Suppression of ``\textit{dog}'' or ``\textit{cat}'' in an interactive scene, preserving the non-target subject.}
    \label{fig:ar_qualitative}
\end{figure*}

Fig.~\ref{fig:ar_qualitative} presents concept-erasure results on both AR models. The left panel illustrates the disentanglement of subject and style: erasing the target instance produces a generic animal while preserving the artistic style, whereas erasing the target style removes the characteristic brushstrokes without altering the subject identity.

The right panel further examines multi-subject interactive scenes. When intervening on either target concept, Uni-AdaVD successfully suppresses the specified concept while preserving the non-target entity. These results further demonstrate the robustness of our value-space intervention across heterogeneous AR generation pipelines.

\begin{figure*}
    \centering
    \includegraphics[width=\textwidth]{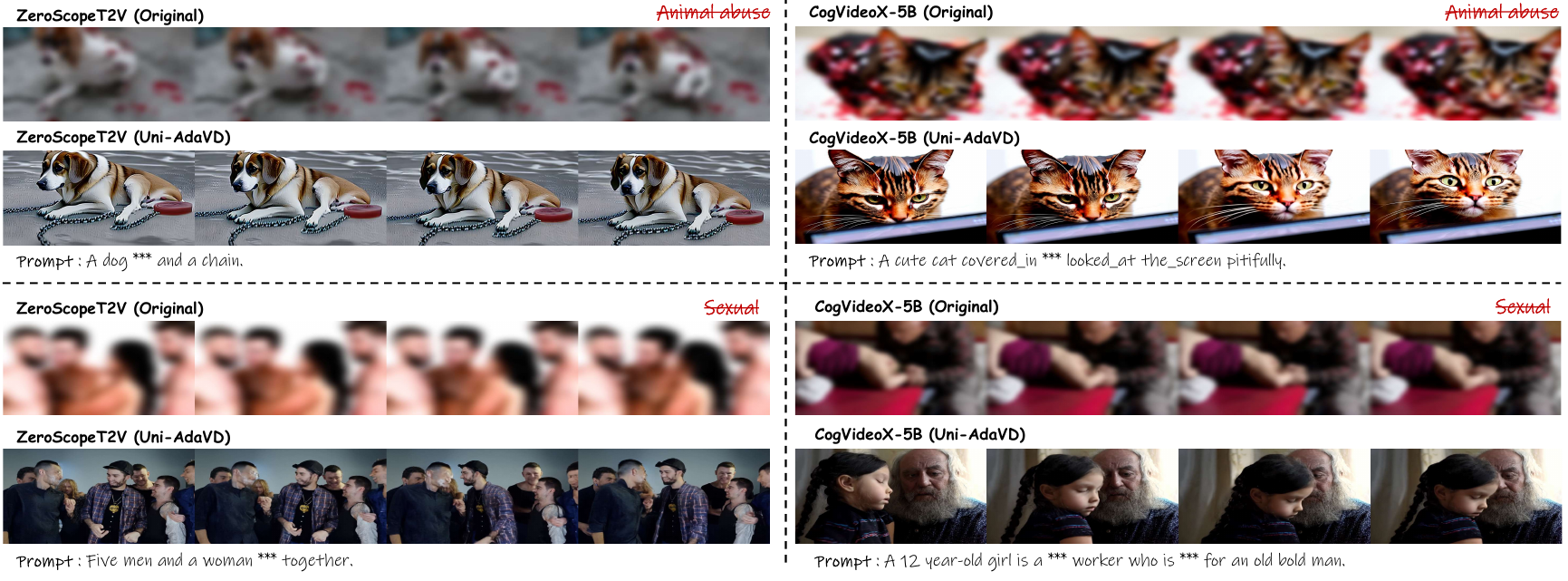}
    \caption{Qualitative results of safety concept erasure on text-to-video models.  Erasure of ``\textit{Animal abuse}'' and ``\textit{Sexual}'' on ZeroScopeT2V and CogVideoX. The original models (top rows, with sensitive content blurred) exhibit severe violations, whereas Uni-AdaVD (bottom rows) suppresses the harmful content while preserving reasonable visual structure.}
    \label{fig:video2}
\end{figure*}

\section{Visualization on T2V Models}
\label{sec:On Transferability to T2V Models}

Following the analysis presented in the main manuscript, this section provides additional qualitative examples of safety concept erasure on video generation models.  As illustrated in Fig.~\ref{fig:video2}, we evaluate the erasure of specific severe safety violations: ``\textit{Animal Abuse}'' and ``\textit{Sexual}'' on ZeroScopeT2V and CogVideoX. While the original unprotected models synthesize frames with explicit violations under malicious prompts, Uni-AdaVD successfully neutralizes the targeted harmful semantics across the generated sequences.

\begin{figure*}
    \centering
    \includegraphics[width=\textwidth]{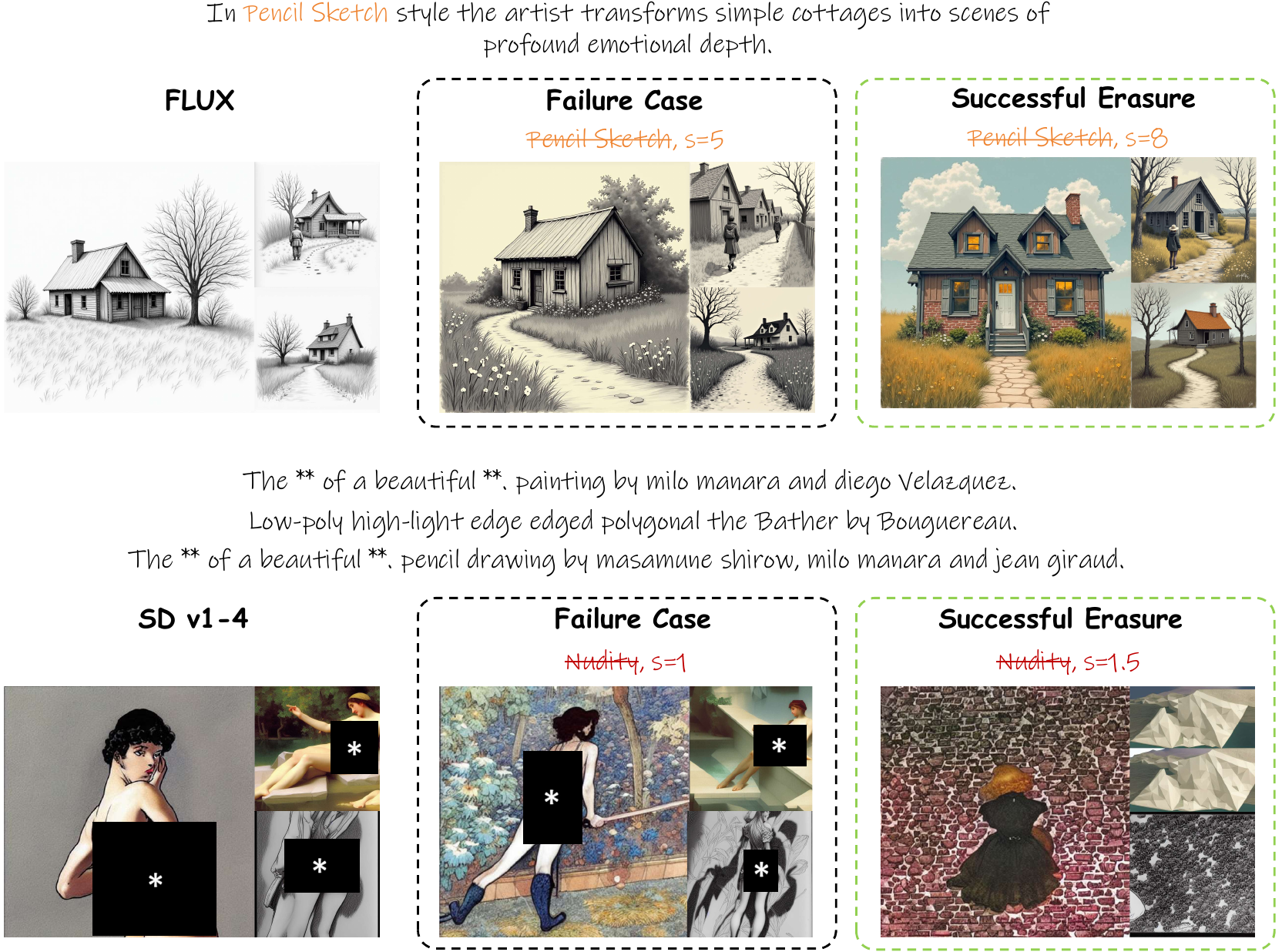}
    \caption{Representative failure cases of Uni-AdaVD and their remedies. Top: in FLUX style erasure, using a smaller scaling factor ($s=5$) leaves residual ``Pencil Sketch'' attributes, whereas increasing it to $s=8$ produces cleaner suppression. Bottom: in SD v1-4 nudity erasure, a conservative setting ($s=1$) may leave residual unsafe semantics, while a slightly larger value ($s=1.5$) yields more complete removal.}
    \label{fig:failure_case}
\end{figure*}

\section{Failure Case Study}
\label{sec:failure_case}

Despite its overall effectiveness, Uni-AdaVD still encounters certain concepts that are difficult to erase completely. Fig.~\ref{fig:failure_case} presents representative failure cases together with their successful corrections.

For artistic style erasure on FLUX, generations associated with ``\textit{Pencil Sketch}'' may exhibit only partial suppression when a relatively conservative scaling factor is used (e.g., $s=5$), still retaining residual grayscale strokes and sketch-like textures. This behavior likely stems from the strong visual prior of certain artistic styles in the pretrained model: the default parameter can attenuate the target style, but may not be sufficiently strong to fully remove the latent representation from the corresponding style manifold. Increasing the scaling factor to $s=8$ effectively overcomes this strong prior, leading to a much more complete suppression of sketch-related cues, as shown on the right side of Fig.~\ref{fig:failure_case}.

A similar phenomenon is observed for implicit concept erasure on SD v1-4. When the erasing strength is set conservatively (e.g., $s=1$), residual unsafe semantics may still remain in the generated image. Implicit concepts, such as nudity triggered by specific contextual combinations in the prompt, can be particularly resistant to mild erasing shifts. By moderately increasing the scaling factor to $s=1.5$, Uni-AdaVD more effectively suppresses the remaining target semantics and produces a benign visual output.

\begin{table*}[t]
\centering
\caption{Complete quantitative comparison of single- and multi-concept instance erasure on the FLUX architecture. The best and second-best results for each primary metric column are marked with \textbf{bold} and \underline{underline}. Columns in gray indicate items that do not directly reflect erasure efficacy or prior preservation performance.}
\label{tab:appendix_final_percentage}
\renewcommand{\arraystretch}{1.3} 

\resizebox{\textwidth}{!}{%
\begin{tabular}{l | cccc | cccc | cccc | cccc | cccc}
\toprule
Concept & \multicolumn{4}{c|}{Snoopy} & \multicolumn{4}{c|}{Mickey} & \multicolumn{4}{c|}{SpongeBob} & \multicolumn{4}{c|}{Dog} & \multicolumn{4}{c}{Legislator} \\ 
\cmidrule(r){1-1} \cmidrule(lr){2-5} \cmidrule(lr){6-9} \cmidrule(lr){10-13} \cmidrule(lr){14-17} \cmidrule(l){18-21}
Metric & CS & FID & SSIM & LPIPS & CS & FID & SSIM & LPIPS & CS & FID & SSIM & LPIPS & CS & FID & SSIM & LPIPS & CS & FID & SSIM & LPIPS \\ 
\midrule
FLUX & 28.08 & -- & -- & -- & 28.05 & -- & -- & -- & 28.97 & -- & -- & -- & 25.74 & -- & -- & -- & 21.92 & -- & -- & -- \\

\specialrule{\lightrulewidth}{0.5pt}{0.5pt}
\multicolumn{21}{c}{\textit{Erase \textbf{Snoopy}}} \\
\specialrule{\lightrulewidth}{0.5pt}{0.5pt}
Metric & CS $\downarrow$ & \textcolor{gray}{FID} & \textcolor{gray}{SSIM} & \textcolor{gray}{LPIPS} & \textcolor{gray}{CS} & FID $\downarrow$ & SSIM $\uparrow$ & LPIPS $\downarrow$ & \textcolor{gray}{CS} & FID $\downarrow$ & SSIM $\uparrow$ & LPIPS $\downarrow$ & \textcolor{gray}{CS} & FID $\downarrow$ & SSIM $\uparrow$ & LPIPS $\downarrow$ & \textcolor{gray}{CS} & FID $\downarrow$ & SSIM $\uparrow$ & LPIPS $\downarrow$ \\ 
\midrule
ESD & 20.91 & \textcolor{gray}{196.96} & \textcolor{gray}{44.50} & \textcolor{gray}{78.35} & \textcolor{gray}{25.26} & 103.05 & 42.73 & 66.74 & \textcolor{gray}{27.23} & 133.41 & 41.15 & 69.50 & \textcolor{gray}{24.18} & 70.07 & 46.41 & 62.30 & \textcolor{gray}{18.72} & 67.57 & 50.78 & 59.72 \\
CA & 27.04 & \textcolor{gray}{22.33} & \textcolor{gray}{83.29} & \textcolor{gray}{24.15} & \textcolor{gray}{27.08} & \underline{22.48} & \underline{78.06} & \underline{21.71} & \textcolor{gray}{27.92} & \underline{23.18} & \textbf{86.16} & \textbf{14.40} & \textcolor{gray}{24.81} & \underline{21.07} & \underline{85.75} & \textbf{15.09} & \textcolor{gray}{20.82} & \underline{23.97} & \underline{87.79} & \underline{24.02} \\
SPM & 27.15 & \textcolor{gray}{36.82} & \textcolor{gray}{81.16} & \textcolor{gray}{26.72} & \textcolor{gray}{27.20} & 22.98 & 69.39 & 36.53 & \textcolor{gray}{27.97} & 25.62 & 75.12 & 29.84 & \textcolor{gray}{24.85} & 23.86 & 76.87 & 29.34 & \textcolor{gray}{21.08} & 29.70 & 69.28 & 40.01 \\
EA & \textbf{18.30} & \textcolor{gray}{144.18} & \textcolor{gray}{59.26} & \textcolor{gray}{69.94} & \textcolor{gray}{26.56} & 36.31 & 66.90 & 40.38 & \textcolor{gray}{27.29} & 40.22 & 70.80 & 37.25 & \textcolor{gray}{24.08} & 28.81 & 75.98 & 30.91 & \textcolor{gray}{19.74} & 35.87 & 69.64 & 40.15 \\
UCE & 25.97 & \textcolor{gray}{60.84} & \textcolor{gray}{70.49} & \textcolor{gray}{44.28} & \textcolor{gray}{27.22} & 23.14 & 69.50 & 36.15 & \textcolor{gray}{25.97} & 26.84 & 70.49 & 44.28 & \textcolor{gray}{24.84} & 24.61 & 76.58 & 29.60 & \textcolor{gray}{20.97} & 28.63 & 69.05 & 40.52 \\
RECE & 22.74 & \textcolor{gray}{88.31} & \textcolor{gray}{58.50} & \textcolor{gray}{63.05} & \textcolor{gray}{22.01} & 58.59 & 44.33 & 72.30 & \textcolor{gray}{22.74} & 68.31 & 58.50 & 63.05 & \textcolor{gray}{22.69} & 69.28 & 52.29 & 66.99 & \textcolor{gray}{17.27} & 98.19 & 45.15 & 71.72 \\
SAFREE & 24.23 & \textcolor{gray}{92.85} & \textcolor{gray}{65.37} & \textcolor{gray}{47.42} & \textcolor{gray}{26.14} & 24.35 & 67.14 & 49.81 & \textcolor{gray}{27.17} & 26.64 & 71.42 & 46.70 & \textcolor{gray}{23.39} & 36.09 & 69.80 & 41.50 & \textcolor{gray}{19.91} & 43.74 & 65.59 & 45.71 \\
\textbf{Ours} & \underline{20.10} & \textcolor{gray}{94.60} & \textcolor{gray}{58.70} & \textcolor{gray}{65.06} & \textcolor{gray}{27.05} & \textbf{16.66} & \textbf{79.77} & \underline{22.40} & \textcolor{gray}{28.03} & \textbf{17.07} & \underline{85.72} & \underline{15.17} & \textcolor{gray}{24.83} & \textbf{9.25} & \textbf{92.79} & \underline{8.05} & \textcolor{gray}{20.90} & \textbf{11.70} & \textbf{90.34} & \textbf{10.81} \\

\specialrule{\lightrulewidth}{0.5pt}{0.5pt}
\multicolumn{21}{c}{\textit{Erase \textbf{Snoopy} and \textbf{Mickey}}} \\
\specialrule{\lightrulewidth}{0.5pt}{0.5pt}
Metric & CS $\downarrow$ & \textcolor{gray}{FID} & \textcolor{gray}{SSIM} & \textcolor{gray}{LPIPS} & CS $\downarrow$ & \textcolor{gray}{FID} & \textcolor{gray}{SSIM} & \textcolor{gray}{LPIPS} & \textcolor{gray}{CS} & FID $\downarrow$ & SSIM $\uparrow$ & LPIPS $\downarrow$ & \textcolor{gray}{CS} & FID $\downarrow$ & SSIM $\uparrow$ & LPIPS $\downarrow$ & \textcolor{gray}{CS} & FID $\downarrow$ & SSIM $\uparrow$ & LPIPS $\downarrow$ \\ 
\midrule
ESD & 20.27 & \textcolor{gray}{237.70} & \textcolor{gray}{48.41} & \textcolor{gray}{75.17} & \textbf{20.08} & \textcolor{gray}{233.55} & \textcolor{gray}{45.83} & \textcolor{gray}{73.80} & \textcolor{gray}{21.95} & 172.66 & 50.94 & 69.92 & \textcolor{gray}{21.38} & 91.98 & 46.38 & 69.92 & \textcolor{gray}{15.88} & 76.58 & 56.90 & 60.26 \\
CA & 25.88 & \textcolor{gray}{40.17} & \textcolor{gray}{80.91} & \textcolor{gray}{38.24} & 26.93 & \textcolor{gray}{42.71} & \textcolor{gray}{75.92} & \textcolor{gray}{39.27} & \textcolor{gray}{27.17} & \underline{32.90} & \textbf{88.35} & \textbf{21.21} & \textcolor{gray}{24.83} & \underline{21.76} & \textbf{88.89} & \textbf{21.23} & \textcolor{gray}{20.43} & \underline{22.54} & \underline{85.89} & \underline{27.39} \\
SPM & 27.12 & \textcolor{gray}{47.32} & \textcolor{gray}{73.78} & \textcolor{gray}{39.49} & 27.46 & \textcolor{gray}{49.33} & \textcolor{gray}{60.61} & \textcolor{gray}{48.47} & \textcolor{gray}{27.34} & 42.05 & 68.08 & 40.50 & \textcolor{gray}{24.87} & 34.90 & 70.47 & 40.15 & \textcolor{gray}{20.66} & 32.43 & 65.00 & 48.13 \\
EA & \textbf{18.36} & \textcolor{gray}{132.08} & \textcolor{gray}{60.36} & \textcolor{gray}{58.38} & \underline{20.31} & \textcolor{gray}{134.46} & \textcolor{gray}{46.78} & \textcolor{gray}{68.38} & \textcolor{gray}{26.27} & 73.06 & 63.27 & 52.50 & \textcolor{gray}{24.02} & 47.62 & 53.27 & 49.79 & \textcolor{gray}{19.70} & 43.91 & 61.02 & 38.20 \\
UCE & 23.41 & \textcolor{gray}{105.54} & \textcolor{gray}{65.02} & \textcolor{gray}{52.77} & 23.63 & \textcolor{gray}{86.80} & \textcolor{gray}{57.05} & \textcolor{gray}{55.73} & \textcolor{gray}{27.19} & 32.17 & 74.77 & 29.91 & \textcolor{gray}{24.90} & 24.92 & 76.10 & 30.08 & \textcolor{gray}{21.00} & 31.23 & 67.54 & 42.70 \\
RECE & 21.58 & \textcolor{gray}{116.82} & \textcolor{gray}{55.90} & \textcolor{gray}{65.32} & 20.94 & \textcolor{gray}{166.51} & \textcolor{gray}{43.98} & \textcolor{gray}{73.95} & \textcolor{gray}{24.14} & 78.39 & 52.96 & 69.09 & \textcolor{gray}{22.73} & 69.28 & 52.27 & 66.98 & \textcolor{gray}{17.46} & 90.97 & 45.68 & 70.98 \\
SAFREE & 21.03 & \textcolor{gray}{119.37} & \textcolor{gray}{62.34} & \textcolor{gray}{58.72} & 22.31 & \textcolor{gray}{129.63} & \textcolor{gray}{50.23} & \textcolor{gray}{61.87} & \textcolor{gray}{26.98} & 64.75 & 81.94 & 41.41 & \textcolor{gray}{23.99} & 54.75 & 51.94 & 52.73 & \textcolor{gray}{17.49} & 56.63 & 58.37 & 49.28 \\
\textbf{Ours} & \underline{19.93} & \textcolor{gray}{117.35} & \textcolor{gray}{57.21} & \textcolor{gray}{59.15} & 21.23 & \textcolor{gray}{136.17} & \textcolor{gray}{48.79} & \textcolor{gray}{62.80} & \textcolor{gray}{27.18} & \textbf{17.78} & \underline{87.99} & \underline{35.30} & \textcolor{gray}{24.33} & \textbf{10.97} & \underline{86.72} & \underline{23.30} & \textcolor{gray}{20.56} & \textbf{16.49} & \textbf{86.59} & \textbf{15.12} \\

\specialrule{\lightrulewidth}{0.5pt}{0.5pt}
\multicolumn{21}{c}{\textit{Erase \textbf{Snoopy}, \textbf{Mickey} and \textbf{SpongeBob}}} \\
\specialrule{\lightrulewidth}{0.5pt}{0.5pt}
Metric & CS $\downarrow$ & \textcolor{gray}{FID} & \textcolor{gray}{SSIM} & \textcolor{gray}{LPIPS} & CS $\downarrow$ & \textcolor{gray}{FID} & \textcolor{gray}{SSIM} & \textcolor{gray}{LPIPS} & CS $\downarrow$ & \textcolor{gray}{FID} & \textcolor{gray}{SSIM} & \textcolor{gray}{LPIPS} & \textcolor{gray}{CS} & FID $\downarrow$ & SSIM $\uparrow$ & LPIPS $\downarrow$ & \textcolor{gray}{CS} & FID $\downarrow$ & SSIM $\uparrow$ & LPIPS $\downarrow$ \\ 
\midrule
ESD & 20.95 & \textcolor{gray}{219.78} & \textcolor{gray}{48.83} & \textcolor{gray}{75.83} & 21.60 & \textcolor{gray}{225.39} & \textcolor{gray}{46.20} & \textcolor{gray}{60.41} & 21.92 & \textcolor{gray}{269.06} & \textcolor{gray}{47.23} & \textcolor{gray}{77.77} & \textcolor{gray}{22.50} & 97.30 & 56.85 & 65.50 & \textcolor{gray}{16.32} & 72.10 & 57.86 & 61.07 \\
CA & 25.01 & \textcolor{gray}{41.91} & \textcolor{gray}{75.48} & \textcolor{gray}{37.96} & 26.44 & \textcolor{gray}{42.26} & \textcolor{gray}{76.52} & \textcolor{gray}{35.17} & 27.69 & \textcolor{gray}{40.61} & \textcolor{gray}{70.26} & \textcolor{gray}{37.18} & \textcolor{gray}{20.70} & \underline{21.56} & \textbf{82.38} & \textbf{15.67} & \textcolor{gray}{20.51} & \underline{23.08} & \textbf{87.58} & \underline{24.19} \\
SPM & 27.06 & \textcolor{gray}{40.79} & \textcolor{gray}{75.93} & \textcolor{gray}{37.08} & 27.36 & \textcolor{gray}{41.40} & \textcolor{gray}{63.17} & \textcolor{gray}{46.35} & 28.08 & \textcolor{gray}{43.36} & \textcolor{gray}{69.10} & \textcolor{gray}{38.63} & \textcolor{gray}{24.71} & 33.88 & 71.64 & 39.30 & \textcolor{gray}{20.56} & 34.99 & 64.99 & 48.83 \\
EA & \textbf{19.80} & \textcolor{gray}{147.66} & \textcolor{gray}{63.70} & \textcolor{gray}{56.37} & \textbf{20.24} & \textcolor{gray}{78.22} & \textcolor{gray}{59.42} & \textcolor{gray}{58.93} & \underline{21.62} & \textcolor{gray}{149.75} & \textcolor{gray}{63.15} & \textcolor{gray}{61.40} & \textcolor{gray}{23.06} & 61.63 & 68.64 & 48.43 & \textcolor{gray}{19.34} & 90.57 & 63.11 & 57.13 \\
UCE & 23.14 & \textcolor{gray}{112.88} & \textcolor{gray}{64.33} & \textcolor{gray}{53.69} & 23.55 & \textcolor{gray}{90.10} & \textcolor{gray}{56.53} & \textcolor{gray}{56.49} & 26.56 & \textcolor{gray}{127.13} & \textcolor{gray}{61.55} & \textcolor{gray}{58.36} & \textcolor{gray}{24.93} & 24.94 & 75.87 & 30.41 & \textcolor{gray}{20.86} & 29.93 & 67.83 & 42.07 \\
RECE & 21.46 & \textcolor{gray}{116.29} & \textcolor{gray}{55.78} & \textcolor{gray}{65.28} & 21.72 & \textcolor{gray}{165.59} & \textcolor{gray}{43.96} & \textcolor{gray}{73.98} & 22.37 & \textcolor{gray}{193.27} & \textcolor{gray}{50.48} & \textcolor{gray}{75.17} & \textcolor{gray}{22.72} & 69.07 & 52.38 & 66.87 & \textcolor{gray}{17.38} & 100.35 & 45.33 & 71.37 \\
SAFREE & \underline{20.47} & \textcolor{gray}{100.37} & \textcolor{gray}{63.56} & \textcolor{gray}{54.39} & 21.63 & \textcolor{gray}{105.24} & \textcolor{gray}{62.88} & \textcolor{gray}{62.35} & 21.79 & \textcolor{gray}{143.98} & \textcolor{gray}{69.82} & \textcolor{gray}{60.53} & \textcolor{gray}{18.22} & 59.49 & 60.73 & 50.96 & \textcolor{gray}{17.49} & 48.11 & 71.37 & 45.02 \\
\textbf{Ours} & 20.90 & \textcolor{gray}{103.79} & \textcolor{gray}{62.06} & \textcolor{gray}{56.72} & \underline{21.58} & \textcolor{gray}{111.01} & \textcolor{gray}{52.80} & \textcolor{gray}{60.41} & \textbf{20.48} & \textcolor{gray}{223.47} & \textcolor{gray}{49.93} & \textcolor{gray}{71.91} & \textcolor{gray}{25.32} & \textbf{11.69} & \underline{79.51} & \underline{21.45} & \textcolor{gray}{20.88} & \textbf{15.59} & \underline{86.21} & \textbf{16.04} \\ 
\bottomrule
\end{tabular}%
}
\end{table*}

\begin{table*}[htbp]
\centering
\caption{Complete quantitative comparison of single-concept instance erasure on SD v3. The best and second-best results for each primary metric column are marked with \textbf{bold} and \underline{underline}. Columns in gray indicate items that do not directly reflect erasure efficacy or prior preservation performance.}
\label{tab:appendix_SD_v3_instance}
\renewcommand{\arraystretch}{1.3} 

\resizebox{\textwidth}{!}{%
\begin{tabular}{l | cccc | cccc | cccc | cccc | cccc}
\toprule
Concept & \multicolumn{4}{c|}{Snoopy} & \multicolumn{4}{c|}{Mickey} & \multicolumn{4}{c|}{SpongeBob} & \multicolumn{4}{c|}{Dog} & \multicolumn{4}{c}{Legislator} \\ 
\cmidrule(r){1-1} \cmidrule(lr){2-5} \cmidrule(lr){6-9} \cmidrule(lr){10-13} \cmidrule(lr){14-17} \cmidrule(l){18-21}
Metric & CS & FID & SSIM & LPIPS & CS & FID & SSIM & LPIPS & CS & FID & SSIM & LPIPS & CS & FID & SSIM & LPIPS & CS & FID & SSIM & LPIPS \\ 
\midrule
SD v3 & 28.48 & -- & -- & -- & 26.89 & -- & -- & -- & 29.49 & -- & -- & -- & 25.22 & -- & -- & -- & 23.47 & -- & -- & -- \\

\specialrule{\lightrulewidth}{0.5pt}{0.5pt}
\multicolumn{21}{c}{\textit{Erase \textbf{Snoopy}}} \\
\specialrule{\lightrulewidth}{0.5pt}{0.5pt}
Metric & CS $\downarrow$ & \textcolor{gray}{FID} & \textcolor{gray}{SSIM} & \textcolor{gray}{LPIPS} & \textcolor{gray}{CS} & FID $\downarrow$ & SSIM $\uparrow$ & LPIPS $\downarrow$ & \textcolor{gray}{CS} & FID $\downarrow$ & SSIM $\uparrow$ & LPIPS $\downarrow$ & \textcolor{gray}{CS} & FID $\downarrow$ & SSIM $\uparrow$ & LPIPS $\downarrow$ & \textcolor{gray}{CS} & FID $\downarrow$ & SSIM $\uparrow$ & LPIPS $\downarrow$ \\ 
\midrule
ESD & \underline{20.94} & \textcolor{gray}{140.17} & \textcolor{gray}{28.03} & \textcolor{gray}{86.78} & \textcolor{gray}{24.85} & 107.43 & 42.14 & 84.56 & \textcolor{gray}{26.73} & 141.51 & 38.09 & 86.24 & \textcolor{gray}{23.06} & 61.38 & 41.98 & 82.11 & \textcolor{gray}{20.40} & 72.05 & 37.88 & 85.41 \\
CA & 27.61 & \textcolor{gray}{35.04} & \textcolor{gray}{62.51} & \textcolor{gray}{47.97} & \textcolor{gray}{26.67} & \underline{29.85} & \underline{62.28} & \underline{37.97} & \textcolor{gray}{28.56} & \underline{29.96} & \underline{62.57} & \underline{46.32} & \textcolor{gray}{24.21} & 26.05 & 65.03 & 44.84 & \textcolor{gray}{22.36} & 38.87 & 36.25 & 80.65 \\

SPM & 28.43 & \textcolor{gray}{31.41} & \textcolor{gray}{63.58} & \textcolor{gray}{48.28} & \textcolor{gray}{26.92} & 31.18 & 60.89 & 48.57 & \textcolor{gray}{28.51} & 34.69 & 60.87 & 48.34 & \textcolor{gray}{24.22} & 27.04 & 64.66 & 45.88 & \textcolor{gray}{22.45} & 32.52 & 61.58 & 48.07 \\

EA & \textbf{19.42} & \textcolor{gray}{160.53} & \textcolor{gray}{25.72} & \textcolor{gray}{88.02} & \textcolor{gray}{22.52} & 156.78 & 43.74 & 91.21 & \textcolor{gray}{25.16} & 166.29 & 38.36 & 86.97 & \textcolor{gray}{23.87} & 198.17 & 44.33 & 79.99 & \textcolor{gray}{22.46} & 59.83 & 31.76 & 85.45 \\

UCE & 28.53 & \textcolor{gray}{33.40} & \textcolor{gray}{61.57} & \textcolor{gray}{42.86} & \textcolor{gray}{26.70} & 30.67 & 58.16 & 44.38 & \textcolor{gray}{28.34} & 43.93 & 54.96 & 47.78 & \textcolor{gray}{24.30} & 22.97 & 67.57 & 38.75 & \textcolor{gray}{22.13} & 32.37 & 60.07 & 41.71 \\

RECE & 28.14 & \textcolor{gray}{46.66} & \textcolor{gray}{61.42} & \textcolor{gray}{47.49} & \textcolor{gray}{27.05} & 31.03 & 61.35 & 42.39 & \textcolor{gray}{28.37} & 37.30 & 60.20 & 48.99 & \textcolor{gray}{24.14} & \underline{22.87} & \underline{74.51} &  \underline{30.92} & \textcolor{gray}{22.16} & \underline{29.00} & \underline{67.08} & \underline{33.92} \\

NP & 27.46 & \textcolor{gray}{89.94} & \textcolor{gray}{30.00} & \textcolor{gray}{79.72} & \textcolor{gray}{27.17} & 51.37 & 47.38 & 78.64 & \textcolor{gray}{28.59} & 50.54 & 32.30 & 76.85 & \textcolor{gray}{24.21} & 60.34 & 35.45 & 74.82 & \textcolor{gray}{21.50} & 48.76 & 37.40 & 80.94 \\
SAFREE & 22.90 & \textcolor{gray}{90.07} & \textcolor{gray}{31.60} & \textcolor{gray}{79.32} & \textcolor{gray}{26.80} & 47.67 & 48.22 & 76.79 & \textcolor{gray}{28.00} & 52.91 & 33.23 & 75.52 & \textcolor{gray}{23.92} & 63.53 & 35.05 & 74.22 & \textcolor{gray}{21.48} & 45.75 & 39.20 & 79.41 \\
\textbf{Ours} & 22.01 & \textcolor{gray}{126.56} & \textcolor{gray}{29.20} & \textcolor{gray}{83.10} & \textcolor{gray}{26.87} & \textbf{9.69} & \textbf{92.26} & \textbf{7.15} & \textcolor{gray}{28.50} & \textbf{12.98} & \textbf{92.34} & \textbf{6.89} & \textcolor{gray}{24.20} & \textbf{5.69} & \textbf{95.86} & \textbf{4.56} & \textcolor{gray}{22.44} & \textbf{11.33} & \textbf{93.03} & \textbf{6.65} \\

\specialrule{\lightrulewidth}{0.5pt}{0.5pt}
\multicolumn{21}{c}{\textit{Erase \textbf{Snoopy} and \textbf{Mickey}}} \\
\specialrule{\lightrulewidth}{0.5pt}{0.5pt}
Metric & CS $\downarrow$ & \textcolor{gray}{FID} & \textcolor{gray}{SSIM} & \textcolor{gray}{LPIPS} & CS $\downarrow$ & \textcolor{gray}{FID} & \textcolor{gray}{SSIM} & \textcolor{gray}{LPIPS} & \textcolor{gray}{CS} & FID $\downarrow$ & SSIM $\uparrow$ & LPIPS $\downarrow$ & \textcolor{gray}{CS} & FID $\downarrow$ & SSIM $\uparrow$ & LPIPS $\downarrow$ & \textcolor{gray}{CS} & FID $\downarrow$ & SSIM $\uparrow$ & LPIPS $\downarrow$ \\ 
\midrule
ESD & \underline{21.33} & \textcolor{gray}{146.55} & \textcolor{gray}{39.14} & \textcolor{gray}{79.26} & \underline{23.59} & \textcolor{gray}{107.18} & \textcolor{gray}{33.47} & \textcolor{gray}{80.74} & \textcolor{gray}{26.74} & 119.19 & 41.81 & 75.84 & \textcolor{gray}{23.17} & 58.52 & 48.95 & 69.58 & \textcolor{gray}{21.49} & 69.01 & 38.10 & 76.95 \\
CA & 27.55 & \textcolor{gray}{34.34} & \textcolor{gray}{55.69} & \textcolor{gray}{48.16} & 26.39 & \textcolor{gray}{31.58} & \textcolor{gray}{54.49} & \textcolor{gray}{49.32} & \textcolor{gray}{28.51} & 36.57 & 53.99 & 51.49 & \textcolor{gray}{24.24} & 23.25 & 64.01 & 43.98 & \textcolor{gray}{22.32} & 35.87 & 50.31 & 52.07 \\

SPM & 28.54 & \textcolor{gray}{37.52} & \textcolor{gray}{63.63} & \textcolor{gray}{46.18} & 26.96 & \textcolor{gray}{36.98} & \textcolor{gray}{61.18} & \textcolor{gray}{48.25} & \textcolor{gray}{28.53} & \underline{32.10} & 60.71 & 48.43 & \textcolor{gray}{24.22} & 23.21 & \underline{74.85} & \underline{29.70} & \textcolor{gray}{22.45} & 30.66 & \underline{71.69} & \underline{29.97} \\

EA & \textbf{20.08} & \textcolor{gray}{173.14} & \textcolor{gray}{35.57} & \textcolor{gray}{80.50} & \textbf{18.94} & \textcolor{gray}{188.33} & \textcolor{gray}{36.41} & \textcolor{gray}{86.68} & \textcolor{gray}{22.60} & 183.30 & 33.09 & 83.47 & \textcolor{gray}{23.19} & 213.08 & 46.64 & 72.45 & \textcolor{gray}{22.80} & 101.01 & 32.02 & 75.64 \\

UCE & 28.23 & \textcolor{gray}{41.92} & \textcolor{gray}{63.22} & \textcolor{gray}{44.60} & 27.07 & \textcolor{gray}{42.84} & \textcolor{gray}{53.72} & \textcolor{gray}{53.60} & \textcolor{gray}{28.47} & 36.21 & \underline{62.52} & \underline{40.10} & \textcolor{gray}{24.07} & \underline{21.25} & 74.63 & 30.84 & \textcolor{gray}{22.14} & 30.77 & 67.92 & 33.13 \\

RECE & 28.05 & \textcolor{gray}{49.65} & \textcolor{gray}{61.13} & \textcolor{gray}{47.78} & 27.04 & \textcolor{gray}{51.98} & \textcolor{gray}{51.14} & \textcolor{gray}{57.52} & \textcolor{gray}{28.41} & 37.46 & 60.94 & 41.98 & \textcolor{gray}{24.12} & 21.41 & 74.02 & 30.53 & \textcolor{gray}{22.10} & \underline{30.61} & 66.59 & 34.84 \\

NP & 28.19 & \textcolor{gray}{75.53} & \textcolor{gray}{41.62} & \textcolor{gray}{64.28} & 26.21 & \textcolor{gray}{67.77} & \textcolor{gray}{41.11} & \textcolor{gray}{64.77} & \textcolor{gray}{28.54} & 52.27 & 45.49 & 58.75 & \textcolor{gray}{24.46} & 48.63 & 47.89 & 61.76 & \textcolor{gray}{21.91} & 50.50 & 38.45 & 63.31 \\
SAFREE & 23.65 & \textcolor{gray}{85.40} & \textcolor{gray}{25.86} & \textcolor{gray}{80.42} & 23.89 & \textcolor{gray}{70.79} & \textcolor{gray}{32.28} & \textcolor{gray}{81.81} & \textcolor{gray}{27.75} & 62.18 & 30.03 & 77.60 & \textcolor{gray}{24.20} & 50.00 & 30.82 & 77.85 & \textcolor{gray}{21.44} & 56.22 & 22.95 & 82.29 \\
\textbf{Ours} & 22.76 & \textcolor{gray}{104.30} & \textcolor{gray}{37.70} & \textcolor{gray}{74.95} & 22.16 & \textcolor{gray}{99.74} & \textcolor{gray}{31.99} & \textcolor{gray}{80.63} & \textcolor{gray}{28.79} & \textbf{27.95} & \textbf{69.70} & \textbf{28.80} & \textcolor{gray}{24.29} & \textbf{22.26} & \textbf{75.51} & \textbf{28.38} & \textcolor{gray}{22.60} & \textbf{29.90} & \textbf{71.78} & \textbf{28.55} \\

\specialrule{\lightrulewidth}{0.5pt}{0.5pt}
\multicolumn{21}{c}{\textit{Erase \textbf{Snoopy}, \textbf{Mickey} and \textbf{SpongeBob}}} \\
\specialrule{\lightrulewidth}{0.5pt}{0.5pt}
Metric & CS $\downarrow$ & \textcolor{gray}{FID} & \textcolor{gray}{SSIM} & \textcolor{gray}{LPIPS} & CS $\downarrow$ & \textcolor{gray}{FID} & \textcolor{gray}{SSIM} & \textcolor{gray}{LPIPS} & CS $\downarrow$ & \textcolor{gray}{FID} & \textcolor{gray}{SSIM} & \textcolor{gray}{LPIPS} & \textcolor{gray}{CS} & FID $\downarrow$ & SSIM $\uparrow$ & LPIPS $\downarrow$ & \textcolor{gray}{CS} & FID $\downarrow$ & SSIM $\uparrow$ & LPIPS $\downarrow$ \\ 
\midrule
ESD & \underline{22.05} & \textcolor{gray}{195.72} & \textcolor{gray}{18.57} & \textcolor{gray}{86.49} & 26.52 & \textcolor{gray}{137.75} & \textcolor{gray}{20.51} & \textcolor{gray}{85.67} & \underline{20.87} & \textcolor{gray}{193.99} & \textcolor{gray}{24.00} & \textcolor{gray}{88.69} & \textcolor{gray}{22.79} & 72.48 & 33.83 & 77.94 & \textcolor{gray}{20.46} & 68.17 & 30.98 & 79.84 \\
CA & 28.32 & \textcolor{gray}{34.33} & \textcolor{gray}{55.71} & \textcolor{gray}{48.17} & 26.86 & \textcolor{gray}{31.51} & \textcolor{gray}{54.48} & \textcolor{gray}{49.29} & 28.51 & \textcolor{gray}{39.44} & \textcolor{gray}{53.95} & \textcolor{gray}{51.53} & \textcolor{gray}{24.23} & 33.39 & 63.98 & 43.98 & \textcolor{gray}{22.49} & 35.89 & 50.34 & 52.05 \\

SPM & 28.39 & \textcolor{gray}{38.23} & \textcolor{gray}{53.53} & \textcolor{gray}{46.35} & 26.94 & \textcolor{gray}{37.55} & \textcolor{gray}{61.23} & \textcolor{gray}{48.25} & 28.55 & \textcolor{gray}{32.98} & \textcolor{gray}{60.62} & \textcolor{gray}{48.50} & \textcolor{gray}{24.19} & 32.82 & \underline{64.69} & 45.88 & \textcolor{gray}{22.50} & \underline{29.22} & 66.49 & 34.15 \\

EA & \textbf{18.55} & \textcolor{gray}{212.30} & \textcolor{gray}{34.16} & \textcolor{gray}{83.69} & \textbf{18.52} & \textcolor{gray}{233.19} & \textcolor{gray}{35.95} & \textcolor{gray}{90.19} & \textbf{19.22} & \textcolor{gray}{257.32} & \textcolor{gray}{30.34} & \textcolor{gray}{92.07} & \textcolor{gray}{19.46} & 246.77 & 45.42 & 78.52 & \textcolor{gray}{19.34} & 189.95 & 29.59 & 84.51 \\

UCE & 28.23 & \textcolor{gray}{41.34} & \textcolor{gray}{63.11} & \textcolor{gray}{44.79} & 27.07 & \textcolor{gray}{40.35} & \textcolor{gray}{54.62} & \textcolor{gray}{52.19} & 28.54 & \textcolor{gray}{37.07} & \textcolor{gray}{59.11} & \textcolor{gray}{44.80} & \textcolor{gray}{24.07} & 30.90 & 64.48 & 41.43 & \textcolor{gray}{22.21} & 29.27 & \underline{67.37} & \underline{33.73} \\

RECE & 28.16 & \textcolor{gray}{46.10} & \textcolor{gray}{61.45} & \textcolor{gray}{47.41} & 27.08 & \textcolor{gray}{46.52} & \textcolor{gray}{52.38} & \textcolor{gray}{55.46} & 28.49 & \textcolor{gray}{38.86} & \textcolor{gray}{57.61} & \textcolor{gray}{46.71} & \textcolor{gray}{24.07} & \underline{30.74} & 64.62 & \underline{41.36} & \textcolor{gray}{22.08} & 29.36 & 66.37 & 34.86 \\

NP & 28.31 & \textcolor{gray}{81.00} & \textcolor{gray}{35.02} & \textcolor{gray}{65.05} & 26.15 & \textcolor{gray}{72.50} & \textcolor{gray}{36.26} & \textcolor{gray}{66.62} & 28.41 & \textcolor{gray}{87.16} & \textcolor{gray}{39.95} & \textcolor{gray}{64.20} & \textcolor{gray}{24.39} & 53.37 & 42.11 & 63.77 & \textcolor{gray}{21.75} & 53.81 & 34.83 & 63.69 \\
SAFREE & 23.87 & \textcolor{gray}{84.44} & \textcolor{gray}{19.89} & \textcolor{gray}{80.16} & \underline{24.89} & \textcolor{gray}{72.91} & \textcolor{gray}{17.97} & \textcolor{gray}{82.42} & 24.86 & \textcolor{gray}{101.00} & \textcolor{gray}{26.60} & \textcolor{gray}{79.76} & \textcolor{gray}{24.11} & 53.60 & 25.13 & 79.68 & \textcolor{gray}{21.41} & 57.21 & 19.18 & 83.15 \\
\textbf{Ours} & 22.09 & \textcolor{gray}{109.91} & \textcolor{gray}{35.82} & \textcolor{gray}{78.64} & 23.21 & \textcolor{gray}{156.07} & \textcolor{gray}{22.99} & \textcolor{gray}{89.09} & 23.06 & \textcolor{gray}{164.94} & \textcolor{gray}{24.44} & \textcolor{gray}{79.87} & \textcolor{gray}{24.84} & \textbf{27.89} & \textbf{64.73} & \textbf{41.21} & \textcolor{gray}{23.05} & \textbf{29.00} & \textbf{69.95} & \textbf{30.13} \\ 
\bottomrule
\end{tabular}%
}
\end{table*}

\begin{table*}[htbp]
\centering
\caption{Full quantitative results across all metrics for instance concept erasure on Switti-AR. The best and second-best results for each primary metric column are marked with \textbf{bold} and \underline{underline}. Columns in gray indicate items that do not directly reflect erasure efficacy or prior preservation performance.}
\label{tab:swittiar_snoopy_appendix_complete}
\renewcommand{\arraystretch}{1.4}

\resizebox{\textwidth}{!}{%
\begin{tabular}{l | cccc | cccc | cccc | cccc | cccc}
\toprule
Concept & \multicolumn{4}{c|}{Snoopy} & \multicolumn{4}{c|}{Mickey} & \multicolumn{4}{c|}{SpongeBob} & \multicolumn{4}{c|}{Dog} & \multicolumn{4}{c}{Legislator} \\ 
\cmidrule(r){1-1} \cmidrule(lr){2-5} \cmidrule(lr){6-9} \cmidrule(lr){10-13} \cmidrule(lr){14-17} \cmidrule(l){18-21}
Metric & CS & FID & SSIM & LPIPS & CS & FID & SSIM & LPIPS & CS & FID & SSIM & LPIPS & CS & FID & SSIM & LPIPS & CS & FID & SSIM & LPIPS \\ 
\midrule
Switti-AR & 28.11 & -- & -- & -- & 26.71 & -- & -- & -- & 29.09 & -- & -- & -- & 25.05 & -- & -- & -- & 21.28 & -- & -- & -- \\

\specialrule{\lightrulewidth}{0.5pt}{0.5pt}
\multicolumn{21}{c}{\textit{Erase \textbf{Snoopy}}} \\
\specialrule{\lightrulewidth}{0.5pt}{0.5pt}
Metric & CS $\downarrow$ & \textcolor{gray}{FID} & \textcolor{gray}{SSIM} & \textcolor{gray}{LPIPS} & \textcolor{gray}{CS} & FID $\downarrow$ & SSIM $\uparrow$ & LPIPS $\downarrow$ & \textcolor{gray}{CS} & FID $\downarrow$ & SSIM $\uparrow$ & LPIPS $\downarrow$ & \textcolor{gray}{CS} & FID $\downarrow$ & SSIM $\uparrow$ & LPIPS $\downarrow$ & \textcolor{gray}{CS} & FID $\downarrow$ & SSIM $\uparrow$ & LPIPS $\downarrow$ \\ 
\midrule

SPM    & 23.65 & \textcolor{gray}{40.46} & \textcolor{gray}{26.77} & \textcolor{gray}{79.98} & \textcolor{gray}{21.45} & \underline{41.34} & \underline{24.75} & \underline{80.82} & \textcolor{gray}{25.19} & \underline{38.08} & \underline{25.53} & \underline{73.55} & \textcolor{gray}{21.82} & \underline{33.01} & 22.71 & 82.59 & \textcolor{gray}{17.86} & \underline{51.57} & \underline{26.55} & \underline{80.38} \\
UCE    & 20.85 & \textcolor{gray}{108.75} & \textcolor{gray}{23.60} & \textcolor{gray}{83.59} & \textcolor{gray}{21.29} & 53.51 & 22.45 & 83.00 & \textcolor{gray}{25.05} & 70.19 & 23.21 & 81.74 & \textcolor{gray}{20.80} & 34.74 & \underline{22.79} & \underline{82.55} & \textcolor{gray}{16.70} & 56.30 & 23.41 & 82.19 \\
SAFREE & \underline{20.65} & \textcolor{gray}{94.03} & \textcolor{gray}{22.78} & \textcolor{gray}{83.92} & \textcolor{gray}{19.81} & 89.10 & 21.02 & 83.57 & \textcolor{gray}{24.77} & 62.68 & 22.77 & 81.62 & \textcolor{gray}{20.76} & 54.29 & 22.58 & 83.60 & \textcolor{gray}{16.78} & 79.39 & 22.58 & 83.60 \\
\textbf{Ours}  & \textbf{20.61} & \textcolor{gray}{95.85} & \textcolor{gray}{23.33} & \textcolor{gray}{82.45} & \textcolor{gray}{26.57} & \textbf{25.69} & \textbf{45.46} & \textbf{50.88} & \textcolor{gray}{28.16} & \textbf{24.12} & \textbf{47.01} & \textbf{49.88} & \textcolor{gray}{24.89} & \textbf{21.46} & \textbf{40.82} & \textbf{51.87} & \textcolor{gray}{20.91} & \textbf{32.38} & \textbf{37.82} & \textbf{55.87} \\ 
\bottomrule
\end{tabular}%
}
\end{table*}

\begin{table*}[htbp]
\centering
\caption{Full quantitative results across all metrics for art style concept erasure on FLUX. The best and second-best results for each primary metric column are marked with \textbf{bold} and \underline{underline}. Columns in gray indicate items that do not directly reflect erasure efficacy or prior preservation performance.}
\label{tab:appendix_flux_styles}
\renewcommand{\arraystretch}{1.4} 

\resizebox{\textwidth}{!}{%
\begin{tabular}{l | cccc | cccc | cccc | cccc | cccc}
\toprule
Concept & \multicolumn{4}{c|}{Pencil Sketch} & \multicolumn{4}{c|}{Van Gogh} & \multicolumn{4}{c|}{Stained Glass} & \multicolumn{4}{c|}{Pixel} & \multicolumn{4}{c}{Gothic} \\ 
\cmidrule(r){1-1} \cmidrule(lr){2-5} \cmidrule(lr){6-9} \cmidrule(lr){10-13} \cmidrule(lr){14-17} \cmidrule(l){18-21}
Metric & CS & FID & SSIM & LPIPS & CS & FID & SSIM & LPIPS & CS & FID & SSIM & LPIPS & CS & FID & SSIM & LPIPS & CS & FID & SSIM & LPIPS \\ 
\midrule
FLUX & 28.22 & -- & -- & -- & 24.51 & -- & -- & -- & 26.82 & -- & -- & -- & 27.91 & -- & -- & -- & 25.64 & -- & -- & -- \\

\specialrule{\lightrulewidth}{0.5pt}{0.5pt} 
\multicolumn{21}{c}{\textit{Erase \textbf{Pencil Sketch}}} \\
\specialrule{\lightrulewidth}{0.5pt}{0.5pt} 
Metric & CS $\downarrow$ & \textcolor{gray}{FID} & \textcolor{gray}{SSIM} & \textcolor{gray}{LPIPS} & \textcolor{gray}{CS} & FID $\downarrow$ & SSIM $\uparrow$ & LPIPS $\downarrow$ & \textcolor{gray}{CS} & FID $\downarrow$ & SSIM $\uparrow$ & LPIPS $\downarrow$ & \textcolor{gray}{CS} & FID $\downarrow$ & SSIM $\uparrow$ & LPIPS $\downarrow$ & \textcolor{gray}{CS} & FID $\downarrow$ & SSIM $\uparrow$ & LPIPS $\downarrow$ \\ 
\midrule

ESD    & 23.22 & \textcolor{gray}{159.35} & \textcolor{gray}{51.36} & \textcolor{gray}{85.78} & \textcolor{gray}{22.31} & 88.66 & 32.90 & 61.49 & \textcolor{gray}{24.33} & 103.62 & 30.38 & 57.75 & \textcolor{gray}{26.58} & 83.54 & 46.40 & 57.91 & \textcolor{gray}{23.24} & 89.30 & 39.49 & 61.43 \\
CA & 27.01 & \textcolor{gray}{19.72} & \textcolor{gray}{86.33} & \textcolor{gray}{12.77} & \textcolor{gray}{24.29} & \underline{22.68} & \textbf{87.79} & \textbf{11.23} & \textcolor{gray}{26.82} & \textbf{24.05} & \underline{76.65} & \underline{15.92} & \textcolor{gray}{27.84} & \textbf{25.02} & \textbf{87.17} & \textbf{12.26} & \textcolor{gray}{24.76} & \textbf{22.74} & \textbf{88.15} & \textbf{11.18} \\
SPM    & 26.50 & \textcolor{gray}{66.24} & \textcolor{gray}{74.59} & \textcolor{gray}{25.88} & \textcolor{gray}{24.40} & 53.42 & 59.08 & 34.88 & \textcolor{gray}{27.67} & 60.99 & 51.50 & 38.02 & \textcolor{gray}{27.87} & 58.30 & 68.32 & 35.37 & \textcolor{gray}{24.79} & 51.35 & 66.65 & 33.36 \\
EA     & \textbf{22.43} & \textcolor{gray}{111.57} & \textcolor{gray}{24.05} & \textcolor{gray}{58.81} & \textcolor{gray}{23.84} & 46.07 & 64.02 & 30.75 & \textcolor{gray}{26.05} & 54.89 & 56.24 & 34.12 & \textcolor{gray}{27.74} & 48.29 & 73.28 & 28.90 & \textcolor{gray}{24.36} & 43.82 & 72.64 & 27.39 \\
UCE    & 26.42 & \textcolor{gray}{65.92} & \textcolor{gray}{73.73} & \textcolor{gray}{26.75} & \textcolor{gray}{24.47} & 53.03 & 59.30 & 34.18 & \textcolor{gray}{27.62} & 60.41 & 51.84 & 37.60 & \textcolor{gray}{27.88} & 56.57 & 69.00 & 34.28 & \textcolor{gray}{24.62} & 52.80 & 65.66 & 34.63 \\
RECE   & 26.40 & \textcolor{gray}{65.88} & \textcolor{gray}{73.75} & \textcolor{gray}{26.79} & \textcolor{gray}{24.46} & 54.19 & 58.85 & 34.76 & \textcolor{gray}{27.63} & 60.94 & 51.50 & 37.88 & \textcolor{gray}{27.89} & 57.63 & 68.92 & 34.43 & \textcolor{gray}{24.53} & 51.00 & 66.50 & 33.41 \\
SAFREE & 25.35 & \textcolor{gray}{93.50} & \textcolor{gray}{64.61} & \textcolor{gray}{41.95} & \textcolor{gray}{23.15} & 69.43 & 52.50 & 44.06 & \textcolor{gray}{24.13} & 84.61 & 40.37 & 55.00 & \textcolor{gray}{25.42} & 90.03 & 58.98 & 50.74 & \textcolor{gray}{23.24} & 68.88 & 56.49 & 49.70 \\
\textbf{Ours} & \underline{22.92} & \textcolor{gray}{128.02} & \textcolor{gray}{43.84} & \textcolor{gray}{67.97} & \textcolor{gray}{24.45} & \textbf{22.44} & \underline{82.95} & \underline{12.70} & \textcolor{gray}{26.70} & \underline{27.94} & \textbf{78.53} & \textbf{14.99} & \textcolor{gray}{27.88} & \underline{28.35} & \underline{85.25} & \underline{14.73} & \textcolor{gray}{24.74} & \underline{25.43} & \underline{84.80} & \underline{13.95} \\

\specialrule{\lightrulewidth}{0.5pt}{0.5pt}
\multicolumn{21}{c}{\textit{Erase \textbf{Van Gogh}}} \\
\specialrule{\lightrulewidth}{0.5pt}{0.5pt}
Metric & \textcolor{gray}{CS} & FID $\downarrow$ & SSIM $\uparrow$ & LPIPS $\downarrow$ & CS $\downarrow$ & \textcolor{gray}{FID} & \textcolor{gray}{SSIM} & \textcolor{gray}{LPIPS} & \textcolor{gray}{CS} & FID $\downarrow$ & SSIM $\uparrow$ & LPIPS $\downarrow$ & \textcolor{gray}{CS} & FID $\downarrow$ & SSIM $\uparrow$ & LPIPS $\downarrow$ & \textcolor{gray}{CS} & FID $\downarrow$ & SSIM $\uparrow$ & LPIPS $\downarrow$ \\ 
\midrule

ESD    & \textcolor{gray}{25.34} & 113.72 & 55.52 & 57.28 & 22.67 & \textcolor{gray}{153.30} & \textcolor{gray}{32.10} & \textcolor{gray}{67.57} & \textcolor{gray}{23.85} & 142.90 & 33.73 & 73.38 & \textcolor{gray}{26.12} & 100.99 & 57.20 & 63.47 & \textcolor{gray}{22.24} & 113.21 & 47.86 & 76.14 \\
CA & \textcolor{gray}{27.51} & \underline{27.92} & \underline{86.83} & \underline{12.16} & 23.37 & \textcolor{gray}{31.15} & \textcolor{gray}{81.17} & \textcolor{gray}{5.13} & \textcolor{gray}{27.68} & \textbf{28.81} & \textbf{82.32} & \textbf{14.54} & \textcolor{gray}{27.85} & \underline{33.45} & \underline{79.75} & \underline{19.20} & \textcolor{gray}{25.16} & \textbf{26.78} & \textbf{85.36} & \textbf{12.97} \\
SPM    & \textcolor{gray}{27.47} & 65.87 & 74.61 & 25.84 & 24.39 & \textcolor{gray}{54.00} & \textcolor{gray}{58.91} & \textcolor{gray}{34.95} & \textcolor{gray}{27.65} & 61.05 & 51.47 & 38.03 & \textcolor{gray}{27.91} & 58.29 & 68.32 & 35.40 & \textcolor{gray}{25.17} & 51.23 & 66.57 & 33.43 \\
EA     & \textcolor{gray}{27.30} & 48.08 & 82.77 & 16.14 & \underline{21.68} & \textcolor{gray}{91.34} & \textcolor{gray}{45.27} & \textcolor{gray}{49.88} & \textcolor{gray}{27.57} & 41.02 & 66.65 & 23.92 & \textcolor{gray}{27.75} & 42.08 & 78.45 & 21.41 & \textcolor{gray}{25.08} & 34.88 & 76.74 & 20.63 \\
UCE    & \textcolor{gray}{27.39} & 66.26 & 74.52 & 26.01 & 24.11 & \textcolor{gray}{62.34} & \textcolor{gray}{54.54} & \textcolor{gray}{41.03} & \textcolor{gray}{27.65} & 61.51 & 51.45 & 38.65 & \textcolor{gray}{27.78} & 58.81 & 68.40 & 35.25 & \textcolor{gray}{25.12} & 55.45 & 64.02 & 37.01 \\
RECE   & \textcolor{gray}{27.24} & 66.92 & 74.38 & 26.25 & 24.10 & \textcolor{gray}{62.27} & \textcolor{gray}{54.51} & \textcolor{gray}{41.07} & \textcolor{gray}{27.61} & 61.53 & 51.18 & 38.91 & \textcolor{gray}{27.66} & 59.32 & 66.44 & 40.25 & \textcolor{gray}{25.16} & 55.98 & 63.89 & 37.19 \\
SAFREE & \textcolor{gray}{26.37} & 55.97 & 68.18 & 34.97 & 22.24 & \textcolor{gray}{63.89} & \textcolor{gray}{52.62} & \textcolor{gray}{44.26} & \textcolor{gray}{26.13} & 63.59 & 49.82 & 41.19 & \textcolor{gray}{26.59} & 68.58 & 63.84 & 43.98 & \textcolor{gray}{24.01} & 64.88 & 58.95 & 44.91 \\
\textbf{Ours} & \textcolor{gray}{27.36} & \textbf{25.15} & \textbf{87.07} & \textbf{11.08} & \textbf{21.64} & \textcolor{gray}{84.18} & \textcolor{gray}{43.74} & \textcolor{gray}{55.31} & \textcolor{gray}{27.67} & \underline{33.10} & \underline{73.33} & \underline{18.77} & \textcolor{gray}{27.82} & \textbf{32.46} & \textbf{81.80} & \textbf{17.38} & \textcolor{gray}{25.19} & \underline{29.91} & \underline{80.56} & \underline{16.52} \\

\specialrule{\lightrulewidth}{0.5pt}{0.5pt}
\multicolumn{21}{c}{\textit{Erase \textbf{Stained Glass}}} \\
\specialrule{\lightrulewidth}{0.5pt}{0.5pt}
Metric & \textcolor{gray}{CS} & FID $\downarrow$ & SSIM $\uparrow$ & LPIPS $\downarrow$ & \textcolor{gray}{CS} & FID $\downarrow$ & SSIM $\uparrow$ & LPIPS $\downarrow$ & CS $\downarrow$ & \textcolor{gray}{FID} & \textcolor{gray}{SSIM} & \textcolor{gray}{LPIPS} & \textcolor{gray}{CS} & FID $\downarrow$ & SSIM $\uparrow$ & LPIPS $\downarrow$ & \textcolor{gray}{CS} & FID $\downarrow$ & SSIM $\uparrow$ & LPIPS $\downarrow$ \\ 
\midrule
ESD    & \textcolor{gray}{26.19} & 102.82 & 29.71 & 78.44 & \textcolor{gray}{22.15} & 103.91 & 38.38 & 67.22 & \textbf{22.06} & \textcolor{gray}{166.13} & \textcolor{gray}{29.71} & \textcolor{gray}{78.44} & \textcolor{gray}{25.26} & 90.58 & 61.21 & 54.66 & \textcolor{gray}{22.44} & 94.88 & 49.05 & 64.59 \\
CA & \textcolor{gray}{27.52} & \textbf{23.44} & \textbf{78.16} & \textbf{17.43} & \textcolor{gray}{24.37} & \textbf{26.08} & \textbf{85.11} & \underline{22.58} & 26.68 & \textcolor{gray}{33.44} & \textcolor{gray}{78.16} & \textcolor{gray}{27.43} & \textcolor{gray}{27.85} & \underline{29.46} & \underline{85.16} & \underline{16.87} & \textcolor{gray}{25.03} & \textbf{16.18} & \underline{75.86} & \underline{22.62} \\
SPM    & \textcolor{gray}{27.52} & 66.35 & 74.60 & 25.88 & \textcolor{gray}{24.39} & 53.52 & 59.06 & 34.89 & 26.64 & \textcolor{gray}{61.01} & \textcolor{gray}{51.52} & \textcolor{gray}{38.00} & \textcolor{gray}{27.88} & 58.37 & 68.35 & 35.37 & \textcolor{gray}{25.16} & 50.90 & 66.66 & 33.32 \\
EA     & \textcolor{gray}{26.83} & 61.40 & 53.30 & 36.48 & \textcolor{gray}{24.22} & 49.83 & 63.64 & 29.86 & 24.65 & \textcolor{gray}{89.31} & \textcolor{gray}{43.30} & \textcolor{gray}{58.48} & \textcolor{gray}{27.42} & 50.71 & 73.31 & 28.08 & \textcolor{gray}{24.92} & 47.50 & 71.14 & 27.72 \\
UCE    & \textcolor{gray}{27.50} & 64.62 & 75.00 & 25.32 & \textcolor{gray}{24.39} & 53.80 & 59.06 & 34.74 & 26.85 & \textcolor{gray}{64.49} & \textcolor{gray}{45.67} & \textcolor{gray}{43.31} & \textcolor{gray}{28.00} & 56.62 & 68.72 & 34.80 & \textcolor{gray}{25.10} & 53.98 & 64.72 & 35.97 \\
RECE   & \textcolor{gray}{26.93} & 78.07 & 60.12 & 44.58 & \textcolor{gray}{24.28} & 59.72 & 62.77 & 37.52 & 26.33 & \textcolor{gray}{80.37} & \textcolor{gray}{37.24} & \textcolor{gray}{54.04} & \textcolor{gray}{27.66} & 89.61 & 58.97 & 50.24 & \textcolor{gray}{24.91} & 60.50 & 55.69 & 47.31 \\
SAFREE & \textcolor{gray}{26.67} & 86.02 & 46.62 & 46.95 & \textcolor{gray}{23.36} & 61.48 & 54.35 & 42.12 & 25.98 & \textcolor{gray}{78.78} & \textcolor{gray}{46.62} & \textcolor{gray}{46.95} & \textcolor{gray}{25.95} & 79.31 & 60.03 & 49.54 & \textcolor{gray}{24.02} & 64.32 & 60.22 & 43.71 \\
\textbf{Ours} & \textcolor{gray}{27.57} & \underline{25.58} & \underline{76.68} & \underline{20.02} & \textcolor{gray}{24.16} & \underline{27.53} & \underline{73.41} & \textbf{20.69} & \underline{23.03} & \textcolor{gray}{163.02} & \textcolor{gray}{30.68} & \textcolor{gray}{72.02} & \textcolor{gray}{27.86} & \textbf{28.72} & \textbf{86.63} & \textbf{15.84} & \textcolor{gray}{24.99} & \underline{19.85} & \textbf{76.50} & \textbf{22.13} \\ 

\bottomrule
\end{tabular}%
}
\end{table*}

\begin{table*}[htbp]
\centering
\caption{Full quantitative results across all metrics for art style concept erasure on SD v3. The best and second-best results for each primary metric column are marked with \textbf{bold} and \underline{underline}. Columns in gray indicate items that do not directly reflect erasure efficacy or prior preservation performance.}
\label{tab:SD_v3_pencil_sketch_appendix}
\renewcommand{\arraystretch}{1.4} 

\resizebox{\textwidth}{!}{%
\begin{tabular}{l | cccc | cccc | cccc | cccc | cccc}
\toprule
Concept & \multicolumn{4}{c|}{Pencil Sketch} & \multicolumn{4}{c|}{Van Gogh} & \multicolumn{4}{c|}{Stained Glass} & \multicolumn{4}{c|}{Pixel} & \multicolumn{4}{c}{Gothic} \\ 
\cmidrule(r){1-1} \cmidrule(lr){2-5} \cmidrule(lr){6-9} \cmidrule(lr){10-13} \cmidrule(lr){14-17} \cmidrule(l){18-21}
Metric & CS & FID & SSIM & LPIPS & CS & FID & SSIM & LPIPS & CS & FID & SSIM & LPIPS & CS & FID & SSIM & LPIPS & CS & FID & SSIM & LPIPS \\ 
\midrule
SD v3 & 29.35 & -- & -- & -- & 25.79 & -- & -- & -- & 29.35 & -- & -- & -- & 29.99 & -- & -- & -- & 26.35 & -- & -- & -- \\

\specialrule{\lightrulewidth}{0.5pt}{0.5pt}
\multicolumn{21}{c}{\textit{Erase \textbf{Pencil Sketch}}} \\
\specialrule{\lightrulewidth}{0.5pt}{0.5pt}
Metric & CS $\downarrow$ & \textcolor{gray}{FID} & \textcolor{gray}{SSIM} & \textcolor{gray}{LPIPS} & \textcolor{gray}{CS} & FID $\downarrow$ & SSIM $\uparrow$ & LPIPS $\downarrow$ & \textcolor{gray}{CS} & FID $\downarrow$ & SSIM $\uparrow$ & LPIPS $\downarrow$ & \textcolor{gray}{CS} & FID $\downarrow$ & SSIM $\uparrow$ & LPIPS $\downarrow$ & \textcolor{gray}{CS} & FID $\downarrow$ & SSIM $\uparrow$ & LPIPS $\downarrow$ \\ 
\midrule 
ESD    & 25.57 & \textcolor{gray}{139.47} & \textcolor{gray}{18.08} & \textcolor{gray}{80.47} & \textcolor{gray}{24.24} & 136.24 & 38.48 & 72.59 & \textcolor{gray}{25.83} & 145.15 & 21.02 & 83.64 & \textcolor{gray}{25.63} & 166.72 & 26.17 & 84.25 & \textcolor{gray}{24.20} & 140.40 & 21.24 & 83.33 \\
CA & 28.23 & \textcolor{gray}{79.83} & \textcolor{gray}{25.52} & \textcolor{gray}{67.47} & \textcolor{gray}{25.72} & 73.42 & 41.54 & 69.51 & \textcolor{gray}{29.14} & 74.71 & 38.29 & 62.72 & \textcolor{gray}{29.62} & 74.44 & 34.14 & 68.83 & \textcolor{gray}{26.10} & 69.94 & 22.54 & 68.87 \\
EA     & 25.53 & \textcolor{gray}{212.85} & \textcolor{gray}{28.70} & \textcolor{gray}{77.16} & \textcolor{gray}{25.76} & 128.29 & 36.59 & 70.63 & \textcolor{gray}{28.78} & 142.71 & 37.49 & 65.55 & \textcolor{gray}{28.30} & 150.35 & 21.79 & 70.23 & \textcolor{gray}{25.54} & 149.50 & 25.90 & 70.85 \\

SPM    & 28.21 & \textcolor{gray}{56.90} & \textcolor{gray}{54.67} & \textcolor{gray}{54.94}  & \textcolor{gray}{25.73} & 53.75  & 53.37  & \underline{57.03}  & \textcolor{gray}{29.28} & 56.39  & 50.37  & 56.76  & \textcolor{gray}{29.01} & \underline{50.28}  & 52.31  & \underline{57.07}  & \textcolor{gray}{25.81} & \underline{54.70}  & \underline{54.99}  & \underline{54.80}  \\
UCE    & 27.43 & \textcolor{gray}{81.68} & \textcolor{gray}{56.96} & \textcolor{gray}{54.78} & \textcolor{gray}{25.72} & \underline{47.10}  & \underline{58.46}  & 57.69 & \textcolor{gray}{29.27} & \underline{53.10}  & 54.42  & \underline{51.00} & \textcolor{gray}{29.16} & 52.22  & \underline{53.34}  & 57.12 & \textcolor{gray}{25.44} & 55.38  & 53.18  & 58.97 \\
RECE   & 27.18 & \textcolor{gray}{84.71} & \textcolor{gray}{53.91} & \textcolor{gray}{58.49} & \textcolor{gray}{25.70} & 53.44  & 53.11  & 58.59 & \textcolor{gray}{29.23} & 57.35  & \underline{54.74}  & 55.25 & \textcolor{gray}{29.02} & 55.57  & 50.29  & 59.83 & \textcolor{gray}{25.43} & 60.02  & 50.28  & 54.84 \\

NP     & 24.52 & \textcolor{gray}{119.64} & \textcolor{gray}{18.71} & \textcolor{gray}{72.48} & \textcolor{gray}{24.72} & 85.84 & 33.94 & 73.30 & \textcolor{gray}{28.21} & 84.55 & 28.16 & 67.70 & \textcolor{gray}{28.25} & 93.03 & 21.45 & 72.18 & \textcolor{gray}{25.43} & 85.00 & 22.82 & 71.93 \\
SAFREE & \underline{23.99} & \textcolor{gray}{119.55} & \textcolor{gray}{19.75} & \textcolor{gray}{71.56} & \textcolor{gray}{24.53} & 81.67 & 35.01 & 70.70 & \textcolor{gray}{28.12} & 82.66 & 30.45 & 66.53 & \textcolor{gray}{28.06} & 90.27 & 20.72 & 71.23 & \textcolor{gray}{24.85} & 81.80 & 23.50 & 70.43 \\
\textbf{Ours} & \textbf{21.65} & \textcolor{gray}{210.85} & \textcolor{gray}{11.25} & \textcolor{gray}{80.43} & \textcolor{gray}{25.75} & \textbf{45.90} & \textbf{59.56} & \textbf{56.84} & \textcolor{gray}{29.25} & \textbf{41.37} & \textbf{54.92} & \textbf{44.15} & \textcolor{gray}{29.85} & \textbf{40.78} & \textbf{59.30} & \textbf{47.55} & \textcolor{gray}{26.08} & \textbf{46.13} & \textbf{52.13} & \textbf{46.89} \\ 
\bottomrule
\end{tabular}%
}
\end{table*}

\begin{table*}[t]
\centering
\caption{Full quantitative results across all metrics for celebrity concept erasure on FLUX. The best and second-best results for each primary metric column are marked with \textbf{bold} and \underline{underline}. Columns in gray indicate items that do not directly reflect erasure efficacy or prior preservation performance.}
\label{tab:face_erasure_appendix}
\renewcommand{\arraystretch}{1.3} 

\resizebox{\textwidth}{!}{%
\begin{tabular}{l | cccc | cccc | cccc | cccc | cccc}
\toprule
Concept & \multicolumn{4}{c|}{Bruce Lee} & \multicolumn{4}{c|}{Marilyn Monroe} & \multicolumn{4}{c|}{Melania Trump} & \multicolumn{4}{c|}{Anne Hathaway} & \multicolumn{4}{c}{Tom Cruise} \\ 
\cmidrule(r){1-1} \cmidrule(lr){2-5} \cmidrule(lr){6-9} \cmidrule(lr){10-13} \cmidrule(lr){14-17} \cmidrule(l){18-21}
Metric & CS & FID & SSIM & LPIPS & CS & FID & SSIM & LPIPS & CS & FID & SSIM & LPIPS & CS & FID & SSIM & LPIPS & CS & FID & SSIM & LPIPS \\ 
\midrule
FLUX & 27.00 & -- & -- & -- & 26.98 & -- & -- & -- & 26.49 & -- & -- & -- & 23.94 & -- & -- & -- & 28.53 & -- & -- & -- \\

\specialrule{\lightrulewidth}{0.5pt}{0.5pt}
\multicolumn{21}{c}{\textit{Erase \textbf{Bruce Lee}}} \\
\specialrule{\lightrulewidth}{0.5pt}{0.5pt}
Metric & CS $\downarrow$ & \textcolor{gray}{FID} & \textcolor{gray}{SSIM} & \textcolor{gray}{LPIPS} & \textcolor{gray}{CS} & FID $\downarrow$ & SSIM $\uparrow$ & LPIPS $\downarrow$ & \textcolor{gray}{CS} & FID $\downarrow$ & SSIM $\uparrow$ & LPIPS $\downarrow$ & \textcolor{gray}{CS} & FID $\downarrow$ & SSIM $\uparrow$ & LPIPS $\downarrow$ & \textcolor{gray}{CS} & FID $\downarrow$ & SSIM $\uparrow$ & LPIPS $\downarrow$ \\ 
\midrule

CA & 24.92 & \textcolor{gray}{21.99} & \textcolor{gray}{86.14} & \textcolor{gray}{23.88} & \textcolor{gray}{25.04} & \underline{25.92} & \underline{87.89} & \underline{12.06} & \textcolor{gray}{25.47} & \underline{25.59} & \underline{88.18} & \underline{11.58} & \textcolor{gray}{22.95} & \underline{25.74} & \underline{88.18} & \underline{11.72} & \textcolor{gray}{27.49} & \underline{24.02} & \underline{88.37} & \underline{11.51} \\
ESD    & \underline{19.84} & \textcolor{gray}{151.91} & \textcolor{gray}{47.97} & \textcolor{gray}{63.37} & \textcolor{gray}{21.73} & 104.87 & 52.44 & 55.59 & \textcolor{gray}{21.53} & 88.33 & 54.51 & 53.11 & \textcolor{gray}{21.06} & 105.82 & 55.40 & 54.00 & \textcolor{gray}{22.18} & 95.38 & 54.26 & 55.32 \\
EA     & \textbf{18.75} & \textcolor{gray}{106.76} & \textcolor{gray}{51.40} & \textcolor{gray}{58.86} & \textcolor{gray}{22.05} & 73.65 & 58.72 & 47.30 & \textcolor{gray}{23.08} & 57.39 & 65.42 & 38.09 & \textcolor{gray}{22.15} & 63.35 & 65.70 & 39.91 & \textcolor{gray}{23.88} & 52.10 & 63.05 & 42.23 \\

UCE    & 25.78 & \textcolor{gray}{96.96} & \textcolor{gray}{42.83} & \textcolor{gray}{67.86} & \textcolor{gray}{24.81} & 63.59 & 42.40 & 64.14 & \textcolor{gray}{25.45} & 62.81 & 41.34 & 64.34 & \textcolor{gray}{21.78} & 70.35 & 44.46 & 63.47 & \textcolor{gray}{27.90} & 48.82 & 43.90 & 62.58 \\
RECE   & 26.59 & \textcolor{gray}{84.02} & \textcolor{gray}{43.30} & \textcolor{gray}{66.47} & \textcolor{gray}{24.98} & 63.98 & 42.33 & 64.34 & \textcolor{gray}{25.62} & 61.58 & 41.37 & 64.19 & \textcolor{gray}{22.13} & 69.84 & 44.37 & 63.74 & \textcolor{gray}{28.28} & 49.83 & 43.93 & 62.56 \\
SPM    & 26.97 & \textcolor{gray}{92.94} & \textcolor{gray}{43.68} & \textcolor{gray}{66.22} & \textcolor{gray}{25.20} & 72.57 & 42.70 & 63.51 & \textcolor{gray}{25.85} & 70.96 & 41.40 & 64.14 & \textcolor{gray}{22.17} & 80.49 & 44.97 & 63.14 & \textcolor{gray}{28.22} & 55.79 & 44.78 & 61.65 \\

SAFREE & 24.43 & \textcolor{gray}{70.82} & \textcolor{gray}{61.83} & \textcolor{gray}{45.68} & \textcolor{gray}{21.99} & 59.21 & 61.75 & 44.09 & \textcolor{gray}{21.81} & 67.59 & 57.60 & 48.85 & \textcolor{gray}{21.75} & 59.46 & 67.45 & 39.94 & \textcolor{gray}{22.94} & 49.55 & 60.50 & 46.95 \\
\textbf{Ours} & 21.06 & \textcolor{gray}{80.63} & \textcolor{gray}{60.10} & \textcolor{gray}{46.94} & \textcolor{gray}{25.21} & \textbf{21.60} & \textbf{90.52} & \textbf{10.06} & \textcolor{gray}{25.52} & \textbf{13.26} & \textbf{88.90} & \textbf{10.90} & \textcolor{gray}{22.94} & \textbf{14.45} & \textbf{93.69} & \textbf{8.22} & \textcolor{gray}{27.71} & \textbf{14.15} & \textbf{92.74} & \textbf{7.42} \\ 
\bottomrule
\end{tabular}%
}
\end{table*}

\end{document}